\documentclass[letterpaper]{article} % DO NOT CHANGE THIS
\usepackage{aaai2027}  % DO NOT CHANGE THIS
\usepackage[hyphens]{url}  % DO NOT CHANGE THIS
\usepackage{graphicx} % DO NOT CHANGE THIS
\usepackage{natbib}  % DO NOT CHANGE THIS AND DO NOT ADD ANY OPTIONS TO IT
\usepackage{caption} % DO NOT CHANGE THIS AND DO NOT ADD ANY OPTIONS TO IT
\usepackage{booktabs}
\usepackage{amsmath}
\usepackage{amssymb}
\usepackage{threeparttable}
\usepackage{multirow}
\usepackage{xspace}

\newcommand{\cmark}{\checkmark}
\newcommand{\xmark}{$\times$}
\newcommand{\tabincell}[2]{\begin{tabular}{@{}#1@{}}#2\end{tabular}}

\title{Eco3S: Complex Socio-Economic System Simulation via Agent-Based Models}
\author{
Shaopeng Wei\textsuperscript{\rm 1},
Yufei Cheng\textsuperscript{\rm 2}\corresponding,
Wenxi Sun\textsuperscript{\rm 2},
Yepeng Ding\textsuperscript{\rm 3},
Yu Zhao\textsuperscript{\rm 2}\corresponding, and
Gang Kou\textsuperscript{\rm 4}\corresponding
}
\affiliations{
\textsuperscript{\rm 1}School of Business, Guangxi University, Nanning 530004, China\\
\textsuperscript{\rm 2}School of Computing and Artificial Intelligence, Southwestern University of Finance and Economics, Chengdu 611130, China\\
\textsuperscript{\rm 3}Information Media Center, Hiroshima University, Hiroshima 739-8511, Japan\\
\textsuperscript{\rm 4}School of Business Administration, Faculty of Business Administration, Southwestern University of Finance and Economics, Chengdu 611130, China\\
shaopeng.wei@gxu.edu.cn, yufeicheng24@gmail.com, 2241201z5016@smail.swufe.edu.cn, yepengd@acm.org, zhaoyu@swufe.edu.cn, kougang@swufe.edu.cn
}

\begin{document}

\maketitle

\begin{abstract}
The rapid development of large language models (LLMs) has renewed interest in agent-based modeling (ABM). However, current LLM-based ABM research faces several key challenges: modeling evolving agent-environment interactions, enabling flexible counterfactual reasoning, and automating simulation workflows for scientific research.
In this paper, we propose \textbf{\textit{Eco3S}}, a socio-economic system simulation framework for economic research and policy analysis that addresses these challenges through three key mechanisms:
(1) \textbf{Co-evolving Environment Design}, a bidirectional feedback loop where agents and the environment co-evolve, producing realistic emergent behaviors; (2) \textbf{Structural Causal Simulation}, a structural causal model (SCM)-inspired counterfactual mechanism that allows flexible interventions for diverse causal inference tasks;
(3) \textbf{Simulation-Analysis-Refinement Paradigm}, a self-corrective mechanism that iteratively refines experimental designs based on prior simulation results.
Experiments on diverse economic scenarios confirm \textit{Eco3S}'s effectiveness in replicating multiple established economic studies (canal decay, origins of governance, and information propagation) and phenomena across domains.
Additional results further demonstrate its scalability and generalizability, highlighting the framework's potential for rigorous economic research and policy-making.
\end{abstract}


\section{Introduction}\label{sec:Intro}
The rapid advancement of large language models (LLMs) has profoundly reshaped numerous disciplines in recent years. Their impact extends far beyond traditional domains of computer science, such as natural language processing~\citep{achiam2023gpt,guo2025deepseek}, to encompass a broad spectrum of social sciences, including economics~\citep{li2024econagent}, finance~\citep{zhao2025alphaagents,yang2025twinmarket}, sociology~\citep{zhang2024electionsim}, and psychology~\citep{demszky2023using}. A particularly influential methodology transformed by this convergence is agent-based modeling (ABM)~\citep{farmer2009economy}, which employs computational agents capable of autonomous perception, decision-making, and action to simulate complex socio-economic dynamics. While early ABM approaches relied on hand-crafted behavioral rules, recent studies have increasingly replaced these with LLMs to endow agents with more nuanced, context-sensitive, and human-like reasoning capabilities~\citep{piao2025agentsociety,wang2025yulanonesim}. This paradigm shift significantly enhances the fidelity and realism of socio-economic simulations, bringing artificial societies closer to the complexities of real world.

In economic research, traditional approaches like econometrics ~\citep{hayashi2011econometrics} and dynamic stochastic general equilibrium (DSGE) models ~\citep{komunjer2011dynamic} often struggle with structural changes and neglect individual heterogeneity. Consequently, agent-based modeling has emerged as a vital alternative~\citep{farmer2009economy}. By allowing agents to make decisions based on unique states rather than uniform rules, ABM captures complex, non-linear, and emergent dynamics that traditional aggregate models overlook.

Currently, numerous studies employ LLMs as the backbone of agents to simulate various economic scenarios. While these works typically define agents, action rules, and environments based on specific situations under investigation~\citep{wang2025yulanonesim,zhang2025socioverse,piao2025agentsociety}, a critical scientific gap persists: how to effectively model and integrate the co-evolutionary dynamics between LLM agents and their complex, dynamically changing physical and social environments.
% Most existing approaches, as summarized in Table \ref{tab:studies_comparison}, often simplify or neglect the endogenous evolution of the physical environment.
Most existing approaches simplify or neglect the endogenous evolution of the physical environment. 
% These works typically define agents, action rules, and environments based on the specific situation under investigation~\citep{wang2025yulanonesim,zhang2025socioverse,piao2025agentsociety}. Each agent is endowed with a profile comprising multiple attributes, and its actions are influenced both by these initialized characteristics and by the dynamic interaction among agents. However, as shown in Table \ref{tab:studies_comparison}, most existing approaches neglect the dynamical evolution of the physical environment, which plays a crucial role in many economic contexts.
Moreover, prior research has primarily focused on replicating real-world phenomena to validate simulation systems, with little attention devoted to rigorous engagement with established economic studies.
Furthermore, the majority of existing frameworks are designed as fixed architectures tailored to specific case studies, which severely constrains their extensibility and broader applicability in economic research and policy analysis.

In this paper, we propose \textit{Eco3S}, a novel LLM-based socio-economic system simulation framework designed for economic research and policy analysis. The framework structures the simulation process into three core modules: configuration, simulation, and auto-analysis. The configuration module initializes simulation scenarios through a Co-evolving Environment Design, establishing a bidirectional feedback loop where agents and the environment co-evolve, enabling the system to capture complex interactions and reproduce realistic emergent behaviors. In the simulation module, we incorporate the principles of Structural Causal Models (SCM) \citep{Pearl_2009} and treat agent behaviors and environment dynamics as a causal structure, enabling flexible interventions and counterfactual reasoning. The auto-analysis module automatically synthesizes simulation statistics and presents them through charts and structured tables. Building upon these modules, we introduce an auto-simulation system based on the Simulation-Analysis-Refinement (SAR) paradigm. This paradigm automates the research process by iteratively evaluating and refining simulation designs based on prior outcomes to ensure alignment with research objectives. We conducted extensive experiments, including manually configured and automatically generated simulations, to validate the effectiveness and versatility of \textit{Eco3S}. 

In summary, the contributions of this work are fourfold:
\begin{itemize}
    \item We propose a \textbf{Co-evolving Environment Design} that captures bidirectional feedback loops between agents and environment to simulate the emergence of complex socio-economic phenomena.
    \item We introduce a \textbf{Structural Causal Simulation} paradigm grounded in SCM that integrates causal principles into ABM to support flexible interventions and counterfactual reasoning.
    \item We propose the \textbf{Simulation-Analysis-Refinement(SAR) paradigm}, a self-corrective mechanism that transforms natural language prompts into executable socio-economic simulations by iteratively refining experimental designs based on prior outcomes.
    % \item Through a series of experiments on established economic scenarios, we demonstrate the framework's effectiveness, scalability, and generalizability.
    \item We conduct extensive experiments, including replicating established economic studies and automatically simulating classical economic phenomena. The results show that \textit{Eco3S} successfully captures complex dynamics, demonstrating its effectiveness, scalability, and generalizability.
\end{itemize}

\section{Related Work}
\subsection{Agent-Based Modeling}

Agent-Based Modeling (ABM)~\citep{farmer2009economy} is a computational paradigm that simulates complex systems through autonomous agents capable of perception, decision-making, and action. A canonical ABM comprises three core components: a set of agents, a set of relationships among them, and the shared environment in which they operate~\citep{macal2010tutorial}.
% Over recent decades, ABM methodologies have evolved significantly from early rule-based systems~\citep{hu2023hierarchical} to reinforcement learning (RL)-driven agents endowed with adaptive learning capabilities~\citep{huang2025competing,Bui2024Mimicking,lee2024episodic}. The advent of LLMs has catalyzed a further paradigm shift in agent design, owing to their remarkable proficiency in language understanding, contextual reasoning, and generative capabilities. Unlike traditional RL agents which rely heavily on extensive environmental interaction and meticulously engineered reward functions, LLM-based agents can execute sophisticated tasks by drawing on vast pretrained knowledge, following natural language instructions, and reasoning over context, often without explicit reward signals.
Over recent decades, ABM methodologies transitioned from rule-based systems~\citep{hu2023hierarchical} to reinforcement learning (RL) agents with adaptive capabilities~\citep{huang2025competing,Bui2024Mimicking,lee2024episodic,yang2025deep,wang2025multi}. The rise of LLMs has driven a shift in agent design through their proficiency in linguistic understanding and contextual reasoning. Unlike RL agents dependent on extensive environmental interaction and engineered reward functions, LLM-based agents execute complex tasks by leveraging pretrained knowledge and reasoning over context without explicit reward signals. This paradigm facilitates context-sensitive behavior in weakly structured environments.
% Current research on LLM-based agents centers on several key directions. A primary focus lies in agent architecture design ~\citep{Wang2025User,tang2024medagents,Ren2024Emergence,mou2024unveiling}, which seeks to create flexible, modular systems capable of adapting their perception, decision-making, and behavioral strategies to meet specific task requirements. Another significant research trajectory examines social phenomena within multi-agent environments, such as group behavior and information diffusion ~\citep{Li2024Large,gao2023s3}. Concurrently, role-playing and identity modeling have emerged as prominent themes ~\citep{xu2024mindecho}, enhancing agent individuality and linguistic variation by embedding diverse backgrounds, goals, and motivations.
Current research on LLM agents follows several key directions. Architecture design focuses on modular systems that adapt perception and strategies across diverse tasks~\citep{Wang2025User,tang2024medagents,Ren2024Emergence,mou2024unveiling}. Multi-agent studies investigate emergent social phenomena, such as group behavior and information diffusion~\citep{Li2024Large,gao2023s3}. Concurrently, role-playing and identity modeling promote agent individuality and linguistic diversity by embedding specific backgrounds and goals~\citep{xu2024mindecho}.
% As multi-agent systems evolve, increasing attention is being paid to variations in population size and interaction modes. The field has progressed from small-scale systems examining individual agent behaviors ~\citep{zhao2024lyfe}, to medium-scale frameworks that simulate complex group dynamics ~\citep{zhao2024competeai,Wang2025User,becker2025mallm}, and further to large-scale architectures designed to generate emergent, macro-level outcomes ~\citep{yang2024oasis,wang2025yulanonesim}.
Multi-agent research has expanded toward diverse population scales and interaction modes. The field spans small-scale individual behavior~\citep{zhao2024lyfe}, medium-scale group dynamics~\citep{zhao2024competeai,Wang2025User,becker2025mallm}, and large-scale architectures targeting emergent macro-level outcomes~\citep{yang2024oasis,wang2025yulanonesim}.

However, most existing studies are tailored to specific scenarios, and the automated simulation of arbitrary scenarios guided by human feedback remains largely unexplored.

\subsection{Social System Simulation}
% Social system simulation employs computational models and virtual environments to replicate the behaviors of individuals, groups, and institutional structures. Its primary objectives include uncovering the generative mechanisms of complex social phenomena, forecasting social trajectories, and forming evidence-based policy design~\citep{li2023quantifying}. Recent advances have increasingly used LLM-based agents to simulate a diverse array of real-world social systems, including healthcare~\citep{chopra2025limits,tang2024medagents,li2024agent}, electoral processes~\citep{zhang2024electionsim,touzel2024simulation}, social media dynamics~\citep{liu2024skepticism,wang2025decoding,rossetti2024social,piao2025agentsociety}, and economic systems~\citep{li2024econagent,gao2024simulating,yang2025twinmarket}, demonstrating both the broad applicability and expressive power of LLM-driven modeling.
Social system simulation replicates the behaviors of individuals, groups, and institutions to uncover generative mechanisms, forecast trajectories, and inform evidence-based policy~\citep{li2023quantifying}. LLM-based agents are increasingly utilized to model diverse domains, including healthcare~\citep{chopra2025limits,tang2024medagents,li2024agent}, electoral processes~\citep{zhang2024electionsim,touzel2024simulation}, social media~\citep{liu2024skepticism,wang2025decoding,rossetti2024social,piao2025agentsociety}, and economic systems~\citep{li2024econagent,gao2024simulating,yang2025twinmarket}.
% For instance, ~\citet{chopra2025limits} integrated LLMs into large-scale ABMs to simulate the spatiotemporal dynamics of the COVID-19 pandemic in New York City, significantly enhancing the behavioral realism and policy relevance of public health interventions. In economics and finance, ~\citet{li2024econagent} introduced EconAgent, a multi-agent system  framework that captures macroeconomic patterns through micro-level agents endowed with perception, reflective reasoning, and autonomous decision-making capabilities. \citet{gao2024simulating} developed ASFM, a financial market simulator featuring realistic order-matching mechanisms to evaluate the systemic impact of regulatory policies. Similarly, ~\citet{yang2025twinmarket} proposed TwinMarket, a hybrid simulation platform that explicitly models investors’ cognitive biases and cross-platform information diffusion between trading systems and social networks, successfully reproducing hallmark market anomalies and emergent collective behaviors.
Specifically, \citet{chopra2025limits} integrated LLMs into large-scale ABMs to simulate New York City's COVID-19 dynamics, improving the realism of public-health interventions. For economics and finance, \citet{li2024econagent} introduced EconAgent to capture macroeconomic patterns via agents capable of reflective reasoning. Similarly, \citet{gao2024simulating} developed ASFM to evaluate regulatory impacts through order matching, while \citet{yang2025twinmarket} proposed TwinMarket to model cognitive biases and information diffusion between social networks and trading systems.

% However, existing studies on simulating economic systems have primarily focused on broad macroeconomic phenomena, often overlooking more intricate mechanisms, such as interactions with physical environments, and lacking integrated counterfactual reasoning tools essential for in-depth causal analysis.
However, existing studies on economic systems have primarily focused on broad macroeconomic phenomena, often overlooking interactions with physical environments and lacking integrated counterfactual reasoning tools for in-depth causal analysis.

\section{Method}
\begin{figure*}[t]
  \centering
  \includegraphics[width=\textwidth]{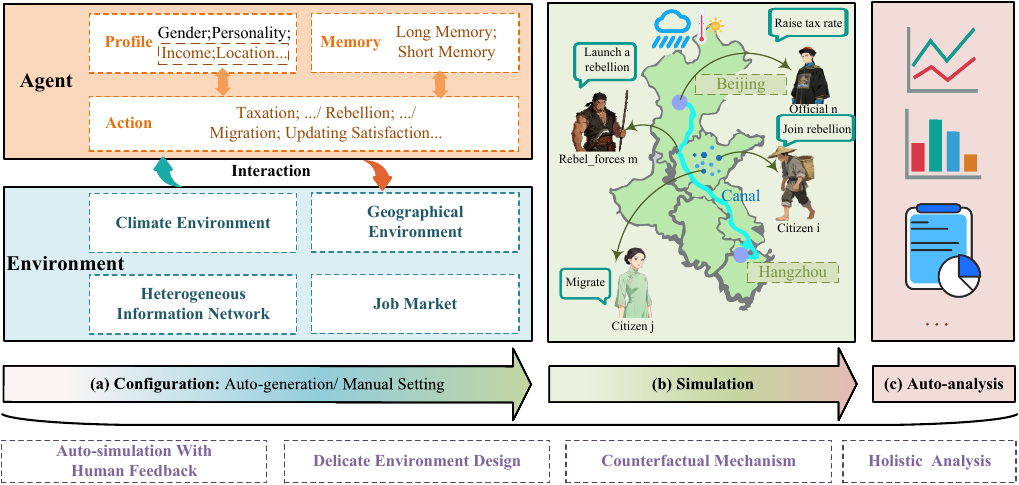}
  \caption{The architecture of the proposed \textit{Eco3S} framework.}
  \label{fig:framework}
  \vspace{-1em}
\end{figure*}

As illustrated in Figure \ref{fig:framework}, the proposed \textit{Eco3S} framework comprises three core modules: configuration, simulation, and auto-analysis. We first present the basic simulation mode, which relies on manual configuration. Building upon this foundation, we introduce the advanced auto-simulation mode augmented with human feedback. 
% A simple demonstration of the system's front-end interface is provided in Appendix \ref{app:system_interface}.

\subsection{Basic Framework}
\textbf{Agent Configuration.}
The configuration module defines the setup for agents, the environment, and their interaction rules. Each agent profile includes static attributes and dynamic states, paired with a dual short- and long-term memory system. To handle LLM context limits, a memory summarization mechanism bounds the context length while preserving key information for future decisions. An agent's action space covers both social interactions and physical tasks, such as migration and infrastructure maintenance.

\textbf{Environment Setting.}
The environment integrates physical, social, and economic subsystems. Climate and geographical variables produce spatially heterogeneous distributions of resources and infrastructure. Social dynamics are modeled via a dynamic Heterogeneous Information Network (HIN) that maps pairwise relationships and group memberships to facilitate context-aware communication. Finally, local labor markets link government investment and macroeconomic conditions to employment opportunities.

\textbf{Co-evolving Environment Design.}
Unlike frameworks in which the environment serves as a static background, \textit{Eco3S} models bidirectional feedback between agents and an evolving environment. Environmental states change through exogenous processes and internal transition rules; agents perceive these states and make context-sensitive decisions based on their profiles and memories; aggregated actions then reshape physical, social, and economic conditions. For example, infrastructure investment affects canal navigability and employment, while migration changes the spatial distribution of population and labor. The framework supports both individual decisions and group decisions produced through multi-round deliberation, enabling micro-level interactions to generate emergent macro-level dynamics.

Formally, the coupled agent-environment transition at each simulation step is defined as follows:
{\small
\begin{equation}
\begin{gathered}
s_{a_i}^{t+1}
  = \mathcal{F}_{a_i}(s_{a_i}^t,\bar{S}_{a_i}^t,S_e^t;\mathcal{P}),\quad
s_{e_k}^{t+1}
  = \mathcal{F}_{e_k}(s_{e_k}^t,\bar{S}_{e_k}^t,S_a^t;\mathcal{P}),\\[-0.2ex]
Y_{\text{outcome}}^{t+1}
  = \mathcal{G}(S_a^{t+1},S_e^{t+1}),
\end{gathered}
\label{eq:state_transition}
\end{equation}
}
% \begin{equation}
% \begin{aligned}
% s_{a_i}^{t+1}
%   &= \mathcal{F}_{a_i}(s_{a_i}^t,\bar{S}_{a_i}^t,S_e^t;\mathcal{P}),\\
% s_{e_k}^{t+1}
%   &= \mathcal{F}_{e_k}(s_{e_k}^t,\bar{S}_{e_k}^t,S_a^t;\mathcal{P}),\\
% Y_{\text{outcome}}^{t+1}
%   &= \mathcal{G}(S_a^{t+1},S_e^{t+1}),
% \end{aligned}
% \label{eq:state_transition}
% \end{equation}
where $S_a^t=\{s_{a_0}^t,s_{a_1}^t,\ldots,s_{a_n}^t\}$ and $S_e^t=\{s_{e_0}^t,s_{e_1}^t,\ldots,s_{e_m}^t\}$ denote the joint states of all agents and environmental components at time $t$. For agent $i$, the set $\bar{S}_{a_i}^t=\{s_{a_j}^t\}_{j\neq i}$ captures the states of peers that influence its decisions via the interaction network; similarly, $\bar{S}_{e_k}^t=\{s_{e_\ell}^t\}_{\ell\neq k}$ represents the environmental context relevant to component $k$. The transition functions $\mathcal{F}_{a_i}$ and $\mathcal{F}_{e_k}$ characterize agent decision-making and environmental evolution, respectively, under the simulation configuration $\mathcal{P}$. Finally, $\mathcal{G}$ aggregates these updated micro-level states into the macro-level outcome $Y_{\text{outcome}}^{t+1}$, explicitly modeling the bidirectional dependence between agent behavior and environmental dynamics.

\textbf{Simulation.}
% The framework can execute the simulation with a specified number of repetitions to ensure statistical reliability. During each simulation run, all agents make decisions simultaneously. Concurrently, the environment evolves either according to exogenous, time-dependent sequences or endogenously in response to agent–environment interactions. In contrast to prior approaches, \textit{Eco3S} incorporates a \textbf{Structural Causal Simulation (SCS)} mechanism into the framework. Inspired by SCM, SCS treats the generative behaviors of agents and environment dynamics as implicit structural equations. The system automatically preserves checkpoints of agent states and environmental conditions, allowing users to perform causal interventions analogous to the $do$-operator. Users can pause the simulation and resume from a saved checkpoint under modified conditions, such as environmental parameters, policy rules, or agent prompts. By replaying trajectories from identical initial states under divergent interventions, \textit{Eco3S} supports rigorous counterfactual analysis: it isolates the causal effect of specific changes on system outcomes without relying on explicit mathematical formulations.This capability is particularly significant for economic research, where counterfactual inference lies at the heart of policy evaluation, mechanism design, and causal understanding of complex socio-economic dynamics.
During each simulation step, agents make decisions while environmental components evolve exogenously or in response to agent actions. Multiple independent runs can be executed for statistical evaluation. \textit{Eco3S} further incorporates \textbf{Structural Causal Simulation (SCS)}, which preserves checkpoints of agent and environment states and supports interventions analogous to the $do$-operator. Researchers can resume from the same checkpoint under modified environmental parameters, policies, or agent prompts, producing comparable trajectories that isolate the effects of alternative interventions.

\textbf{Auto-analysis.}
Finally, the auto-analysis module automatically interprets simulation results in light of the initial configuration, generating tailored visualizations (e.g., charts and tables), statistical summaries, and narrative reports based on the observed outcomes.

% Finally, the auto-analysis module automatically interprets simulation results in light of the initial configuration, generating tailored visualizations (e.g., charts and tables), statistical summaries, and narrative reports based on the observed outcomes, as defined in the following:
% {\small
% \begin{equation}
% Y_{\text{outcome}}^{t+1} = \textit{Eco3S}(S_a^t, S_e^t; \mathcal{P});\quad
% s_{a_i}^{t+1} = \textit{Eco3S}(s_{a_i}^t, \bar{S}_{a_i}^t, S_e^t; \mathcal{P});\quad
% s_{e_i}^{t+1} = \textit{Eco3S}(s_{e_i}^t, \bar{S}_{e_i}^t, S_a^t; \mathcal{P})
% \end{equation}
% }

% where  \( S_a^t = \{s_{a_0}^t, s_{a_1}^t, \dots, s_{a_n}^t\} \) and \( S_e^t = \{s_{e_0}^t, s_{e_1}^t, \dots, s_{e_m}^t\} \) denote the states of all agents and environmental components at time \( t \), respectively;
% \( \bar{S}_{a_i}^t = \{s_{a_j}^t\}_{j \neq i} \) and \( \bar{S}_{e_i}^t = \{s_{e_j}^t\}_{j \neq i} \) represent the sets of all other agent and environmental states excluding the \(i\)-th element;  
%  \( \cal{P} \) denotes the simulation hyperparameters, such as the number of agents and simulation steps; and  
% \( Y_{\text{outcome}}^{t+1} \) is the macro-level simulation outcome at time \( t+1 \).
% This module delivers a clear, intuitive, and comprehensive interface for researchers to explore, compare, and communicate findings. 

\subsection{Auto-simulation}
\label{subs:auto-simulation}
\begin{figure}[!ht]
\vspace{-1em}
  \centering
  \includegraphics[width=\columnwidth]{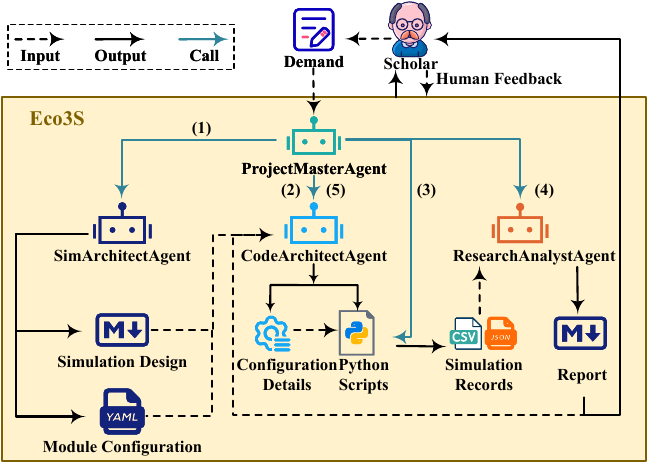}
  \caption{The architecture of the proposed auto-simulation with human feedback.}
  \label{fig:auto-simulation}
  \vspace{-1em}
\end{figure}

% To enable end-to-end automation of socio-economic simulations from natural language, we introduce the Simulation-Analysis-Refinement (SAR) paradigm. Inspired by human research process, SAR dynamically orchestrates specialized LLM agents across the entire simulation life-cycle. As illustrated in Figure~\ref{fig:auto-simulation}, the system comprises four core agents: ProjectMasterAgent, SimArchitectAgent, CodeArchitectAgent, and ResearchAnalystAgent. These agents operate in a collaborative, iterative loop, incorporating human feedback when necessary, to jointly design, implement, execute, and refine experiments based on high-level natural language inputs. 
We introduce the Simulation-Analysis-Refinement (SAR) paradigm to enable the automated synthesis of socio-economic simulations from natural-language requests. SAR orchestrates four specialized LLM agents, comprising the ProjectMasterAgent, SimArchitectAgent, CodeArchitectAgent, and ResearchAnalystAgent, to manage the iterative research life cycle. As illustrated in Figure~\ref{fig:auto-simulation}, these agents collaboratively design, implement, and refine experiments while integrating human-in-the-loop feedback.

\textbf{Demand Analysis.}
% The workflow begins when a researcher submits a natural language simulation request. The ProjectMasterAgent, serving as the central orchestrator, parses this unstructured input into a structured requirement. The SimArchitectAgent then interprets the user's intent using system documentation and produces a detailed, formally structured simulation design document. Based on this foundational blueprint, it further generates a high-level module configuration that specifies which functional subsystems should be activated, along with their interconnection logic.
The workflow commences with a researcher's natural-language request. As the central orchestrator, the ProjectMasterAgent formalizes this unstructured input into explicit research requirements. The SimArchitectAgent subsequently interprets these requirements via system documentation to generate a structured simulation design and a module configuration defining the functional subsystems and their connectivity.

\textbf{Configuration Generation.}
% Subsequently, the CodeArchitectAgent, functioning as a full-stack software engineer, consumes these blueprints to automatically generate executable Python code. It synthesizes simulator scripts, runtime configuration files, and agent-specific prompt templates. Critically, this agent employs an incremental refinement strategy: it first populates code from templates, then iteratively enhances functionality module-by-module using context-aware LLM calls, and finally validates coherence through self-check loops.
The CodeArchitectAgent implements the design as executable Python artifacts, comprising simulator scripts, runtime configurations, and agent-specific prompt templates. It instantiates reusable templates and subsequently integrates scenario-specific functions via context-aware generation. Automated self-checks ensure consistency across the generated code, configurations, and prompts prior to execution.

\textbf{Simulation Execution.}
% Once the simulation artifacts are ready, the ProjectMasterAgent executes the generated program. If runtime errors occur, it captures the stack trace and instructs the CodeArchitectAgent to perform automatic debugging, modifying code or configurations until the simulation runs successfully. Upon successful execution, the system produces structured outputs.
Upon completion, the ProjectMasterAgent executes the program. If execution fails, it captures the stack trace for the CodeArchitectAgent to repair the code or configuration prior to re-execution. Successful runs yield structured simulation outputs for subsequent analysis.

\textbf{Results Analysis and Configuration Optimization.} 
% The ResearchAnalystAgent evaluates simulation outcomes against the Key Performance Indicators (KPIs) and qualitative trends defined in the design document. If the results deviate from expectations, it generates a diagnostic report pinpointing problematic configurations or agent behaviors. Guided by this diagnosis, the ProjectMasterAgent triggers an optimization loop: it directs the CodeArchitectAgent to adjust relevant configurations or prompts and initiates a new simulation run. This autonomous Simulation-Analysis-Refinement (SAR) cycle repeats until the outcomes meet predefined quality criteria or reach a maximum iteration limit.
% To ensure practical convergence, the framework incorporates two key designs. First, a hard iteration limit (typically 10) prevents infinite loops, with our experiments showing successful optimization usually within five cycles. Second, it adopts a \textit{satisficing} principle. Since finding a strict mathematical optimum is often unrealistic for complex socio-economic systems, the framework focuses on qualitative plausibility, requiring only that the simulation exhibits an overall trend that effectively addresses the research question. This calibrated baseline then serves as the foundation for Structural Causal Simulation, where users can intervene on the replicated setting for counterfactual analysis.
The ResearchAnalystAgent evaluates simulation outcomes against the KPIs and qualitative trends in the design document. If deviations occur, it produces a diagnostic report pinpointing problematic configurations or agent behaviors. The ProjectMasterAgent then instructs the CodeArchitectAgent to adjust the relevant parameters or logic before triggering a new iteration. Rather than seeking a strict mathematical optimum, SAR employs a diagnostic-guided \textit{satisficing} principle; the process terminates once the results exhibit plausible trends or reach the iteration limit, typically 10. Empirically, optimization usually succeeds within five cycles. The resulting calibrated simulation serves as a baseline for subsequent counterfactual interventions.

\textbf{Human Feedback Control.}
% Throughout this process, human scholars may intervene at key steps (e.g., after design or initial code generation) to provide feedback. Moreover, at any stage, users can leverage \textit{Eco3S}’s counterfactual mechanism to modify parameters and restart from saved checkpoints, facilitating policy exploration. By integrating structured decomposition, autonomous code synthesis, and data-driven iteration, our agent system realizes a closed-loop pipeline that transforms abstract research questions into validated, interpretable economic simulations, dramatically lowering the technical barrier for social science researchers.
Researchers can intervene at critical stages, such as post-design or following initial code generation, to refine requirements and configurations. In later phases, they may adjust parameters or policies and resume execution from saved checkpoints. This human-in-the-loop architecture integrates automated decomposition, code synthesis, and outcome-driven refinement while ensuring researcher control over experimental assumptions and policy exploration.

\section{Experiments}
We evaluate \textit{Eco3S} across three established social-science settings: \textit{Canal Decay and Rebellion} investigates infrastructure-driven stability; \textit{Origins of Governance} examines how geography shapes collective decision-making; and \textit{Information Propagation} replicates a field experiment on dissemination strategies. The former two cases leverage the full agent-environment architecture, while the latter isolates cognitive belief updating by omitting the physical environment. To ensure statistical robustness, each scenario is executed over five independent runs.

\subsection{Canal Decay and Rebellion}
\textbf{Simulation Mechanism.}
% The simulation is grounded in the historical collapse of the Grand Canal in China, which forced the Qing government to shift grain transport to maritime routes. This trade disruption provides a quasi-natural experiment to study social stability. Using the empirical finding of \citet{cao2022rebel} as a benchmark, our simulated outcome deviates from their reported effect by only 8 percentage points.
% The simulation is governed by five core mechanisms:
% (1) Canal navigability varies annually in response to exogenous climate conditions and endogenous government maintenance.
% (2) As navigability declines, transportation costs along the canal increase. 
% (3) Government agents dynamically allocate their limited budget across three competing priorities (i.e., canal maintenance, military expenditures, and public services), and concurrently adjust both tax rates and the strategic mix between canal- and sea-based tribute transportation.
% (4) Resident agents make adaptive decisions regarding employment and migration. 
% (5) Rebel agents, operating under budget constraints, decide whether to initiate rebellion, select target towns, and disseminate propaganda to influence resident sentiment and recruitment. 
This experiment simulates the Grand Canal's decline and the Qing government's pivot to maritime transport, providing a historical setting to analyze the impact of trade disruption on social stability. Each year proceeds via five linked mechanisms: (1) navigability evolves with climate conditions and maintenance; (2) declining navigability increases transport costs; (3) government agents allocate budgets across maintenance, defense, and public services while adjusting taxes and transport modes; (4) residents adapt employment and migration to local conditions; and (5) rebels determine mobilization and propaganda strategies. These processes form a feedback loop where infrastructure decay alters labor and satisfaction, impacting rebellion and subsequent policy. Benchmarked against \citet{cao2022rebel}, the simulated rebellion risk deviates from the reported effect by only 8 percentage points.

\begin{figure}[!ht]
  \centering
  \begin{tabular}{@{}c@{\hspace{0.01\columnwidth}}c@{}}
    \includegraphics[width=0.48\columnwidth]{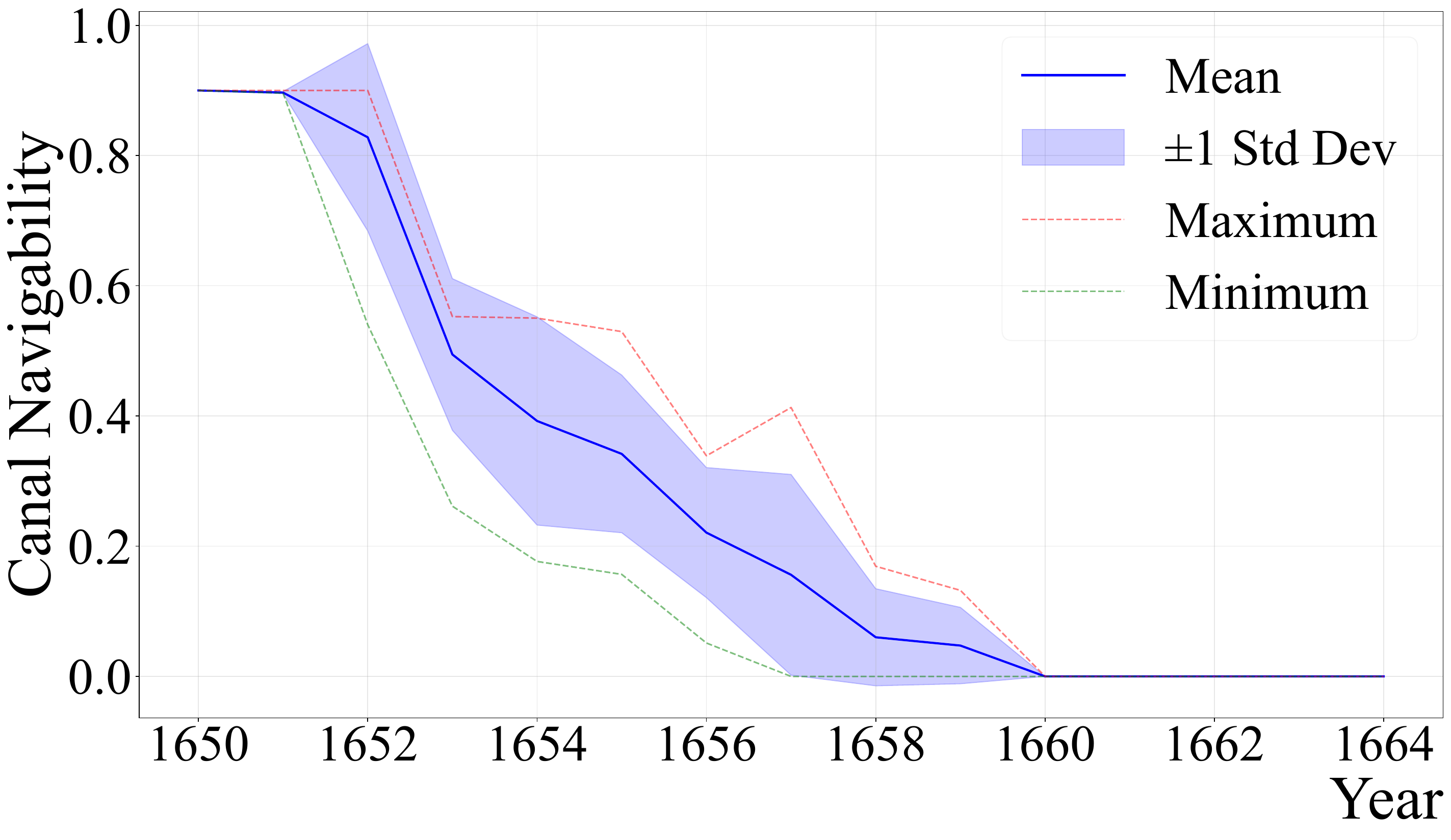} &
    \includegraphics[width=0.48\columnwidth]{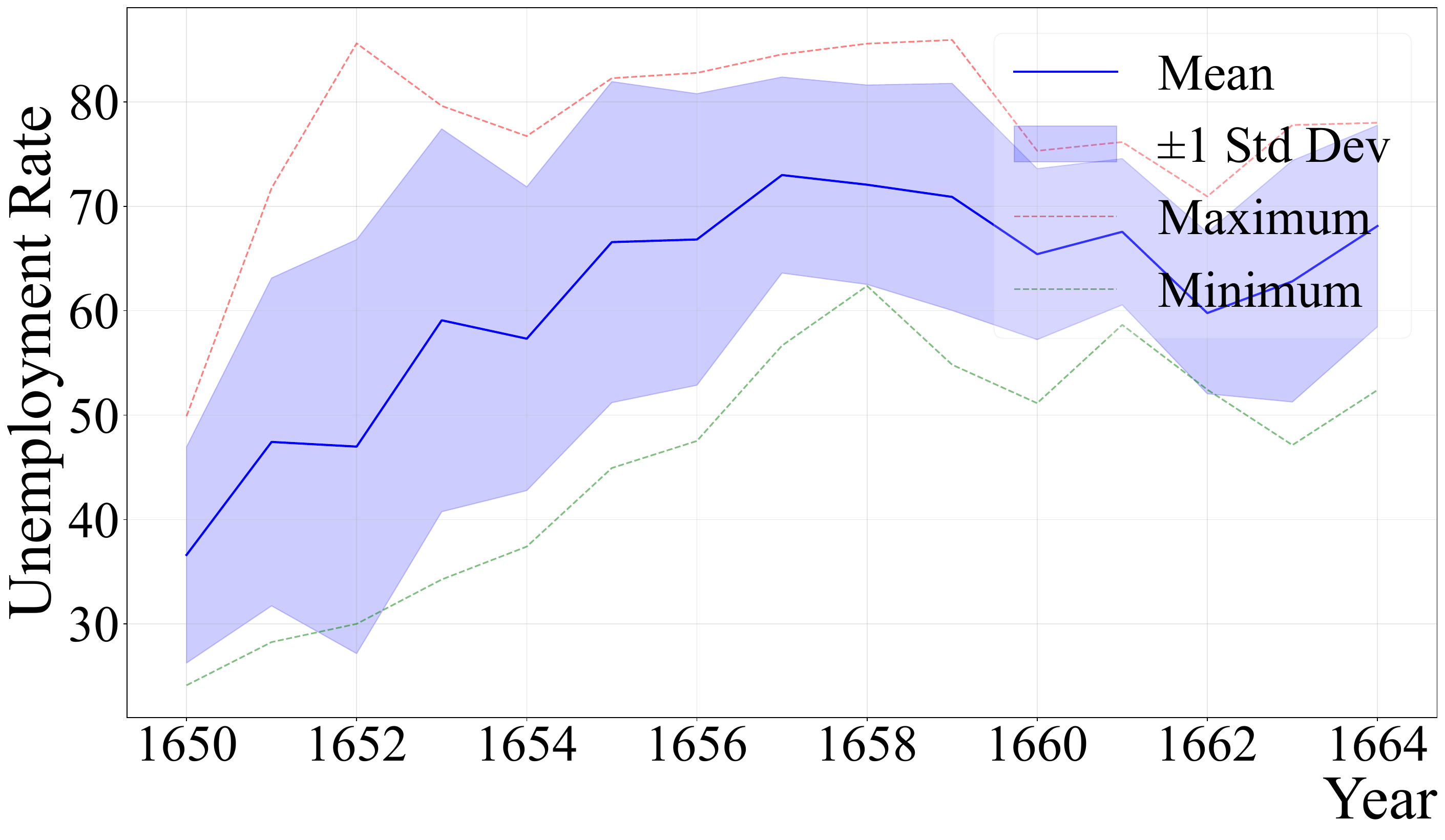} \\[-2ex]
    {\scriptsize (a) Canal Navigability} &
    {\scriptsize (b) Unemployment Rate} \\[0.8ex]
    \includegraphics[width=0.48\columnwidth]{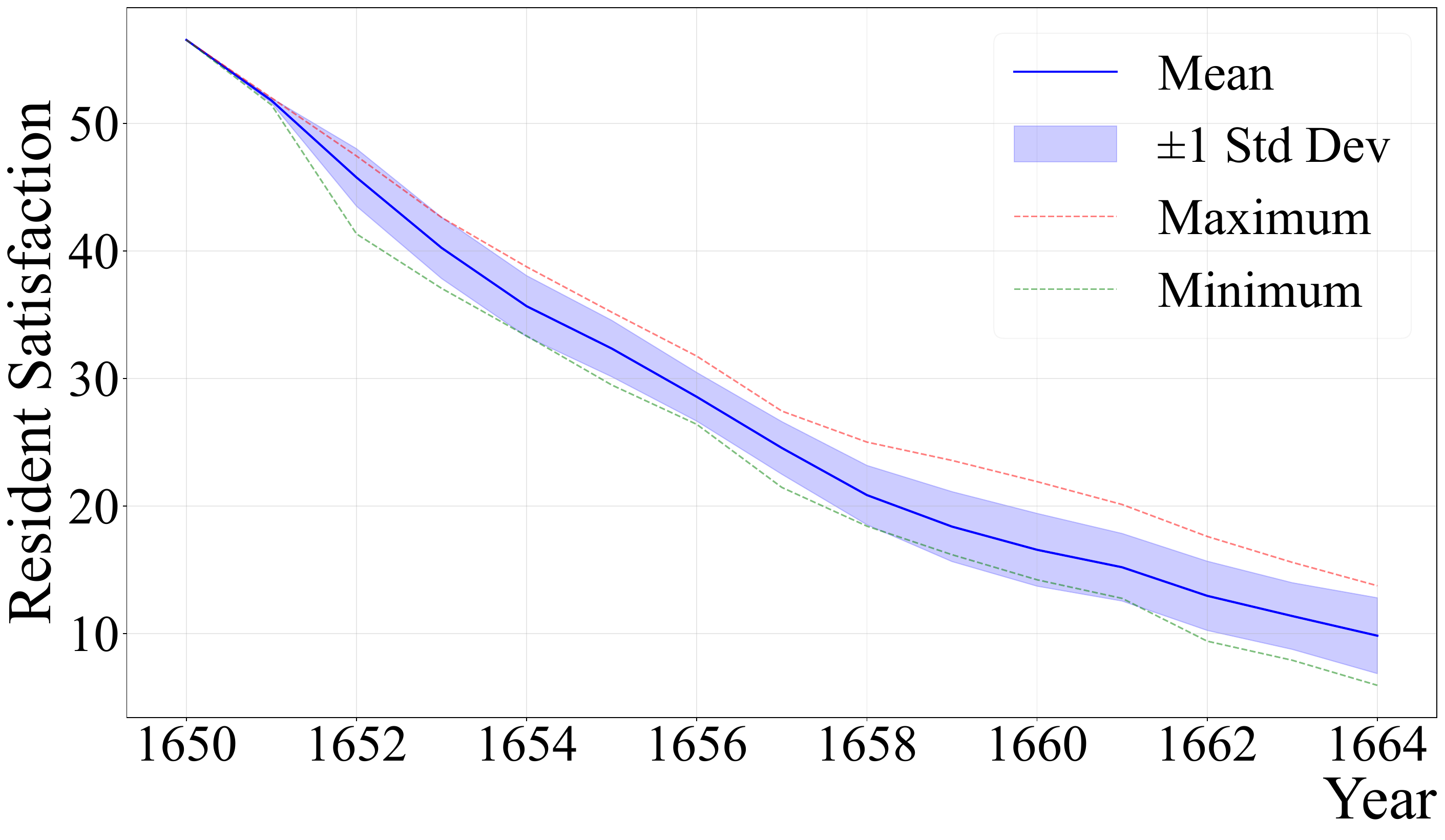} &
    \includegraphics[width=0.48\columnwidth]{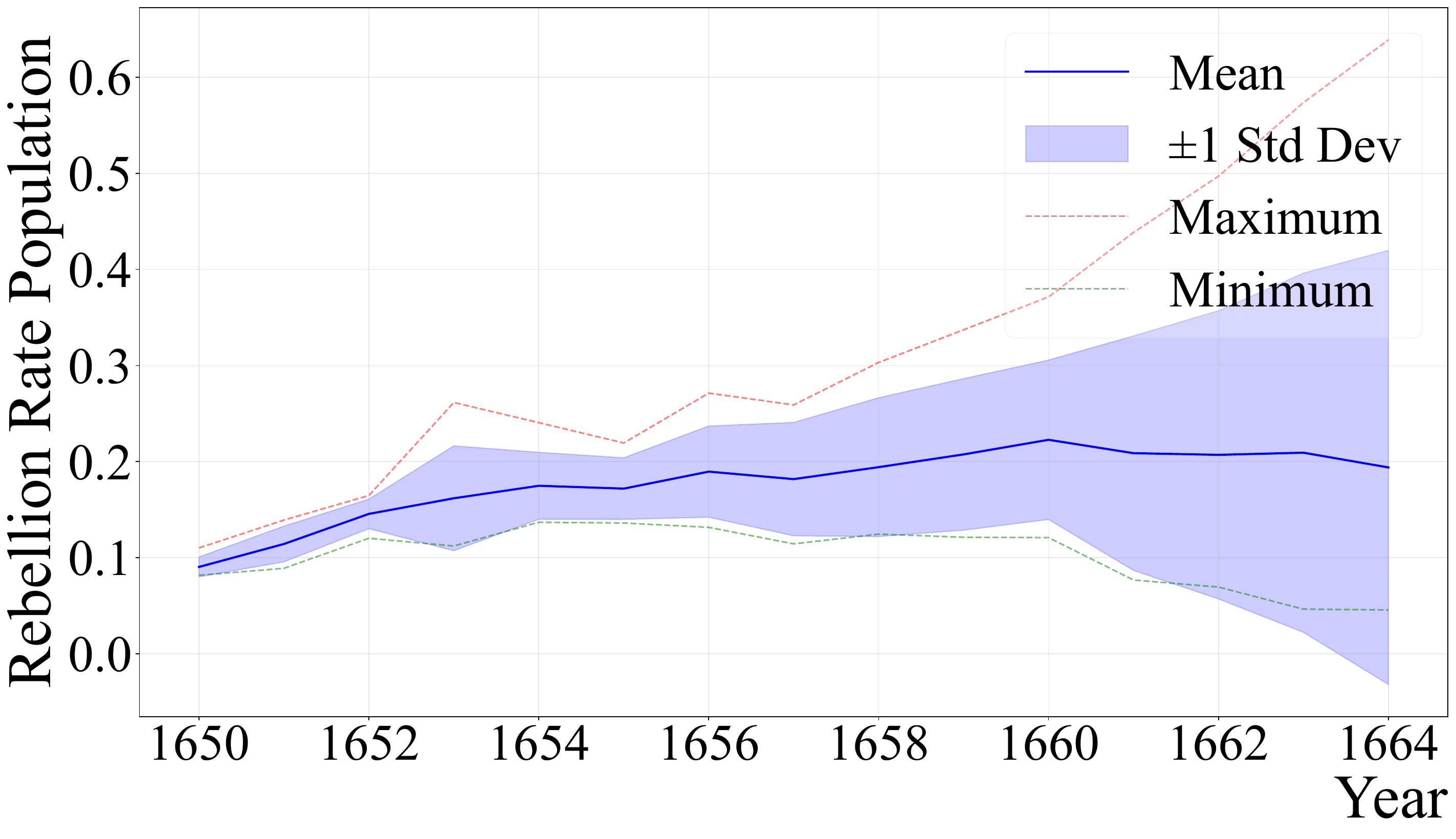} \\[-2ex]
    {\scriptsize (c) Reisdent Satisfaction} &
    {\scriptsize (d) Rebellion Population Rate}
  \end{tabular}
  \caption{Qualitative analysis of the Canal Decay experiment.}
  \label{fig:canal_decay_qualitative_analysis}
  \vspace{-1em}
\end{figure}

\textbf{Overall Dynamics.}
Figure~\ref{fig:canal_decay_qualitative_analysis} illustrates the system-level evolution. As canal navigability declines, the simulation observes rising unemployment and declining satisfaction, which subsequently drive an increase in the rebellion population. This progression captures the transmission from infrastructure decay to social instability.

\begin{figure}[!ht]
  \centering
  \includegraphics[width=0.49\columnwidth]{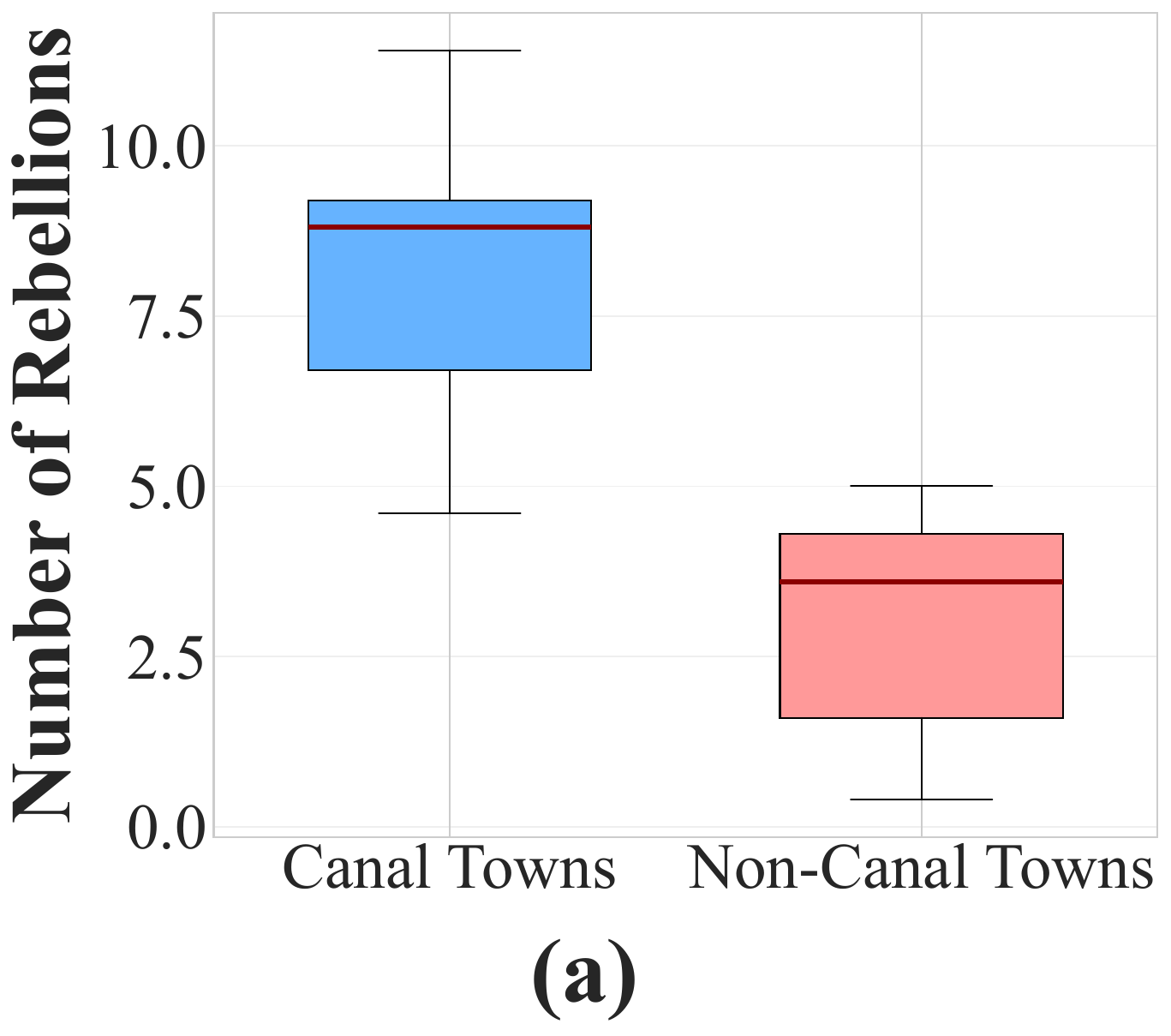}
  \includegraphics[width=0.49\columnwidth]{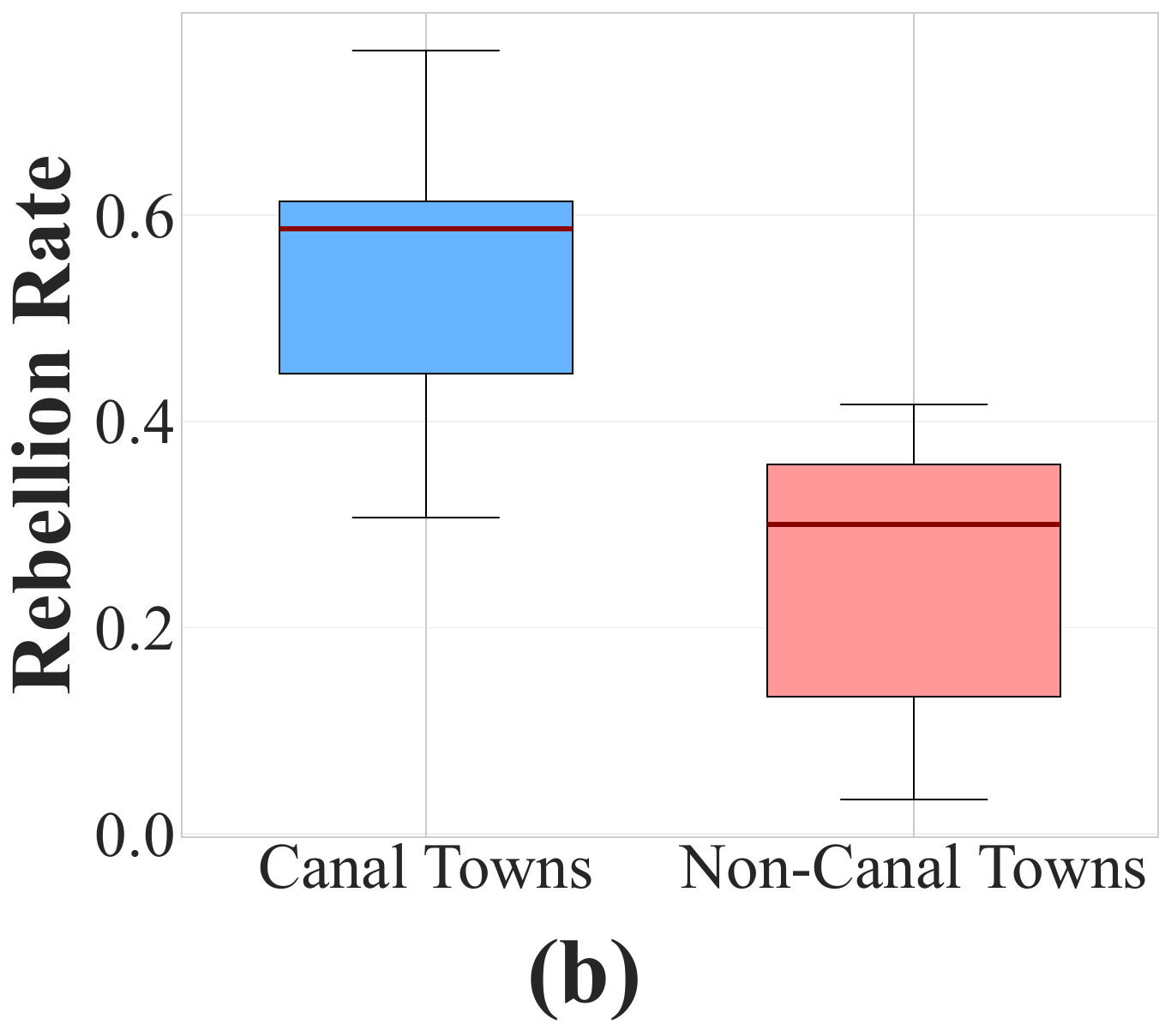}
  \vspace{-1em}
  \caption{Comparison of Spatial Distribution of Rebellion.}
  \label{fig:rebellion_boxplot}
  
\end{figure}

\textbf{Spatial Pattern.}
Table~\ref{tab:canal_decay_spatial_analysis} and Figure~\ref{fig:rebellion_boxplot} indicate that canal-adjacent towns exhibit a significantly higher rebellion rate than non-canal towns ($0.54\pm0.12$ vs.\ $0.24\pm0.13$; $p<0.05$, Cohen's $d=2.40$). This 125\% increase aligns with the 117\% empirical benchmark of \citet{cao2022rebel}, confirming that \textit{Eco3S} reproduces the spatial concentration of unrest observed historically.

% \begin{table}[!ht]
% \centering
% \footnotesize
% \setlength{\tabcolsep}{2.5pt}
% \renewcommand{\arraystretch}{0.9}
% \begin{threeparttable}
% \begin{tabular}{@{}lccc@{}}
% \toprule
% & Number & \#Rebellion & Rebellion Rate \\
% \midrule
% Canal     & 15 & $8.17 \pm 1.83$ & $0.54 \pm 0.12$ \\
% Non-canal & 12 & $2.92 \pm 1.55$ & $0.24 \pm 0.13$ \\
% Cohen's $d$ & -- & 3.09 & 2.40 \\
% $P$ value   & -- & $<0.01$ & $<0.05$ \\
% \bottomrule
% \end{tabular}
% \end{threeparttable}

% \vspace{1pt}
% \makebox[\columnwidth][c]{%
%   \footnotesize For Cohen's $d$, $|d|\geq0.8$ denotes a statistically large effect.%
% }
% \caption{Spatial Analysis of Rebellion Rate}
% \label{tab:canal_decay_spatial_analysis}
% \vspace{-1em}
% \end{table}

\begin{table}[!ht]
\centering
{\small
\setlength{\tabcolsep}{3pt}
\begin{tabular}{@{}cccc@{}}
\toprule
                & \multicolumn{1}{l}{Number} & \multicolumn{1}{l}{\#Rebellion } & \multicolumn{1}{l}{Rebellion Rate} \\ \midrule
Canal     & 15                         & $8.17 \pm 1.83$                      & $0.54 \pm 0.12$                \\
Non-canal & 12                         & $2.92 \pm 1.55$                      & $0.24 \pm 0.13$                \\
Cohen's d      & -                          & 3.09                              & 2.40                           \\
P value         & -                          & \textless{}0.01                     & \textless{}0.05   \\ \bottomrule                
\end{tabular}
}
\caption{Spatial analysis of rebellion rates; $|d|\geq0.8$ denotes a large effect.}
\label{tab:canal_decay_spatial_analysis}
\end{table}

% \textbf{Temporal Heterogeneity.}
% We conduct a temporal analysis of the rebellion rate by regressing it on simulation year using standard ordinary least squares (OLS) model:
% \begin{equation}
% Rebellion\_rate = \alpha \times Year + \beta + \epsilon.
% \end{equation}
% The overall progression of conflict is depicted in Figure \ref{fig:rebellion_temporal_bar}, which shows a consistent upward trend in the total number of rebellions across both types of towns as canal functionality deteriorates.
% It should be noted that, due to temporal heterogeneity in the rebellion rate, the OLS model may not be fully appropriate for this spatial analysis. Nevertheless, our focus lies on identifying statistically significant trends rather than establishing precise causal relationships.
% As shown in Table~\ref{tab:canal_decay_temporal_analysis}, rebellion rates in both adjacent and non-adjacent towns increase significantly with simulation year. This trend is primarily driven by the deteriorating condition of the canal and the consequent rise in unemployment, as illustrated in Figure~\ref{fig:canal_decay_qualitative_analysis}(a). Moreover, the intercept for adjacent towns is higher than that for non-adjacent towns, reflecting spatial heterogeneity in rebellion propensity. And the slopes $\alpha$ are nearly identical, suggesting similar temporal trends across locations.
% These findings quantitatively corroborate historical evidence linking transportation infrastructure collapse to heightened conflict, as documented by ~\citet{cao2022rebel}.

\begin{figure}[!ht]
  \centering
  \includegraphics[width=0.9\columnwidth]{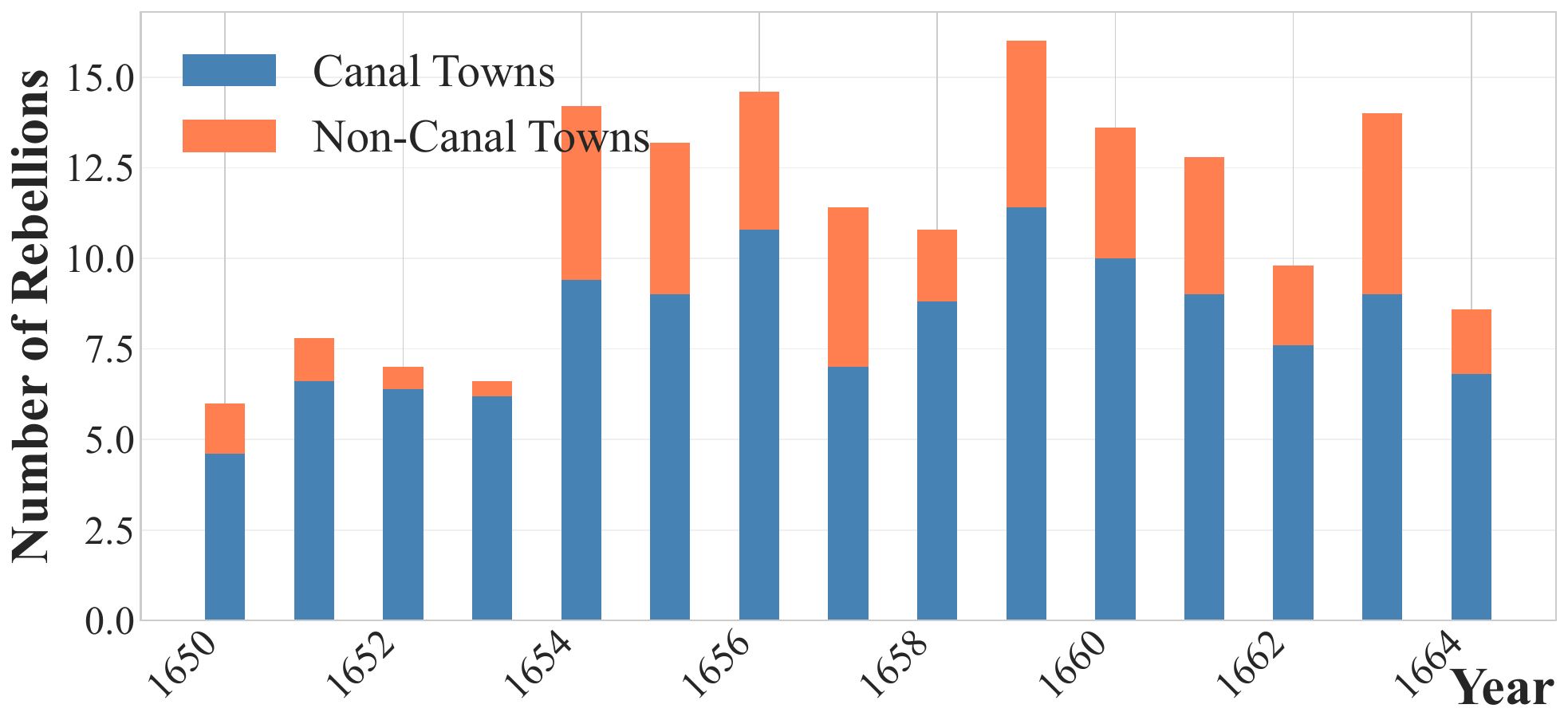}
  \vspace{-1em}
  \caption{Stacked Distribution of Rebellions.}
  \label{fig:rebellion_temporal_bar}
  \vspace{-1em}
\end{figure}

\textbf{Temporal Trend.}
We employ an OLS model to characterize the temporal trend:
\begin{equation}
\mathrm{RebellionRate}_t=\alpha,\mathrm{Year}_t+\beta+\epsilon_t.
\end{equation}
As shown in Table~\ref{tab:canal_decay_temporal_analysis}, rebellion rates increase significantly for both town types. Their near-identical slopes suggest comparable temporal dynamics, while the higher intercept for canal-adjacent towns reflects a persistent spatial disparity in rebellion propensity. Figure~\ref{fig:rebellion_temporal_bar} further confirms the rising conflict frequency as canal functionality collapses. Collectively, these results replicate the temporal link between transportation failure and conflict documented by \citet{cao2022rebel}.

% \begin{table}[!ht]
% \centering
% \begin{threeparttable}
% \begin{tabular}{ccc}
% \toprule
% Coefficient           & Canal & Non-canal \\ \midrule
% $\alpha$ & 0.0123***      & 0.0127***          \\
% $\beta$  & 0.4589      & 0.1546          \\
% \#Samples        & 225         & 180      \\ \bottomrule       
% \end{tabular}

% \end{threeparttable}
% \makebox[\columnwidth][c]{%
%   \footnotesize *** indicates the coefficient is significant at the 99\% level.%
% }
% \caption{Temporal Analysis of Rebellion Rate}
% \label{tab:canal_decay_temporal_analysis}
% \vspace{-1em}
% \end{table}

\begin{table}[!ht]
\centering
\begin{tabular}{lcc}
\toprule
Term & Canal & Non-canal \\
\midrule
$\alpha$ & $0.0123^{***}$ & $0.0127^{***}$ \\
$\beta$  & 0.4589 & 0.1546 \\
\#Samples & 225 & 180 \\
\bottomrule
\end{tabular}
\caption{OLS trends in rebellion rates ($^{***}p<0.01$).}
\label{tab:canal_decay_temporal_analysis}
\end{table}

\subsection{Origins of Governance}

Drawing on \citet{allen2023economic}, we investigate the cooperative theory of state formation. Specifically, shifting river patterns can render private irrigation unsustainable, thereby driving the demand for an authority to coordinate public canal infrastructure.

\textit{Eco3S} implements this by modeling a feedback loop between the river system, fiscal policy, and resident choices. Exogenous climate conditions influence river navigability and agricultural yields. Urban income, which is tied to navigability, drives GDP growth and generates the tax revenue necessary for the government budget. Government agents then allocate these funds between general public expenditure and river maintenance while adjusting the tax rate. These policy decisions directly influence urban employment opportunities and long-term navigability. Residents evaluate their income, tax burden, and local climate to choose between state affiliation as urban residents, self-sufficient farming, or migration. These choices collectively update the polity's population and fiscal capacity in subsequent periods.

\begin{figure}[!ht]
  \centering
  \begin{tabular}{@{}c@{\hspace{0.01\columnwidth}}c@{}}
    \includegraphics[width=0.48\columnwidth]{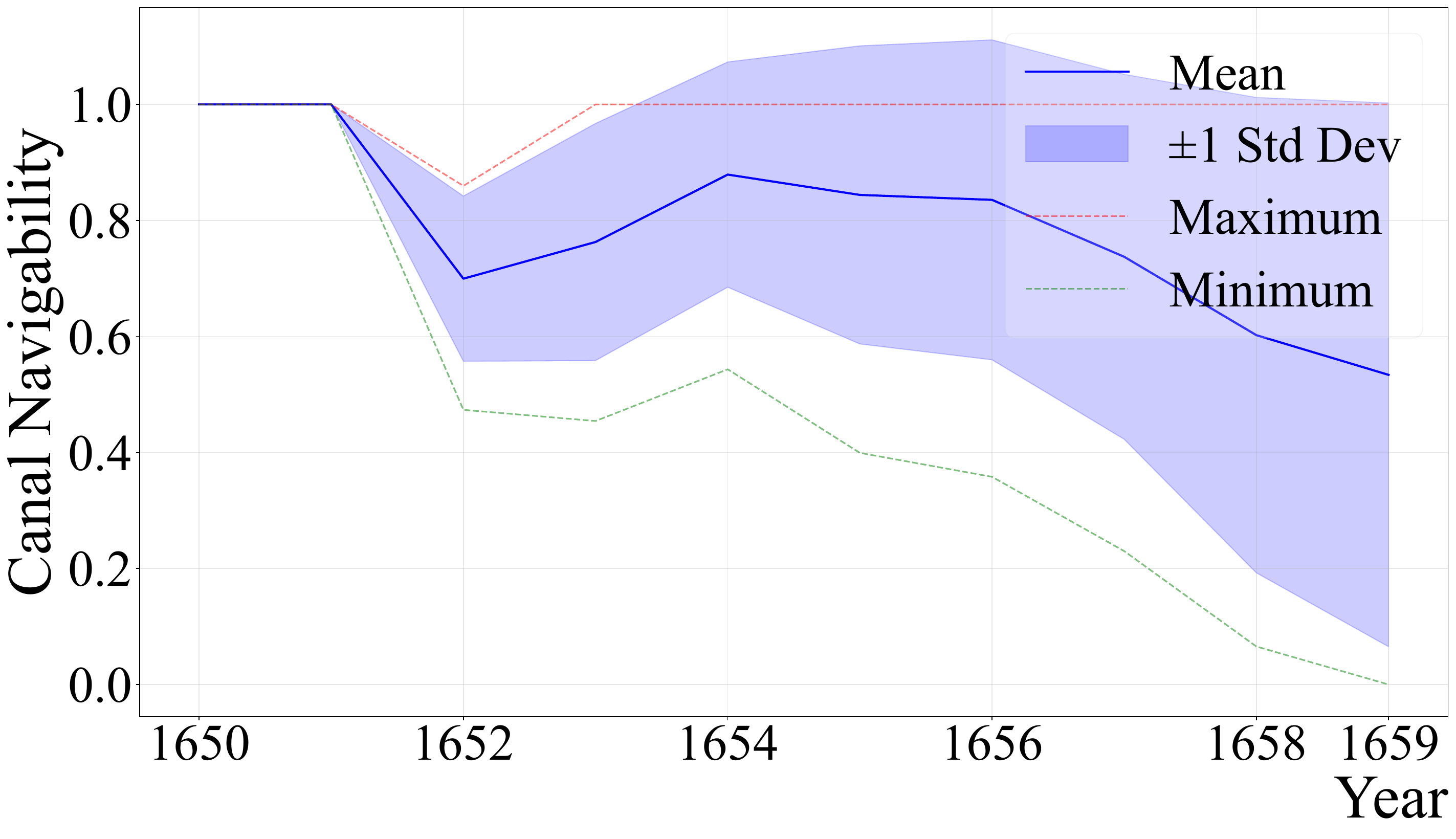} &
    \includegraphics[width=0.48\columnwidth]{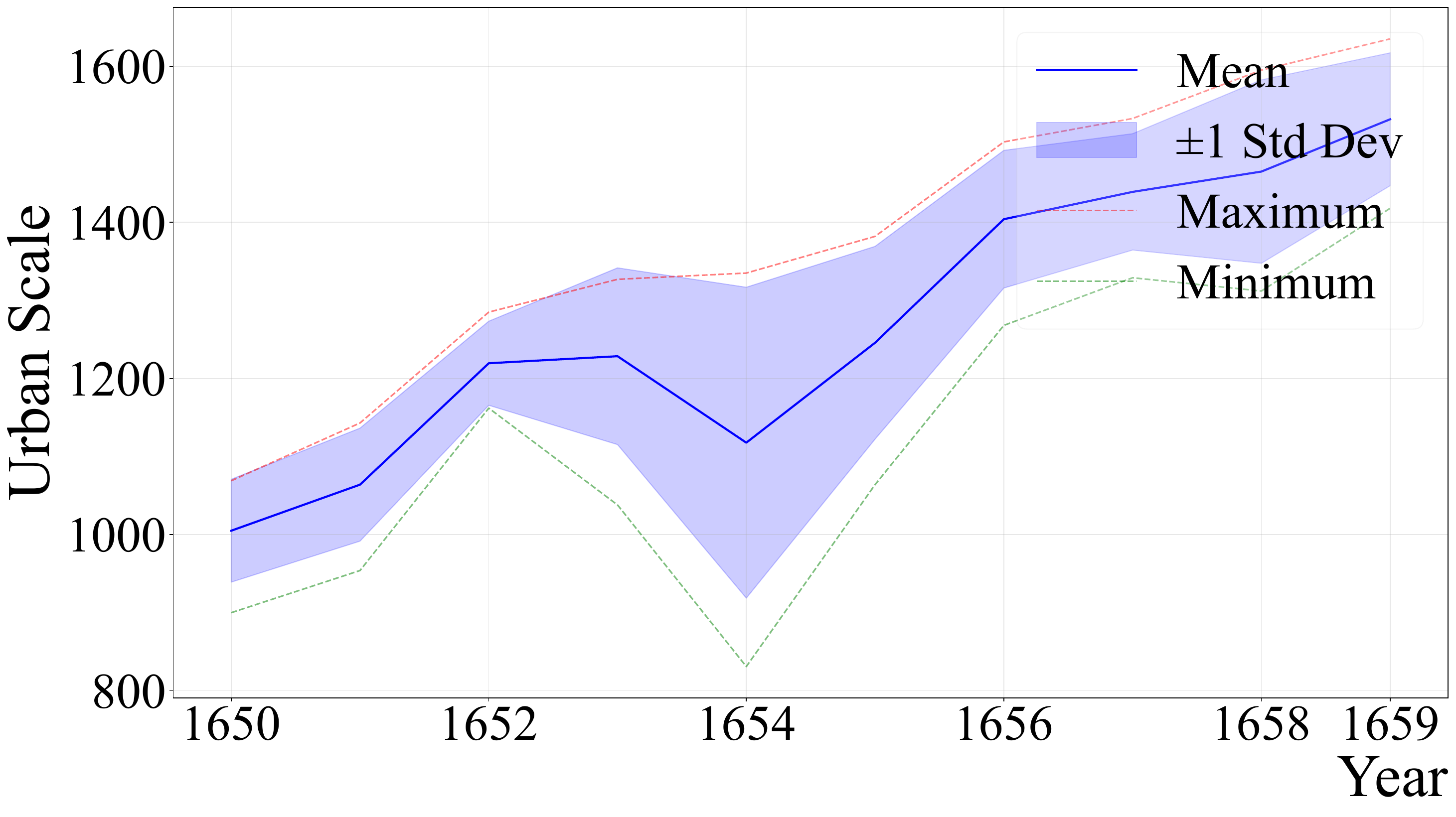} \\[-2ex]
    {\scriptsize (a) River Navigability} &
    {\scriptsize (b) Urban Population}
  \end{tabular}
  \caption{Results of the Origins of Governance experiment.}
  \label{fig:Origins_of_Governance_qualitative_analysis}
  \vspace{-1em}
\end{figure}

As illustrated in Figure~\ref{fig:Origins_of_Governance_qualitative_analysis}, coordinated maintenance preserves navigability despite natural degradation, while the urban population exhibits steady growth. This trend suggests that residents opt for state affiliation when collective governance offers reliable infrastructure and economic stability. The result aligns with \citet{allen2023economic} by confirming that shared infrastructure needs can incentivize the emergence of collective governance by making it an economically rational choice.

\subsection{Information Propagation}

We replicate the $2\times2$ factorial design of \citet{banerjee2024less} by varying message reach (Broadcasting vs.\ Seeding) and meta-knowledge (Common Knowledge vs.\ No Common Knowledge). Agents receive policy information via one of four strategies; they then decide whether to initiate discussions based on personality traits and information certainty. Post-simulation metrics include conversation volume, factual knowledge retention, and the quality of rational choices.

\begin{figure}[t]
  \centering
  \begin{tabular}{
    @{}c@{\hspace{0.01\columnwidth}}
    c@{\hspace{0.01\columnwidth}}c@{}
  }
    \includegraphics[width=0.32\columnwidth]
      {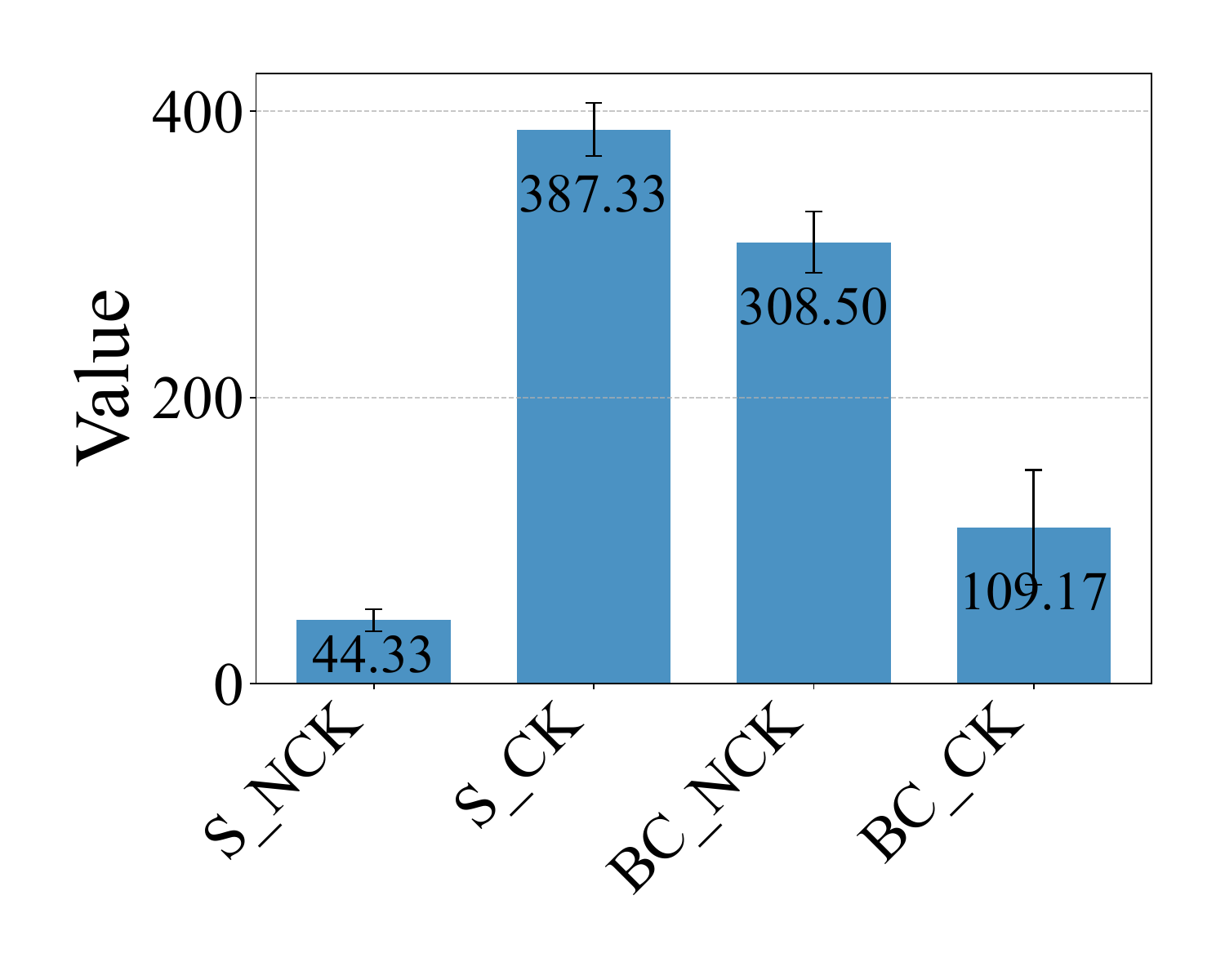} &
    \includegraphics[width=0.32\columnwidth]
      {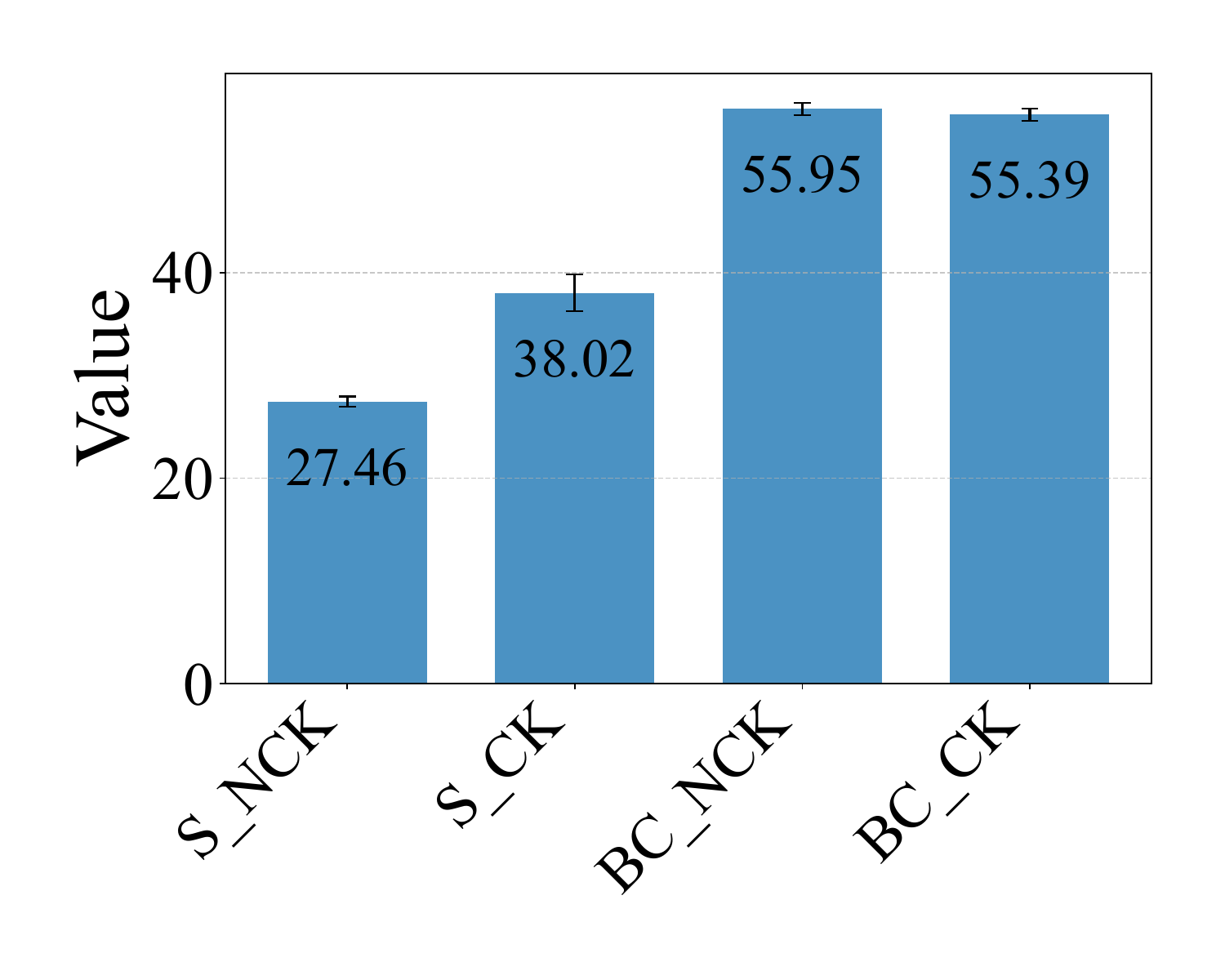} &
    \includegraphics[width=0.32\columnwidth]
      {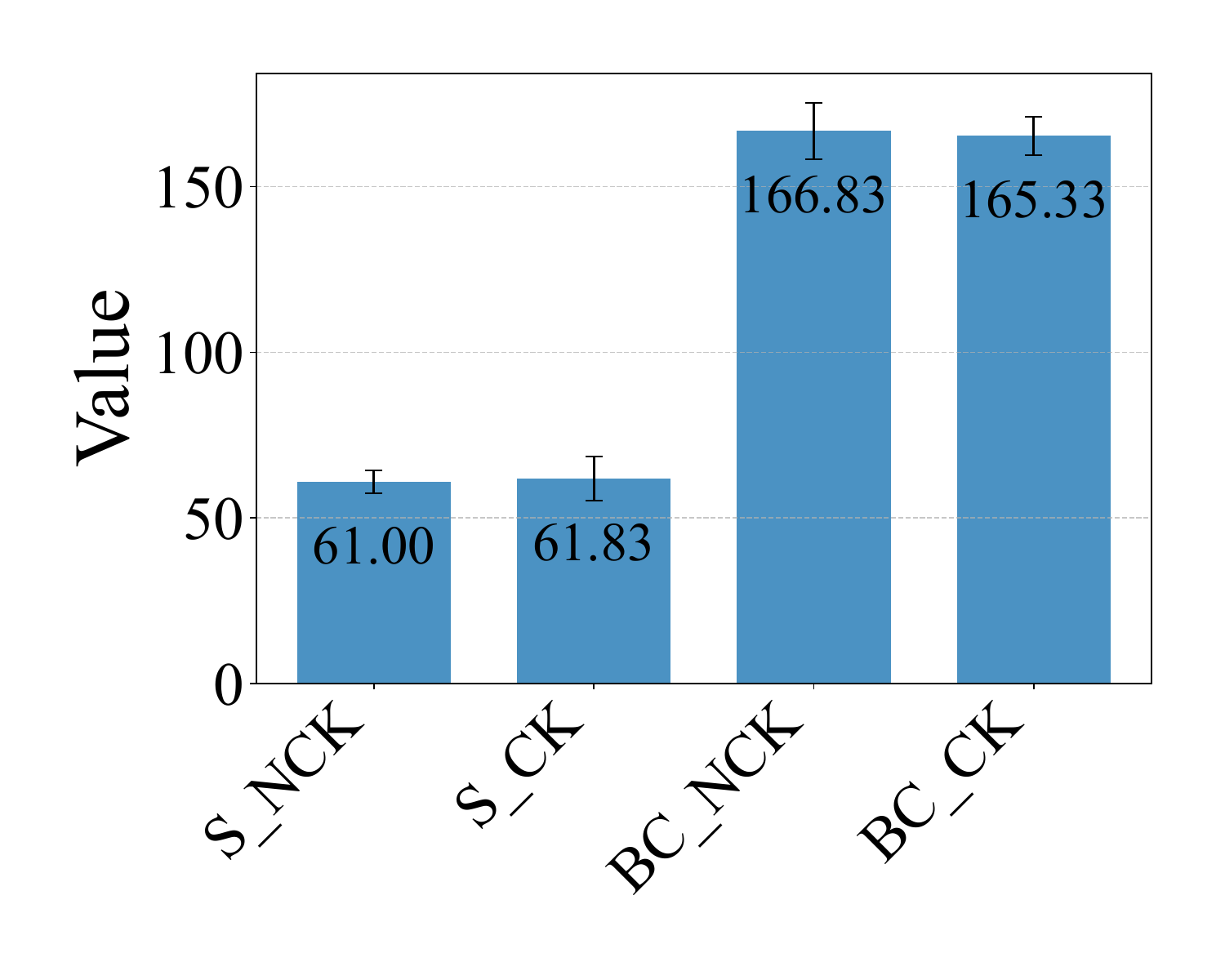} \\[-2ex]
    {\small (a) Conversations} &
    {\small (b) Knowledge} &
    {\small (c) Incentives}
  \end{tabular}
  \caption{Results of the Information Delivery experiment.}
  \label{fig:infop_qualitative_analysis}
\end{figure}

Figure~\ref{fig:infop_qualitative_analysis} and Table~\ref{tab:info_delivery_comparison} demonstrate that common knowledge exerts divergent effects depending on the delivery strategy. Under seeding, common knowledge increases conversation volume by 773.7\%, whereas under broadcasting, it reduces conversations by 64.6\%. This reduction closely aligns with the 63\% decrease reported in the original field study. Similar directional patterns are observed in knowledge accuracy and rational choice quality. Notably, informing only five seeds with common knowledge yields knowledge levels comparable to broadcasting without common knowledge. These results corroborate the image-concern mechanism proposed by \citet{banerjee2024less}: publicly designating a few informed seeds encourages dialogue, while revealing universal awareness suppresses clarification and social learning.

% \begin{table}[t]
% \centering
% \begin{threeparttable}

% {\small
% \setlength{\tabcolsep}{2pt}
% \renewcommand{\arraystretch}{1.05}
% \begin{tabular}{@{}llccc@{}}
% \toprule
% \textbf{Outcome}
% & \textbf{Source}
% & \textbf{S}
% & \textbf{BC}
% & \textbf{Cons.} \\
% \midrule

% \multirow{2}{*}{Conversation}
% & Field
% & $+103.0\%\uparrow$
% & $-63.0\%\downarrow$
% & \multirow{2}{*}{$\checkmark$} \\
% & \textit{Eco3S}
% & $+773.7\%\uparrow$
% & $-64.6\%\downarrow$
% & \\

% \addlinespace

% \multirow{2}{*}{Knowledge}
% & Field
% & $+5.6\%\uparrow$
% & $-3.2\%\downarrow$
% & \multirow{2}{*}{$\checkmark$} \\
% & \textit{Eco3S}
% & $+38.5\%\uparrow$
% & $-1.0\%\downarrow$
% & \\

% \addlinespace

% \multirow{2}{*}{Choice Quality}
% & Field
% & $+81.0\%\uparrow$
% & $-48.0\%\downarrow$
% & \multirow{2}{*}{$\checkmark$} \\
% & \textit{Eco3S}
% & $+1.3\%\uparrow$
% & $-0.9\%\downarrow$
% & \\

% \bottomrule
% \end{tabular}
% }

% \caption{Comparison of information-delivery dynamics with field results.}
% \label{tab:info_delivery_comparison}

% \begin{tablenotes}[flushleft]
% \small
% \item[] \textit{Note:} Values denote percentage changes when Common
% Knowledge is introduced. S denotes Seeding; BC denotes Broadcasting.
% Field results are adapted from \citet{banerjee2024less}; simulation
% results are averaged over five runs.
% \end{tablenotes}

% \end{threeparttable}
% \end{table}

\begin{table}[t]
\centering
% 【AAAI-27 格式修正】删除禁止的 resizebox，通过缩短表头和减小列间距适配单栏。
{\small
\setlength{\tabcolsep}{2pt}
\renewcommand{\arraystretch}{1.05}
\begin{tabular}{@{}llccc@{}}
\toprule
\textbf{Outcome}
& \textbf{Source}
& \textbf{S}
& \textbf{BC}
& \textbf{Cons.} \\
\midrule

\multirow{2}{*}{Conversation}
& Field
& $+103.0\%\uparrow$
& $-63.0\%\downarrow$
& \multirow{2}{*}{$\checkmark$} \\
& \textit{Eco3S}
& $+773.7\%\uparrow$
& $-64.6\%\downarrow$
& \\

\addlinespace

\multirow{2}{*}{Knowledge}
& Field
& $+5.6\%\uparrow$
& $-3.2\%\downarrow$
& \multirow{2}{*}{$\checkmark$} \\
& \textit{Eco3S}
& $+38.5\%\uparrow$
& $-1.0\%\downarrow$
& \\

\addlinespace

\multirow{2}{*}{Choice Quality}
& Field
& $+81.0\%\uparrow$
& $-48.0\%\downarrow$
& \multirow{2}{*}{$\checkmark$} \\
& \textit{Eco3S}
& $+1.3\%\uparrow$
& $-0.9\%\downarrow$
& \\

\bottomrule
\end{tabular}
}
\caption{Comparison of information-delivery dynamics with field results.}
\label{tab:info_delivery_comparison}
\vspace{-1em}
\end{table}

% These results are explained by image concerns: agents are reluctant to ask questions or seek clarification when doing so might signal ignorance or low competence. In the seeding condition with common knowledge, however, only a few are designated as informed, making it socially acceptable and even expected. The findings align with a signaling model in which the visibility of information receipt alters the strategic incentives to communicate. Ultimately, the experiment supports a "less is more" principle: carefully designed, limited information release combined with transparent seeding can outperform universal broadcasting by fostering richer social learning and better cognitive outcomes.

\subsection{Structural Causal Simulation}
% In reality, counterfactual states, what would have happened under different conditions, cannot be directly observed. Grounded in the principles of SCM, \textit{Eco3S} implements a \textbf{Structural Causal Simulation (SCS)} framework. It provides a controlled, generative environment to quantify the causal effects of policy interventions or environmental shifts, where dynamic agent interactions serve as the underlying causal mechanisms. We formally define the causal effect $Y_{\text{causal}}$ as the relative change in a system outcome:

Structural Causal Simulation evaluates counterfactuals by replaying a common baseline state under alternative interventions. For system parameters $P$, intervention parameters $P'$, and evaluation period $\Delta_t$, the relative causal effect is

{\small
\begin{equation}
Y_{\text{causal}}(P, P', \Delta_t) = \frac{Eco3S(P', \Delta_t) - Eco3S(P, \Delta_t)}{Eco3S(P, \Delta_t)},
\end{equation}
}

where \( P \) and \( P' \) denote the system parameters under the baseline and intervention scenarios, respectively,  and $\Delta_t $ specifies the evaluation period.

% function $f$ under an altered parameter configuration $p'$ compared to the original baseline configuration $p$, calculated over a specific simulation interval from the start step $t_{\text{start}}$ to the end step $t_{\text{end}}$:

% We conduct a series of counterfactual experiments within the Canal Decay and Rebellion simulation to demonstrate the flexibility and analytical utility of our modeling framework. Specifically, we evaluate three alternative scenarios that isolate the effects of key historical drivers: (1) no climate shock, (2) efficient government maintenance, and (3) no sea-transportation.
% As illustrated in Table \ref{tab:canal_decay_counterfactual_effect}, under the no climate shock condition, the canal remains navigable throughout the simulation horizon and is not abandoned on average. Under efficient maintenance, navigability is preserved for an extended period, though the canal is ultimately decommissioned, suggesting that maintenance alone cannot fully offset other systemic pressures. In contrast, when sea transportation is unavailable, the canal remains continuously operational across the entire simulation period. These results highlight the pivotal role of the shift to sea-based tribute logistics, which is the proximate cause of the Grand Canal’s abandonment, consistent with historical analyses in the literature.
We test three interventions: efficient maintenance, the removal of climate shocks, and the removal of sea transport. Table~\ref{tab:canal_decay_counterfactual_effect} shows that all three reduce outcome intensity, duration, and ratio relative to the baseline, with sea transport removal exerting the largest effect. While maintenance and climate influence the decline, the transition to maritime tribute transport remains the most decisive factor in canal abandonment.

% \begin{table}[t]
% \centering
% \caption{Causal effects of the Canal Decay experiment relative to the baseline.}
% \label{tab:canal_decay_counterfactual_effect}
% \footnotesize
% \setlength{\tabcolsep}{2.5pt}
% \renewcommand{\arraystretch}{0.9}

% \begin{tabular}{@{}lccc@{}}
% \toprule
% \multirow{2}{*}{\textbf{Experiment}}
% & \multicolumn{3}{c}{\textbf{Causal Effect}} \\
% \cmidrule(lr){2-4}
% & \shortstack{Mean Peak\\Intensity}
% & Duration
% & Ratio \\
% \midrule
% Efficient Maintenance & $-0.120$ & $-0.109$ & $-0.015$ \\
% w/o Climate Shock     & $-0.104$ & $-0.131$ & $-0.045$ \\
% w/o Sea Transport     & $-0.296$ & $-0.205$ & $-0.120$ \\
% \bottomrule
% \end{tabular}
% \vspace{-2em}
% \end{table}

\begin{table}[t]
\centering
{\small
\setlength{\tabcolsep}{3pt}
\renewcommand{\arraystretch}{1.05}

\begin{tabular}{@{}lccc@{}}
\toprule
\multirow{2}{*}{\textbf{Experiment}}
& \multicolumn{3}{c}{\textbf{Causal Effect}} \\
\cmidrule(lr){2-4}
& \shortstack{Mean Peak\\Intensity}
& Duration
& Ratio \\
\midrule
Efficient Maintenance & $-0.120$ & $-0.109$ & $-0.015$ \\
w/o Climate Shock     & $-0.104$ & $-0.131$ & $-0.045$ \\
w/o Sea Transport     & $-0.296$ & $-0.205$ & $-0.120$ \\
\bottomrule
\end{tabular}
}
\caption{Causal effects of the Canal Decay experiment relative to the baseline.}
\label{tab:canal_decay_counterfactual_effect}
\vspace{-1em}
\end{table}

\subsection{Comparison with Baseline Simulation Approaches}
\begin{figure}[!ht]
  \centering
  \begin{tabular}{@{}c@{\hspace{0.01\columnwidth}}c@{}}
    \includegraphics[width=0.48\columnwidth]{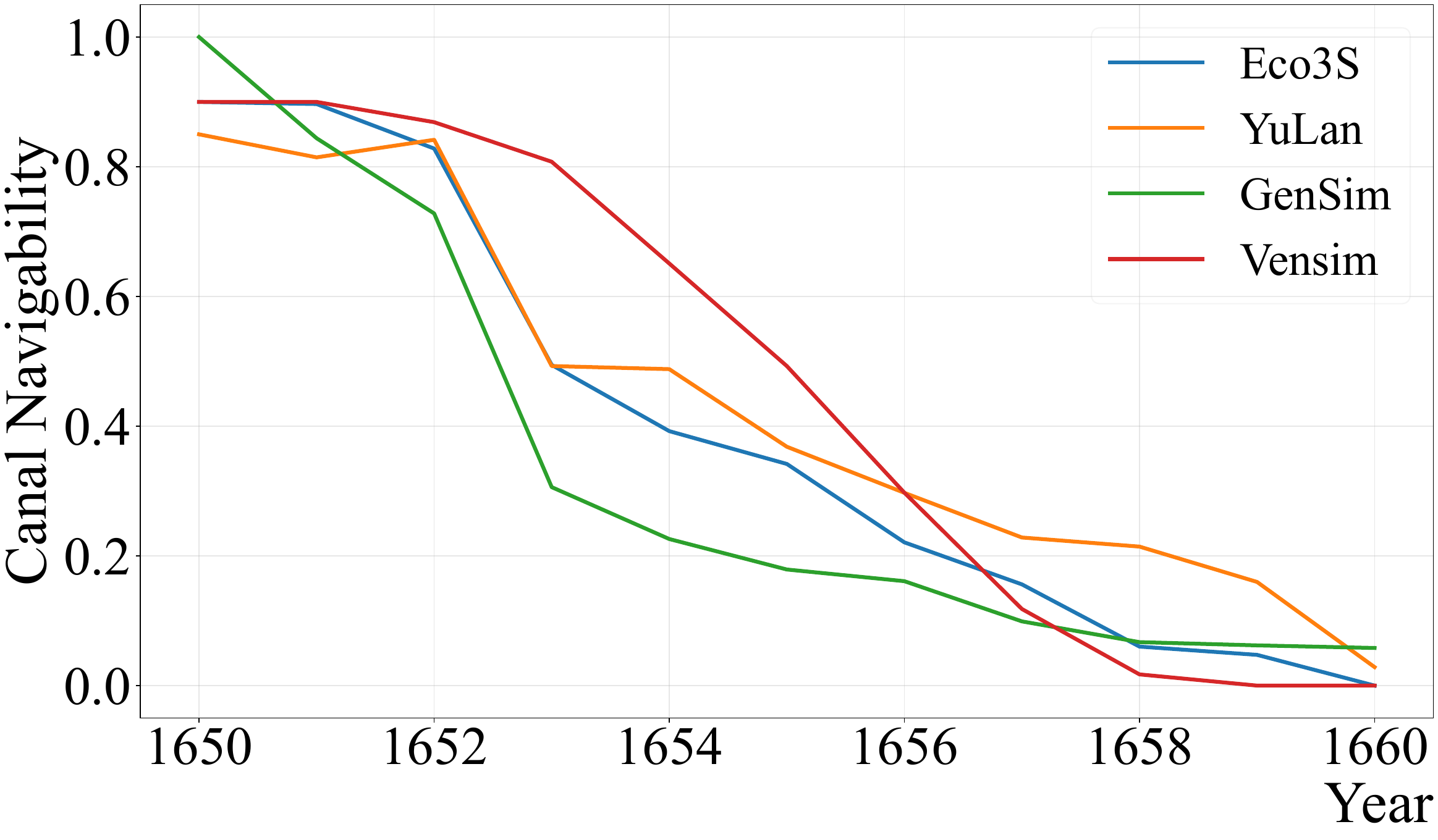} &
    \includegraphics[width=0.48\columnwidth]{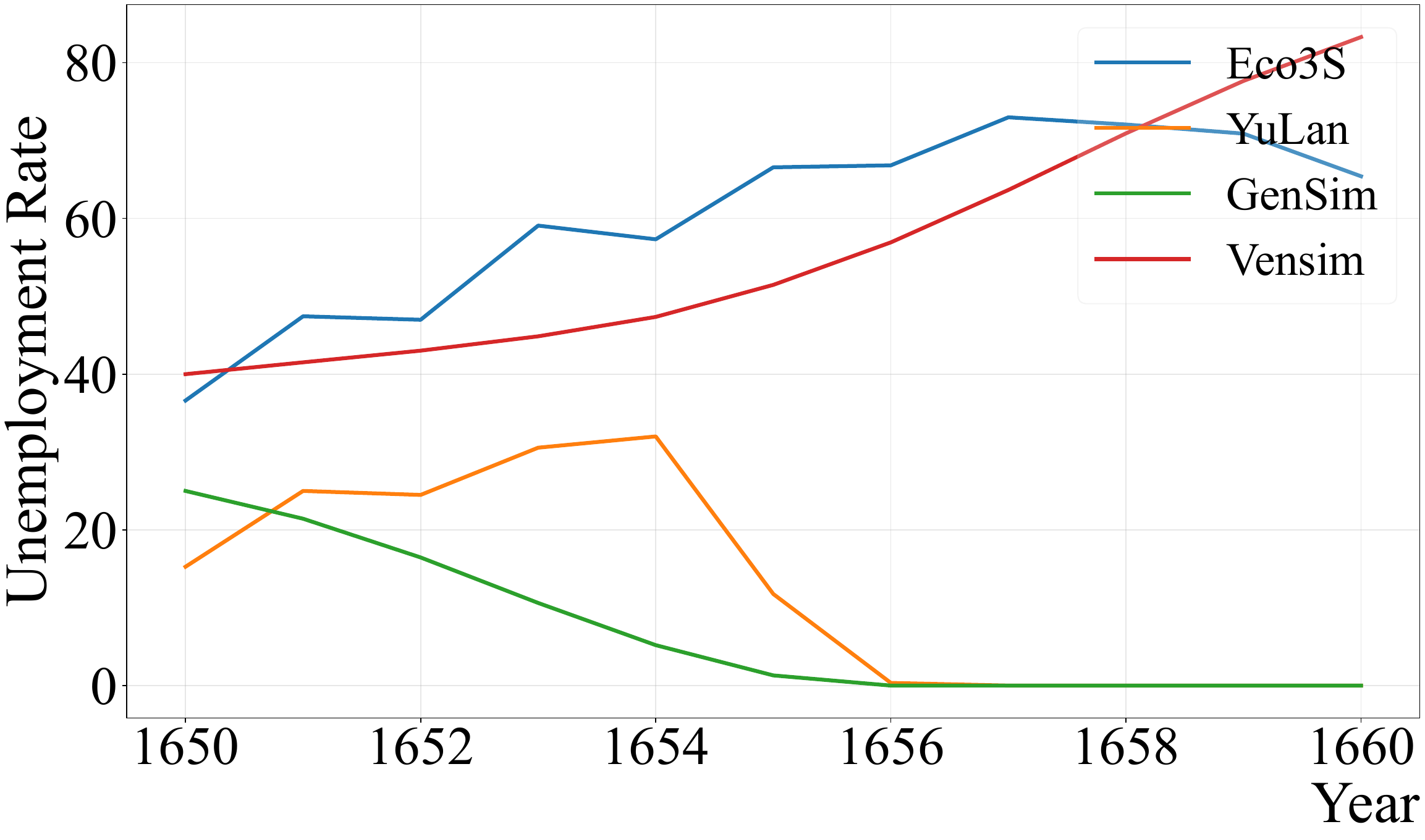} \\[-2ex]
    {\scriptsize (a) River Navigability} &
    {\scriptsize (b) Unemployment Rate} \\[0.8ex]
    \includegraphics[width=0.48\columnwidth]{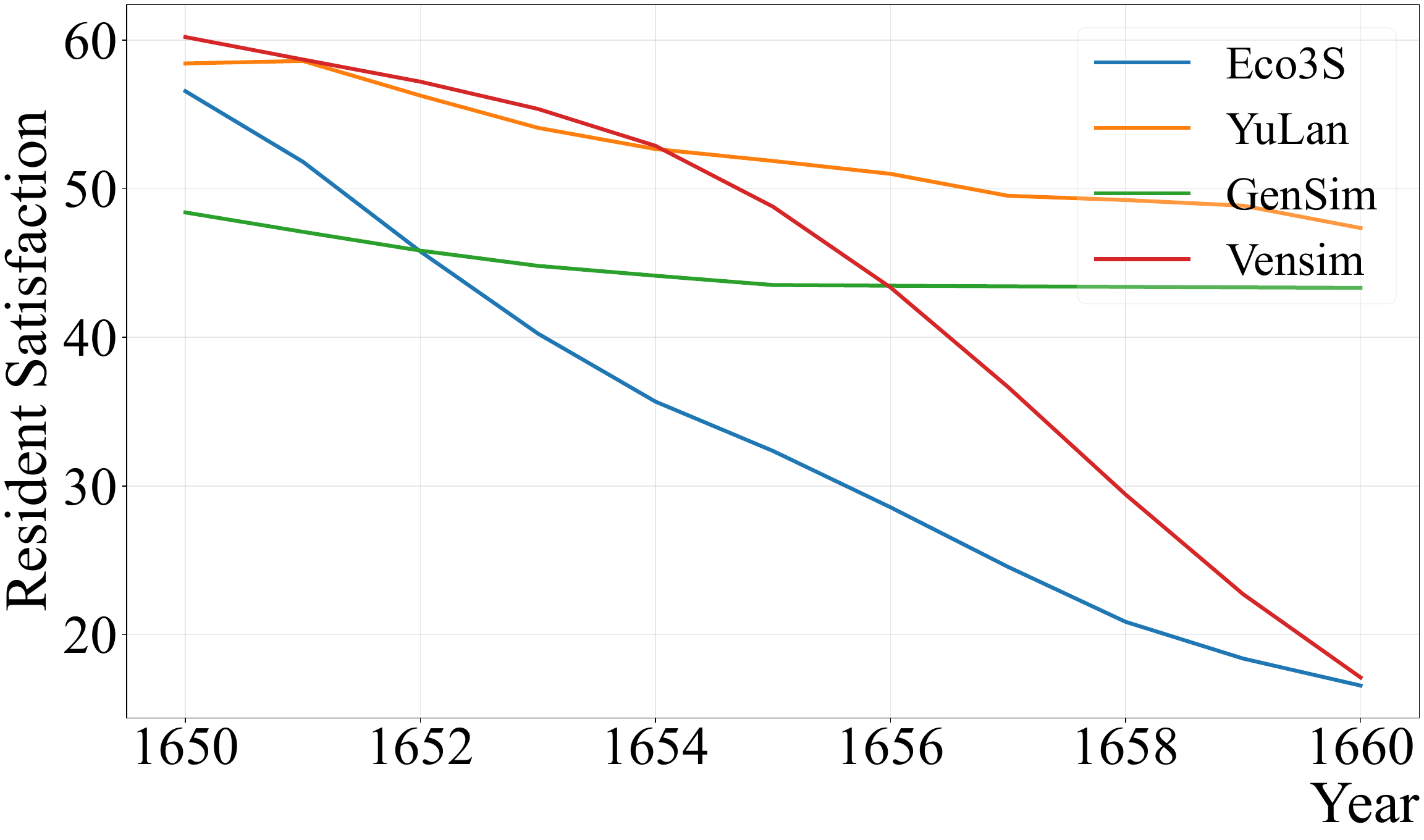} &
    \includegraphics[width=0.48\columnwidth]{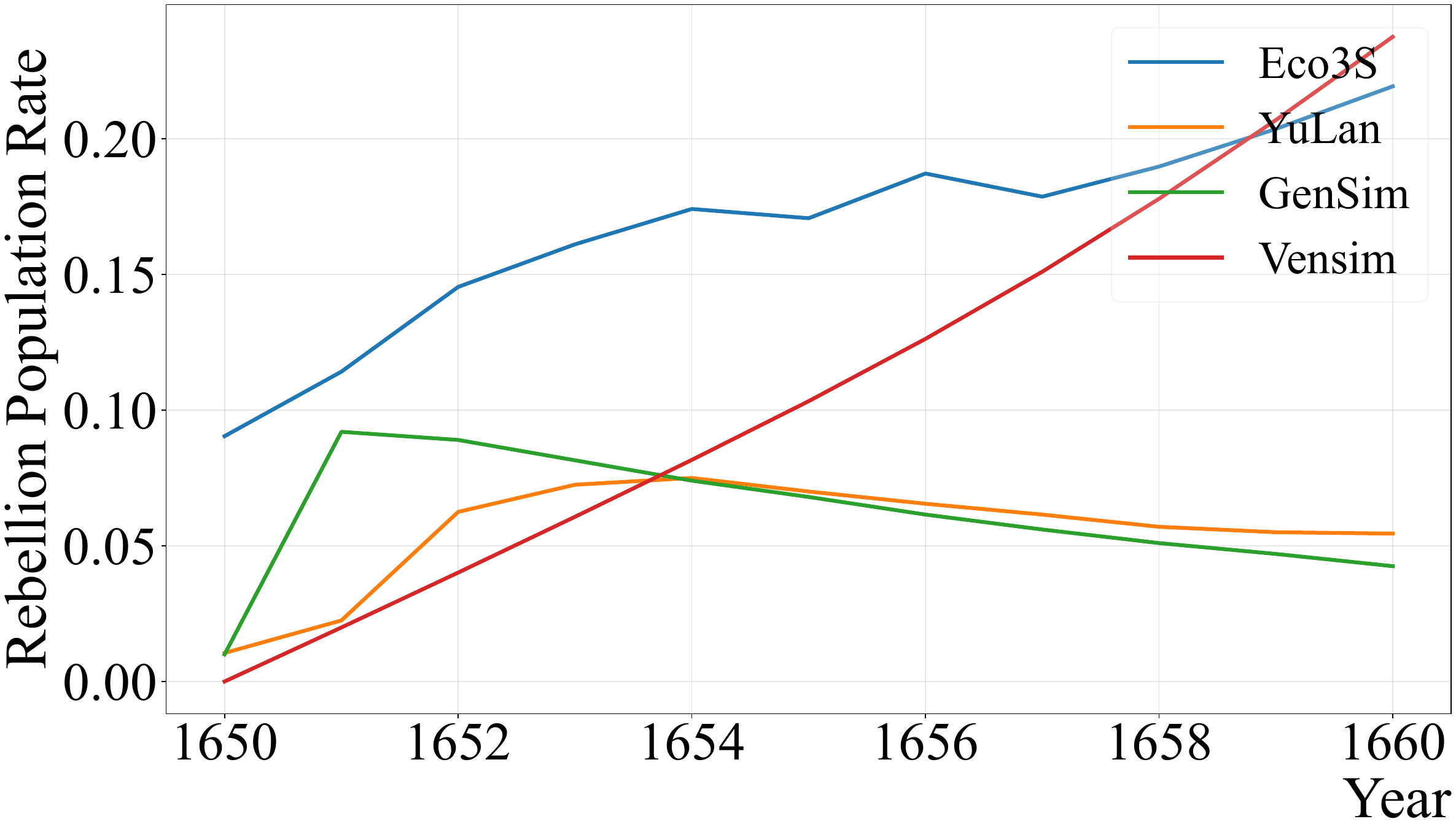} \\[-2ex]
    {\scriptsize (c) Average Satisfaction} &
    {\scriptsize (d) Rebellion Population Rate}
  \end{tabular}
  \vspace{-1em}
  \caption{Benchmark comparison for Canal Decay.}
  \label{fig:canal_decay_platforms}

\end{figure}

% To validate the advancement of \textit{Eco3S}, we conduct head-to-head experiments against two representative LLM-based simulation platforms: YuLan-OneSim~\citep{wang2025yulanonesim} and GenSim~\citep{tang2025gensim}, as well as a traditional System Dynamics (SD) modeling tool (Vensim PLE). We implement the Canal Decay scenario on all platforms with 2,000 agents over a 10-year period.
% As shown in Figure \ref{fig:canal_decay_platforms}, Vensim exhibits trends that are largely consistent with those of \textit{Eco3S}. However, since it relies on macroscopic mathematical equations rather than individual-level modeling, Vensim produces excessively smooth aggregate curves. Compared to this traditional SD approach, agent-based modeling in \textit{Eco3S} demonstrates three core advantages: (1) Individual Heterogeneity: it simulates thousands of unique agents with diverse profiles and social ties instead of a homogeneous mass; (2) Spatio-Temporal Heterogeneity: it integrates a co-evolving environment to capture how localized infrastructure decay triggers regional instability, which aggregate models fail to reflect; and (3) Flexible Decision-making: it utilizes complex agent-level logic rather than fixed mathematical equations. By modeling bottom-up behaviors, \textit{Eco3S} successfully captures emergent phenomena (like local rebellions) often overlooked in traditional SD models.
We evaluate \textit{Eco3S} against YuLan-OneSim~\citep{wang2025yulanonesim}, GenSim~\citep{tang2025gensim}, and Vensim PLE using a 2,000-agent, 10-year Canal Decay scenario. Figure~\ref{fig:canal_decay_platforms} indicates that Vensim captures aggregate trends but produces smooth trajectories via macroscopic equations. By contrast, \textit{Eco3S} models individual heterogeneity and localized shocks, where micro-level decisions and spatial variance drive emergent aggregate dynamics.

Figures~\ref{fig:canal_decay_platforms}(a) and (b) indicate that all platforms capture the decline in navigability, yet their labor market responses diverge. In \textit{Eco3S}, unemployment rises alongside canal deterioration, while YuLan-OneSim and GenSim exhibit weaker or logically inconsistent responses. This disparity underscores the efficacy of Co-evolving Environment Design, which couples environmental states with employment via explicit job markets and spatial differentiation.
Social stability indicators in panels (c) and (d) reveal a similar contrast. Declining satisfaction in \textit{Eco3S} leads to rising rebellion, while baseline platforms produce flatter trajectories. Limitations in YuLan-OneSim, such as fixed behavior graphs, and GenSim, which lacks social networks, hinder behavioral adaptation and sentiment diffusion. By distinguishing between town types and modeling population dynamics, \textit{Eco3S} captures regional economic shocks and rebellion disparities that homogeneous baselines with fixed populations cannot represent.

\subsection{Ablation Study}

We evaluate the full model against three ablations via five independent runs: government decisions without deliberation, rule based decisions instead of LLM reasoning, and the removal of the HIN.

\begin{figure}[!ht]
  \centering
  \begin{tabular}{@{}c@{\hspace{0.01\columnwidth}}c@{}}
    \includegraphics[width=0.48\columnwidth]{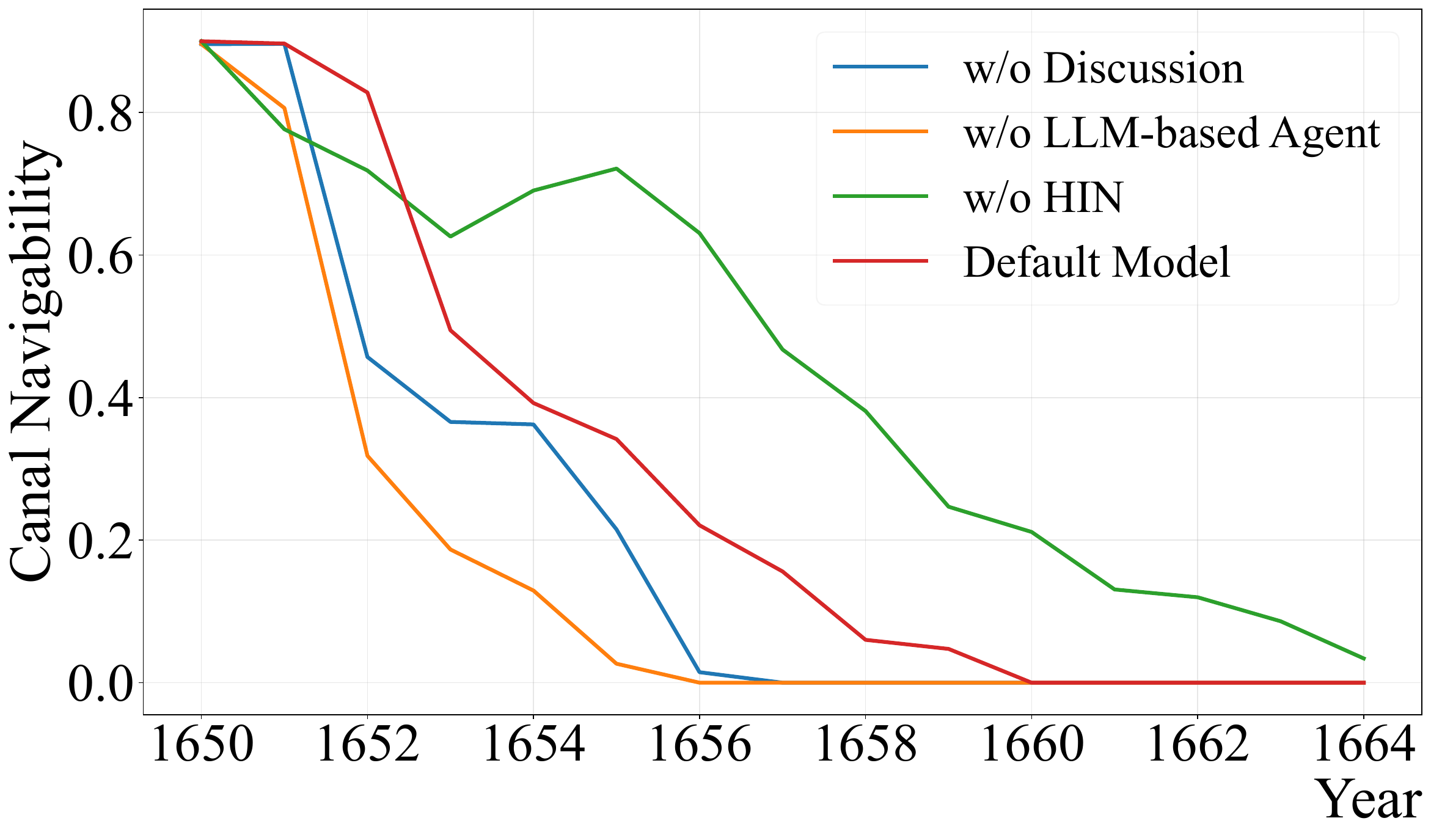} &
    \includegraphics[width=0.48\columnwidth]{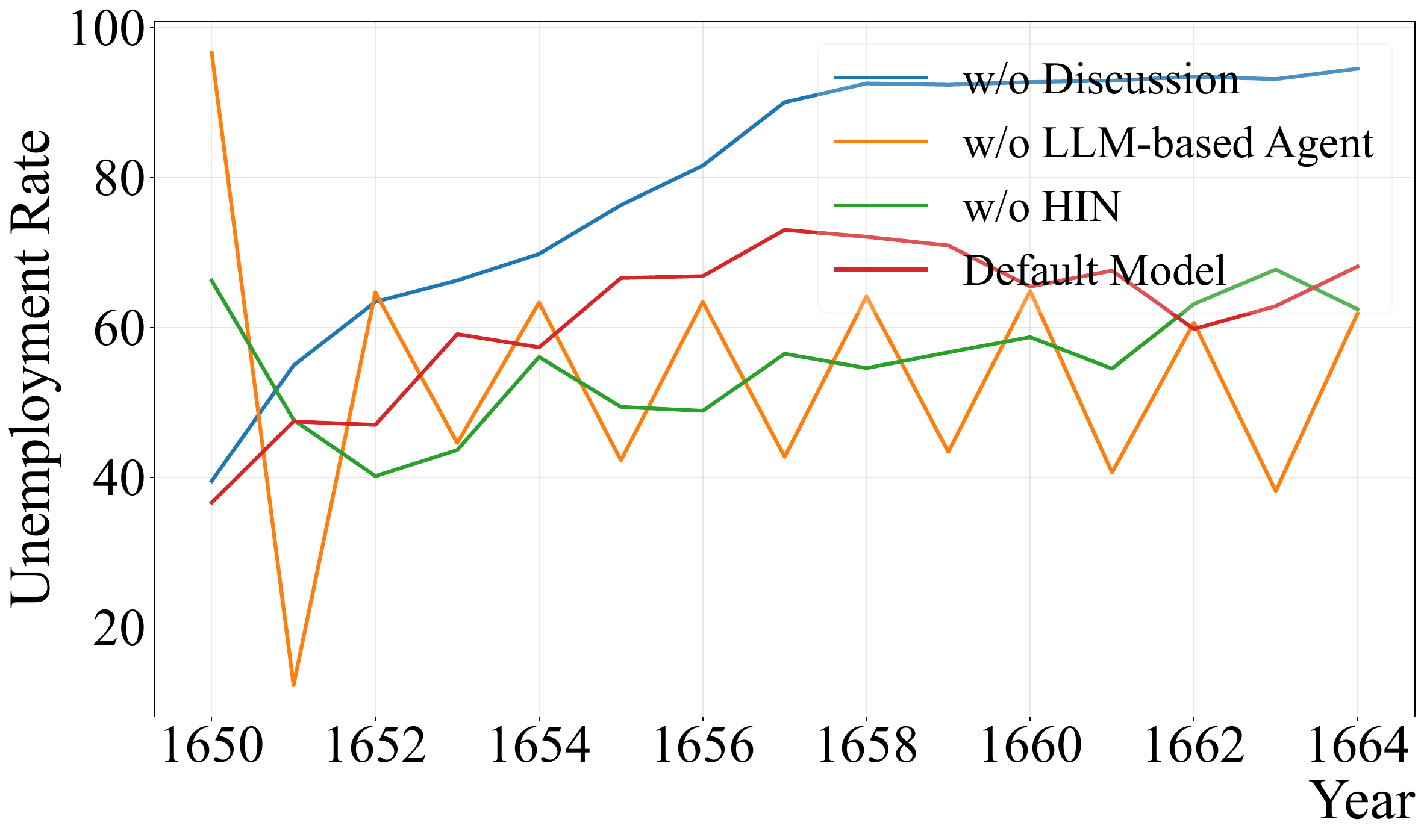} \\[-2ex]
    {\scriptsize (a) River Navigability} &
    {\scriptsize (b) Unemployment Rate} \\[0.8ex]
    \includegraphics[width=0.48\columnwidth]{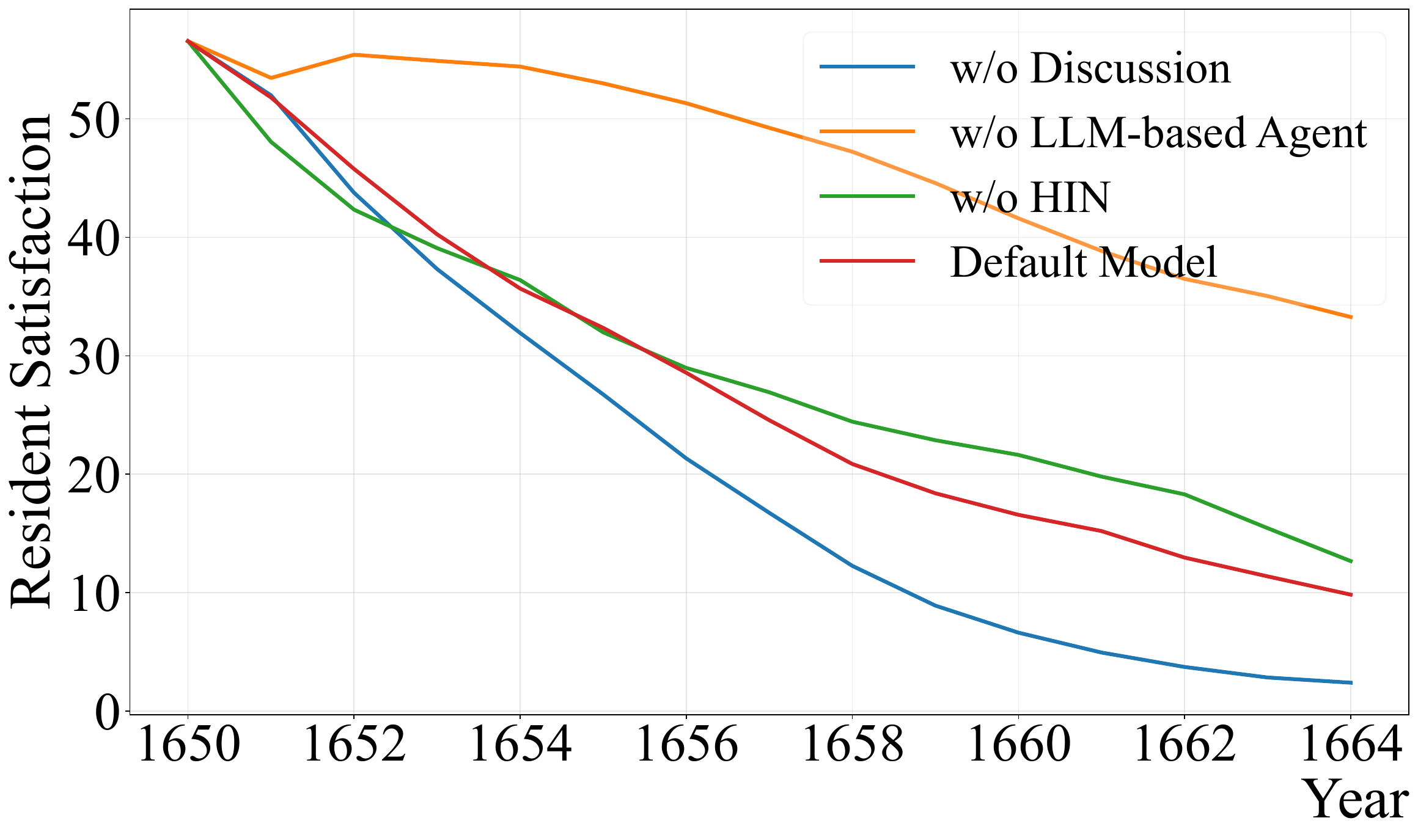} &
    \includegraphics[width=0.48\columnwidth]{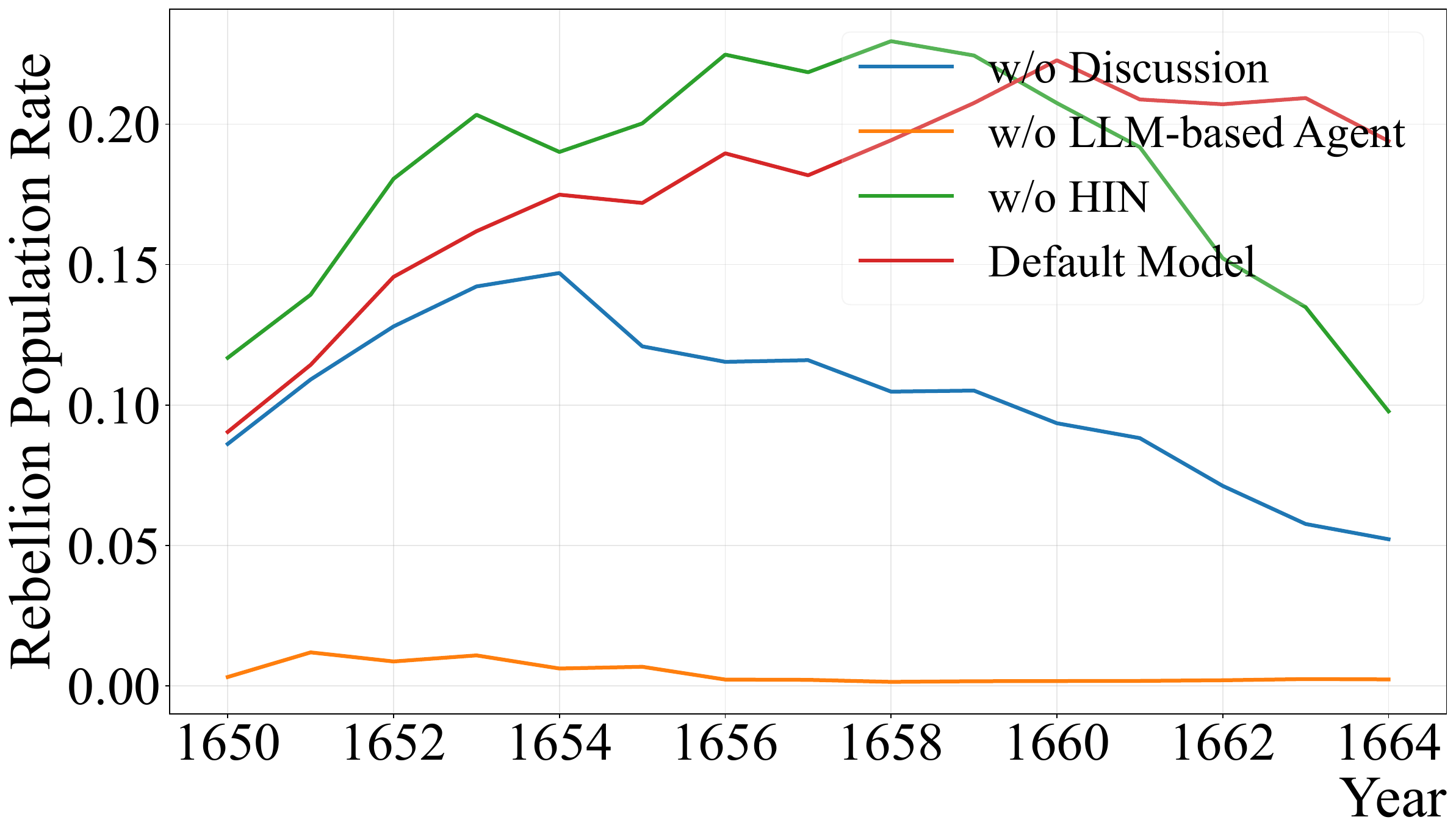} \\[-2ex]
    {\scriptsize (c) Average Satisfaction} &
    {\scriptsize (d) Rebellion Population Rate}
  \end{tabular}
  \caption{Ablation study of the Canal Decay experiment.}
  \label{fig:canal_decay_ablation_study}
  \vspace{-1em}
\end{figure}

As illustrated in Figure~\ref{fig:canal_decay_ablation_study}(a) through (c), omitting official deliberation accelerates navigability loss, increases unemployment, and reduces satisfaction, suggesting that iterative discussion stabilizes resource allocation. Panel (d) reveals a lower rebellion rate for this ablation. This does not indicate improved social conditions; instead, more extreme government policies, such as higher tax rates and intensified control, suppress unrest even as satisfaction deteriorates.
Substituting LLM reasoning with fixed rules also disrupts the simulated dynamics. The unemployment curve fluctuates without a coherent response to navigability loss, while rebellion fails to exhibit a sustained increase. This suggests that static rules cannot replicate the context aware coupling across environmental, economic, and social states. Removing the HIN yields a distinct effect where rebellion is initially higher due to stochastic initialization but subsequently falls below full model levels, as restricted information exchange hinders the diffusion of rebellious sentiment.

Collectively, these ablations demonstrate that deliberative governance, LLM reasoning, and heterogeneous social interactions are essential to the emergent dynamics in \textit{Eco3S}.

\subsection{System Scalability, Robustness, and Complexity Analysis}

We evaluate the scalability, behavioral robustness, and computational complexity of the \textit{Eco3S} framework.
% To validate scalability and robustness, we conducted experiments across varying agent population sizes ($N \in [100, 10,000]$) and different simulation horizons ($T \in [5, 20]$). Experimental results demonstrate that scaling up the agent population effectively mitigates stochastic micro-level fluctuations, thereby significantly reinforcing the macroscopic robustness of emergent socio-economic phenomena. This confirms that \textit{Eco3S} can robustly scale to large and complex economic scenarios without suffering from behavioral degradation. 
We evaluate scalability across populations $N \in [100, 10,000]$ and simulation horizons $T \in [5, 20]$. Increasing the population size mitigates stochastic micro level fluctuations and stabilizes emergent macro level patterns, confirming that \textit{Eco3S} scales without behavioral degradation.
% Details are provided in Appendix \ref{app:performance}.
We further evaluated backend independence using five LLMs from different providers. While most models successfully replicated the core qualitative dynamics, Qwen3-8B exhibited a higher rebellion rate and an inconsistent temporal trajectory. This suggests that the framework is robust across mainstream providers, though outcomes remain contingent on the reasoning capacity of the base model.

For a single simulation run, \textit{Eco3S} bounds agent state storage via memory summarization, yielding a space complexity of $\mathcal{O}(N + |E|)$, where $N$ and $|E|$ denote the number of agents and network edges, respectively. Additionally, by leveraging concurrent processing, the time complexity for simulating $T$ steps is reduced to $\mathcal{O}(\frac{N \times T}{P})$, where $P$ is the parallel processing capacity. 

Beyond single runs, the SAR pipeline introduces a meta-level search complexity of $\mathcal{O}(K \times C_{sim})$, where $C_{sim}$ is the single-run cost and $K$ is the number of refinement iterations. Rather than striving for an exact mathematical optimum, \textit{Eco3S} adopts a diagnostic-guided \textit{satisficing} principle. 
The refinement process is capped at \(K\leq10\) iterations and converged within five iterations in our experiments.
% This ensures $K$ converges rapidly and robustly ($K \le 10$), maintaining overall computational feasibility. 

\subsection{Auto-simulation}

We evaluate the Auto-simulation system across four representative phenomena initiated from high-level research prompts. The synthesized simulations successfully replicate financial herding, marked by a 167\% increase in average holdings; hysteresis between customer satisfaction and loyalty; endogenous asset bubbles, which saw a 191\% price surge despite declining fundamentals; and Schelling segregation, where the dissimilarity index rose from 0.29 to 0.74. All cases converge within the prescribed iteration limits. These results demonstrate that the SAR pipeline can generate executable models that recover established qualitative patterns while significantly reducing manual configuration effort.

\section{Discussion}
The implementation of \textit{Eco3S} highlights several key considerations for LLM-driven economic simulations:
(1) The "black box" nature of LLM reasoning requires improved transparency; future work will explore attention-weight analysis to support rigorous validation.
(2) While the current framework captures dynamic climate and river systems, the physical environment could be enriched with urban transportation networks, mountains, and natural resource distributions, along with more flexible simulation mechanisms.
(3) Evaluating simulation validity without historical benchmarks remains challenging. Potential solutions include cross-configuration convergence testing (multiple structurally distinct designs for the same question) and adversarial robustness validation (using a separate LLM critic to challenge conclusions).

\section{Conclusion}
% In this paper, we introduce \textit{Eco3S}, a novel, end-to-end simulation framework designed to advance computational approaches in economic research and policy evaluation. 
% By integrating Co-evolving Environment Design, Structural Causal Simulation, and the Simulation-Analysis-Refinement paradigm, the framework addresses the limitations of static environments and manual simulation workflows in existing studies. 
% % \textit{Eco3S} integrates four core capabilities within a unified architecture: (1) fine-grained environmental modeling, (2) counterfactual experimentation, (3) automated simulation with human-in-the-loop feedback, and (4) built-in analytical modules for post-simulation interpretation.
% We demonstrate the framework’s versatility and validity through an extensive suite of experiments spanning both canonical economic phenomena and historically grounded case studies drawn from high-impact research. These experiments confirm that \textit{Eco3S} not only reproduces established theoretical insights but also enables rigorous, scalable exploration of complex socio-economic dynamics.
% We envision \textit{Eco3S} as a powerful tool for both ex ante and ex post validation of economic theories and policy interventions, offering researchers and policymakers a sandbox.
% % for stress-testing hypotheses, anticipating unintended consequences, and refining designs before real-world implementation.
% Future work will incorporate reinforcement learning to improve the system’s ability to understand researcher intent and autonomously execute tailored simulations.

This paper introduces \textit{Eco3S}, an end-to-end framework designed for computational economic research and policy evaluation. By integrating co-evolving environment design, structural causal simulation, and the Simulation-Analysis-Refinement paradigm, the framework models dynamic socio-economic systems, facilitates counterfactual reasoning, and automates simulation workflows. Evaluations spanning historical case studies and canonical economic phenomena demonstrate that \textit{Eco3S} replicates established findings while ensuring scalability across diverse settings. The system provides a controlled environment for both ex ante and ex post analysis, allowing researchers to stress-test hypotheses and assess policy consequences prior to implementation. Future research will explore reinforcement learning to further refine intent interpretation and simulation tailoring.

\bibliography{eco3s_refs}

\appendix
\setcounter{secnumdepth}{2}
% \documentclass[letterpaper]{article} % DO NOT CHANGE THIS
% \usepackage[submission]{aaai2027} % DO NOT CHANGE THIS
% \usepackage[hyphens]{url} % DO NOT CHANGE THIS
% \usepackage{graphicx} % DO NOT CHANGE THIS
% \urlstyle{rm} % DO NOT CHANGE THIS
% \def\UrlFont{\rm} % DO NOT CHANGE THIS
% \usepackage{natbib} % DO NOT CHANGE THIS AND DO NOT ADD OPTIONS
% \usepackage{caption} % DO NOT CHANGE THIS AND DO NOT ADD OPTIONS
% \frenchspacing % DO NOT CHANGE THIS
% \usepackage{amsmath}
% \usepackage{amssymb}
% \usepackage{booktabs}
% \usepackage{threeparttable}
% \usepackage{multirow}
% \usepackage{xspace}

% \pdfinfo{
% /TemplateVersion (2027.1)
% }

% \setcounter{secnumdepth}{2}
% \renewcommand{\thesection}{S\arabic{section}}
% \renewcommand{\thesubsection}{\thesection.\arabic{subsection}}
% \renewcommand{\thefigure}{S\arabic{figure}}
% \renewcommand{\thetable}{S\arabic{table}}
% \renewcommand{\theequation}{S\arabic{equation}}

% \newcommand{\cmark}{\checkmark}
% \newcommand{\xmark}{$\times$}
% \newcommand{\tabincell}[2]{\begin{tabular}{@{}#1@{}}#2\end{tabular}}

% \title{Supplementary Material for\\Eco3S: Complex Socio-Economic System Simulation via Agent-Based Models}
% \author{Anonymous Submission}
% \affiliations{}

% \begin{document}
% \maketitle

% This supplementary document provides the detailed study comparison, experimental configurations, additional auto-simulation results, robustness and scalability analyses, and system interface descriptions referenced by the main paper.

\section{Main Experiment Details}
\label{appendix:main_experiment}

\subsection{Parameter Development and Selection}
\label{appendix:parameter_selection}

We conducted non-exhaustive pilot tests rather than optimizing parameters against the reported evaluation metrics. Final settings were selected by jointly considering the stability of the principal qualitative trends, diversity in agent-level behavior, and computational cost. The values examined during development and the scenario-specific final settings are summarized in Table~\ref{tab:parameter_selection}.

\begin{table*}[t]
\centering
\small
\setlength{\tabcolsep}{4pt}
\begin{tabular}{@{}p{0.16\textwidth}p{0.21\textwidth}p{0.23\textwidth}p{0.32\textwidth}@{}}
\toprule
\textbf{Parameter} &
\textbf{Values Examined} &
\textbf{Final Setting} &
\textbf{Selection Criterion} \\
\midrule
Response probability &
$0.05$, $0.2$, $0.3$, $0.4$, $0.5$, $0.8$ &
$0.05$ for the historical simulations; $0.2$ for Information Delivery &
Population-dependent LLM-call volume and runtime, subject to preservation of the principal qualitative trends \\
\addlinespace
LLM temperature &
$0.6$, $0.7$, $0,8$, $1.0$ &
$1.0$ &
Greater diversity in agent-level decisions while retaining stable aggregate behavior \\
\addlinespace
Population size &
$100$, $200$, $500$, $2{,}000$, $10{,}000$ &
$2{,}000$ for the historical simulations; $200$ for Information Delivery &
Balance between aggregate stability and computational and API cost \\
\addlinespace
Simulation horizon &
$3$, $5$, $10$, $15$, $20$ steps &
$15$ for Canal Decay; $10$ for Origins of Governance; $3$ for Information Delivery &
Coverage of the relevant temporal dynamics without unnecessary computation \\
\bottomrule
\end{tabular}
\caption{Parameter values examined during development and the criteria used to select the final experimental settings.}
\label{tab:parameter_selection}
\end{table*}

Response probability was selected jointly with population size because the expected number of agent responses, and consequently the number of LLM calls, increases with both quantities. For the historical simulations with $2{,}000$ agents, we selected a response probability of $0.05$ to keep execution costs manageable. Higher probabilities substantially increased runtime without materially changing the principal qualitative trends. For the Information Delivery experiment with $200$ agents, we used a response probability of $0.2$ to retain sufficient opportunities for short-run information exchange at the smaller population scale.

The Information Delivery experiment models a representative village using a synthetic population of $200$ agents. Its three simulation steps intentionally correspond to the three-day interval between intervention and outcome measurement in the original field experiment~\citep{banerjee2024less}. We tested LLM temperatures of $0.7$ and $1.0$ and selected $1.0$, the higher of the tested values, to encourage diversity in individual decisions while retaining stable aggregate behavior.

% 【Checklist 4.8 补充说明】
% 本小节说明实验实际使用的主要计算设备及本地计算与远程 API 的分工。
\subsection{Computing Infrastructure}

Most experiments were conducted on a Lenovo Legion Y9000P IAX10H laptop equipped with an Intel Core Ultra 9 275HX processor and an NVIDIA GeForce RTX 5070 Ti Laptop GPU with 12\,GB of GPU memory. A subset of the experiments was conducted on a Windows 10 workstation equipped with an AMD Ryzen 5 5600X processor and an NVIDIA T600 GPU. The local machines were used for simulation execution, data processing, and result analysis. LLM inference was performed through remote APIs in most experiments; the local-model configuration included in the cross-model robustness evaluation was the exception.

Unless otherwise specified, each reported quantitative experimental configuration was independently executed at least five times.

% 【Checklist 4.12 补充说明】
\subsection{Statistical Analysis}
Statistical significance was assessed for inferential comparisons with appropriate repeated observations. Spatial differences in rebellion outcomes were evaluated using independent-samples $t$-tests together with Cohen's $d$ effect sizes, while temporal trends were assessed using OLS regression. Comparisons in the baseline, ablation, robustness, and scalability experiments are treated as descriptive analyses of repeated simulation trajectories; no claims of statistical significance are made for these comparisons.

% 【Checklist 4.7 随机性与复现边界说明】
The reported experiments reused fixed initial agent profiles and identical scenario configurations across compared conditions. Independent runs retained stochastic runtime operations and nondeterministic LLM outputs; therefore, exact trajectory-level reproduction is not guaranteed. The accompanying analysis code documents how the corresponding run-level outputs are processed to reproduce the reported aggregate statistics.

\subsection{Rebel on the Canal}
\textbf{Agents.}
The simulation features two main types of agents: institutional actors (government and rebels) and residents.

Government and rebel organizations are structured hierarchically. Both government agents and rebel agents are categorized by rank into regular officials, secretary, and senior officials. Government agents are categorized into three policy factions, specifically pro-canal, pro-maritime, and neutral, which influence their stances on infrastructure investment. All agents are assigned personality types with randomly selected words to modulate decision-making styles. Decision-making within each organization follows a four-step process. First, all members asynchronously submit initial opinions. Subsequent rounds involve iterative discussion informed by prior exchanges. A secretary then synthesizes these inputs into a shared information pool. Finally, designated leaders make binding decisions based on this consolidated view.
% Each agent maintains both short-term and long-term memory. Short-term memory stores up to three recent dialogue entries. When this buffer fills, a LLM summarizes the conversation using the agent’s identity and prior context, appending the result to a list. In addition to personal memory (e.g., past statements), agents have access to collective memory, which records annual policy decisions and their macro-level consequences, such as changes in employment rates, public satisfaction, and canal navigability.

Government agents allocate annual resources across five dimensions: total public budget (an integer), the proportion of goods transported via canal (a float between 0 and 1), maintenance investment for the canal (integer), military support (integer), and tax adjustment (a float in [–0.1, +0.1]). Canal maintenance improves navigability, creates jobs for "canal maintenance workers," and reduces available fiscal resources. Rebel agents, by contrast, decide on propaganda budget, the scale of rebellion to stage, and the target towns, along with the resource allocations for each location.

Resident agents represent the working-age population. Each has attributes including gender, remaining lifespan, employment status, job type, income, health index (1–5), geographic coordinates, assigned town, and a randomly generated two-word personality descriptor (e.g., "cautious and pragmatic"). Resident satisfaction is represented on a five-level qualitative scale: "hate the regime, vow to overthrow it"; "deeply resentful, denounce its incompetence"; "indifferent, disengaged from politics"; "tolerable, willing to comply"; and "loyal, ready to die for the state." This categorical measure directly influences the probability of joining a rebellion. Residents possess only short-term memory (up to three interactions). Ordinary residents may choose to join a rebellion (more likely when dissatisfied), seek new employment if unemployed, migrate to another town upon job loss (guided by inter-town connectivity), update their satisfaction based on economic conditions and propaganda exposure, and post public remarks reflecting their mood. A subset of residents affiliated with rebels do not make active decisions but may post inflammatory statements with fixed probability.
% Government agents allocate resources across public budget, canal transport proportion, maintenance, military support, and tax adjustment. Rebel agents decide on propaganda, rebellion scale, and target towns. Resident agents represent the working-age population with attributes like gender, income, and a two-word personality descriptor. Resident satisfaction is measured on a five-level qualitative scale, directly influencing rebellion probability. Residents may choose to seek employment, migrate, or join a rebellion based on economic conditions and propaganda.

Regarding conflict dynamics, when rebellions occur, the attrition rates depend on the force ratio $\alpha = G/R$, where $G$ and $R$ represent the government and rebel forces, respectively. Letting $c = 0.1$ be the baseline attrition, $\alpha_{\max}$ the threshold for extreme attrition, and $d \in [0.5, 1.0]$ a decay coefficient, the attrition probabilities are calculated as $p_G = c/\alpha^d + \epsilon$ if $\alpha \leq \alpha_{\max}$ (otherwise $0.01 + \epsilon$) and $p_R = c \cdot \alpha^d + \epsilon$, with a stochastic noise term $\epsilon \sim \mathcal{U}(-0.03, 0.03)$. The actual numerical losses for each side are then given by $L_G = \lfloor p_G \cdot G \rfloor$ and $L_R = \lfloor p_R \cdot R \rfloor$.

\textbf{Environment.}
The environment consists of several interconnected subsystems.
The map system includes a two-dimensional river grid encoding the spatial layout of the Grand Canal, a scalar navigability index indicating current usability, a town connectivity graph in adjacency-list format for migration modeling, and a town attribute registry storing metadata such as name, coordinates, and town type (e.g., canal-side or inland).
Specifically, the canal navigability $\phi$ varies annually following the updating rule $\phi_{t+1} = \max(0, \phi_t \cdot (1 - \delta) - \gamma \cdot 0.6)$, where $\delta$ is the natural decay rate and $\gamma$ is a climate impact factor.
% The geographical distribution of the canal segments and the relative positions of the simulated towns (including canal-side and inland categories) are visualized in Figure \ref{fig:map_of_canal}.

% \begin{figure}
%     \FIGURE
%     {\includegraphics[width=0.45\textwidth]{ map_visualization.pdf}}
%     {Map of the Grand Canal \label{fig:map_of_canal}}
%     {}
%     \vspace{-1em}
% \end{figure*}

Transport costs are dynamically computed. A global baseline transport cost scales linearly with total population. The canal transport cost is defined as the baseline cost multiplied by $(2 - \text{navigability index})$, ensuring it never falls below the baseline cost. Maritime transport is fixed at one-fifth of the baseline cost, creating an economic incentive to shift to sea routes as the canal degrades. Canal maintenance also directly affects the number of available infrastructure maintenance positions in the labor market.

Social relationships are modeled using two complementary structures. A heterogeneous graph captures identity-based ties such as friendships and professional connections, initialized with a power-law degree distribution $f(x) = c x^{-a}$. A hypergraph structure encodes community-based bonds: kinship clusters are formed via $K$-means clustering on resident coordinates (with minor stochastic noise), while hometown groups are defined by shared town membership. Agent $i$ expresses opinions with a probability of $d_i/d_{max}$, where $d_i$ and $d_{max}$ denote the degree of agent $i$ and the maximum degree of the network, respectively. Both types of graphs are updated during simulation, with edges stochastically rewired every three to five years.

Towns are represented as structured entities containing basic information (name, location, type), a registry of residents, a demographic management module, and a dynamic local employment market. During initialization, the population is distributed proportionally across towns, with any remainder assigned to the primary town. Employment structures vary by town type; for instance, canal-side towns feature a higher density of transport-related jobs.

Each town’s labor market includes six occupational categories: farmer, merchant, rebel, official/soldier, canal maintenance worker, and others. For each category, the system tracks total positions, employed residents (including wage levels), and base salary. Job availability responds dynamically to government investment and economic demand.

\subsection{The Economic Origins of Government}

\textbf{Agents.}
At the core of the simulation are two classes of agents: state officials and residents. State officials are characterized by two attributes: rank (ordinary official, secretary, or senior official) and personality type. 
The memory mechanism is same as that in Canal Decay experiment. Annually, officials collectively decide on two primary actions: (1) investment in canal maintenance, which directly improves navigability and thereby boosts urban incomes, and (2) adjustment of the tax rate, which influences both government revenue and citizen satisfaction.

Residents represent the labor-age population and are categorized as either rural farmers or urban citizens. Each resident possesses a profile that includes remaining lifespan, employment status (self-sufficient or integrated into a polity), occupation, income, satisfaction (a score from 0 to 100), health index (1–5), geographic location $(x, y)$ and associated town, and a randomly assigned pair of personality descriptors. 
% Like officials, residents maintain short-term memory of up to three recent interactions. 
Their annual decisions are also mediated by LLM-based reasoning. 
 Rural farmers choose either to join a polity for stable income at the cost of taxes, or to remain self-sufficient and subject to climatic variability. Urban residents may elect to stay in their current city, migrate to another, or exit the polity entirely. All such choices update individual satisfaction and may trigger spatial relocation based on inter-town connectivity.

\textbf{Environment.}
% The environment comprises a spatial map represented by several complementary data structures. A two-dimensional spatial layout matrix denotes the distribution of canals, while a scalar navigability index (ranging from 0 to 1) quantifies their transport efficiency. Inter-town migration is facilitated by a settlement connectivity graph, and comprehensive town metadata—including nomenclature, spatial coordinates, and settlement categories—is maintained in a centralized attribute registry.
Income is dynamically calculated based on environmental and institutional parameters. Farmers' income is defined as a baseline wage modulated by an environmental impact coefficient, reflecting the susceptibility of agriculture to climatic variance. The income of urban residents is coupled with the state of canal infrastructure; residents in non-riparian settlements receive only 70\% of the income relative to their riverside counterparts. Gross Domestic Product (GDP) is aggregated as the total income of all urban residents, serving as a proxy for economic productivity and state fiscal capacity.

Weather system introduces annual fluctuations via a climate impact factor (0–1), which diminishes agricultural yield and accelerates the natural degradation of canal infrastructure. The decay of the navigability index follows the same mechanism as defined in Canal Decay experiment. Strategic government investment in maintenance can mitigate or reverse this deterioration, establishing a feedback loop between administrative action and economic stability.

Each simulation cycle follows a structured temporal sequence. Initially, the climate impact factor is sampled, and the canal navigability index is updated accordingly. Resident incomes are then recalculated relative to the revised environmental and infrastructural state. Macro-level indicators including GDP, the government budget (derived from taxation), and the population birth rate (modeled as a function of aggregate satisfaction) are updated next. The administrative phase follows, where officials engage in LLM-mediated deliberation to finalize fiscal and investment policies. During the agent behavior phase, individuals concurrently assess their survival status; those whose income falls below a subsistence threshold or who exceed their biological lifespan are removed from the system. Surviving agents make autonomous decisions regarding political affiliation and migration, potentially disseminating public statements through social networks to influence future collective behavior. Finally, each year concludes with a synchronization step: key performance metrics (population, GDP, satisfaction, etc.) are logged, and policy outcomes are archived into the governing agents' long-term memory to inform subsequent decision-making.

% This cyclical process enables the emergence of complex dynamics between environmental constraints, infrastructural investment, fiscal policy, and collective behavior—offering a computational lens on the historical relationship between rivers, agriculture, and the origins of centralized authority.

\subsection{Information Delivery During India’s Demonetization}

The simulated population consists of autonomous agents representing residents, each endowed with basic demographic and psychological attributes including gender, remaining lifespan, a life satisfaction index (on a scale of 0--100), a health index (1--5), a geographical identifier representing a rural settlement, and a personality profile defined by latent behavioral traits. These attributes remain invariant throughout each experimental run but are re-initialized across strategy comparisons to ensure statistical independence.

A central feature of the model is the agents' dual memory system. Short-term memory records information during each step, including official messages (if any) and utterances received through social interactions. At the end of every step, agents use a LLM to synthesize their short-term experiences into a coherent, stable representation of the policy event, which constitutes their long-term "knowledge memory." This distilled knowledge serves as the basis for subsequent decision-making and is directly queried during post-intervention assessments. Critically, all memory states and behavioral counters are reset before the initiation of each dissemination strategy to maintain experimental purity.

Agents make decisions about whether to discuss the policy based on a context-aware planning process also mediated by an LLM. Their decision inputs include their personal state (e.g., personality, satisfaction), the content of any official message received, a "public notice" field that encodes meta-information about who else received the message (the key variable distinguishing common knowledge conditions), and their current knowledge memory. If an agent decides to speak, it generates natural language commentary that is propagated through its social network connections.

The social network structure is fixed throughout each simulation run to isolate the effects of information strategy from network evolution. It combines heterogeneous pairwise ties with hyperedges representing group affiliation. Pairwise ties are generated via a power-law distribution to emulate real-world networks with hubs. 
Hyperedges are constructed using K-means clustering on agent attributes and spatial proximity, capturing dense local communities. When an agent speaks, its message is delivered to all neighbors in this combined graph, triggering memory updates in recipients and enabling multi-step information diffusion.

Four distinct dissemination strategies are implemented, forming a 2×2 factorial design: two delivery modes (Broadcast vs. Seeding) crossed with two knowledge conditions (Common Knowledge vs. No Common Knowledge). In Broadcast (BC) strategies, every resident receives the full official message. In Seeding (S) strategies, only a small set of "seed" agents are given the message. These agents are selected using a composite centrality measure that combines degree, betweenness, and closeness.The Common Knowledge (CK) manipulation concerns what agents are told about others’ access to information. Under BC\_CK, all agents receive the public notice: "You know that all villagers received the government message, and all villagers know that you received it." Under BC\_NCK, no such notice is provided. In S\_CK, seed agents are told: "Only some villagers received the message, and everyone knows you are one of them," while non-seeds hear: "Only some villagers received the message." In S\_NCK, all agents receive only the latter statement, obscuring who actually possesses the information.

Each strategy is executed over a fixed simulation period. Within each time step, the following sequence occurs: (i) the designated message and public notice are delivered according to the active strategy; (ii) every agent processes this input and decides whether to initiate a discussion; (iii) any generated speech propagates through the static social network, updating neighbors’ short-term memories; and (iv) all agents summarize their accumulated inputs into updated long-term knowledge representations.

After the final time step of a given strategy, two standardized evaluations are conducted. First, a knowledge survey queries every agent on factual aspects of the demonetization policy; responses are parsed by an LLM to compute per-question and aggregate accuracy scores. Second, an incentivized choice task presents agents with a binary economic decision analogous to the field experiment: choosing between holding a now-invalid 500-rupee note (which can still be deposited in banks) or accepting a voucher redeemable in a few days for either 200 rupees or an equivalent value in lentils. The proportion selecting the rational option (the bankable note) serves as a behavioral proxy for policy comprehension and trust.
Besides, the total volume of spontaneous policy-related discussions across the network during the intervention period is also recorded as a measure of social learning intensity.

\section{Auto-simulation Details}
\label{appendix:Technical_Details_Auto-simulation}

The auto-simulation pipeline in \textit{Eco3S} relies on a set of specialized AI agents and structured data artifacts to enable end-to-end automation.
%  The complete operational sequence—from requirement parsing and code synthesis to iterative optimization—is illustrated in the sequence diagram in Figure \ref{fig:auto_simulation_framework}, which details the collaborative interactions between the four core AI agents. 
 The following elaborates on the internal responsibilities of each agent and the semantic roles of the artifacts they produce or consume. 

\textbf{Agent Responsibilities. }
\textbf{(1) ProjectMasterAgent} serves as the central orchestrator. It parses user input into a structured requirement containing simulation name, description, and type (e.g., decision-based or survey-based). It then sequentially invokes other agents, manages execution of the generated simulation, captures runtime errors, and coordinates iterative refinement loops. In interactive mode, it pauses after design and coding phases to solicit human feedback.
\textbf{(2) SimArchitectAgent} acts as the system architect. Given the structured requirement, it generates a high-level design specification that defines agent types (e.g., residents, officials), environmental components (e.g., climate, geography, job market, heterogeneous information networks), interaction protocols (e.g., taxation, migration, rebellion triggers), and decision-making modes (individual vs. group consensus). It also produces a module activation plan indicating which functional subsystems should be included.
\textbf{(3) CodeArchitectAgent} functions as a full-stack developer. It synthesizes executable Python code for the simulator core and program entrypoint based on the design specification and module plan. It also generates detailed runtime configuration files that govern simulation parameters, initial agent states, and LLM prompt templates for behavioral control. Crucially, it supports two forms of automated repair: (1) fixing runtime crashes by analyzing error traces and modifying code/configurations; (2) adjusting configurations in response to diagnostic feedback to improve result quality.
\textbf{(4) ResearchAnalystAgent} serves as the evaluation expert. After a successful simulation run, it compares the output data against the expected behaviors described in the original design. If outcomes deviate (e.g., rebellion rates are too low), it produces a diagnostic report that identifies the most likely faulty components (e.g., overly generous job market settings) and suggests concrete, actionable modifications to relevant configurations or agent prompts.

\textbf{Artifact Types and Functions.}
The agents communicate through a set of well-defined artifact types: 
\textbf{(1) Structured Requirement}: A machine-readable representation of the user’s natural language input, enabling deterministic downstream processing. 
\textbf{(2) System Design Specification}: A comprehensive document detailing the simulated entities, environment dynamics, interaction rules, and decision mechanisms. This serves as the single source of truth for implementation. 
\textbf{(3) Module Activation Plan}: A declarative configuration that specifies which functional modules (e.g., job market, climate system) are enabled, acting as a high-level switchboard for code generation. 
\textbf{(4) Executable Source Code}: The generated Python implementation, including the simulator logic and execution entrypoint. This code is fully functional without manual intervention. 
\textbf{(5) Runtime Configuration Set}: A collection of structured files that define numerical parameters, initial states, and LLM prompts for agent behaviors. These decouple policy logic from program structure, enabling easy tuning and counterfactual experiments. 
\textbf{(6) Simulation Output Data}: Structured records (e.g., time-series of agent actions, environmental states, macro indicators) produced during execution, used for evaluation. 
\textbf{(7) Diagnostic Report}: A machine-readable analysis that links observed outcome deviations to specific configuration or behavioral flaws, providing precise guidance for iterative refinement.

\textbf{Iterative Refinement and Error Handling.}
The system supports two closed-loop mechanisms:
\textbf{(1) Runtime Error Recovery}: When simulation execution fails, the ProjectMasterAgent captures the error trace and instructs the CodeArchitectAgent to analyze the root cause. The latter identifies relevant code and configuration components, then applies targeted fixes. This loop repeats until the simulation runs successfully.
\textbf{(2) Result-Driven Optimization}: If the ResearchAnalystAgent deems results unsatisfactory, it generates a diagnostic report. The ProjectMasterAgent initiates a refinement loop: the CodeArchitectAgent adjusts the runtime configuration (or prompts) based on the report, and the simulation is re-run. This cycle continues until results meet quality thresholds or a maximum iteration count is reached.

\section{More Experiments on Auto-simulation}
\label{app:More Experiments on Auto-simulation}

This section validates the generalizability of the \textit{Eco3S} auto-simulation pipeline through four classical economic cases. The full system prompts and hyperparameter configurations will be provided through an anonymized artifact repository.
% The source code is publicly available at 

\subsection{Herding Effect in Financial Markets}
The \textbf{herding effect simulation} models investor behavior where agents make buy, sell, or hold decisions based on market signals and social interactions. Core mechanisms include a dynamic social network for information diffusion and a feedback loop between trading and asset prices.

\begin{figure}[!ht]
  \centering
    \includegraphics[width=0.49\columnwidth]{ 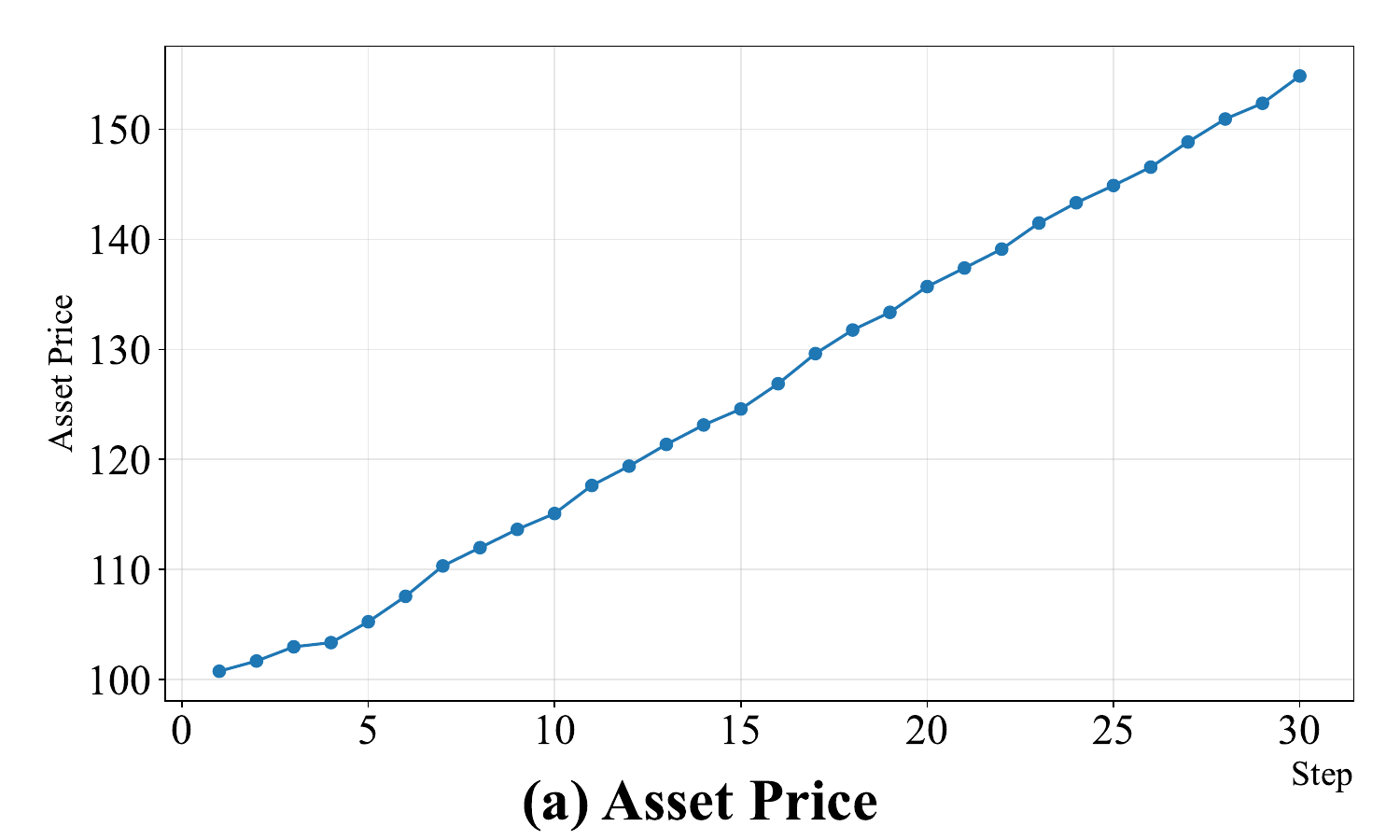}
    \includegraphics[width=0.49\columnwidth]{ 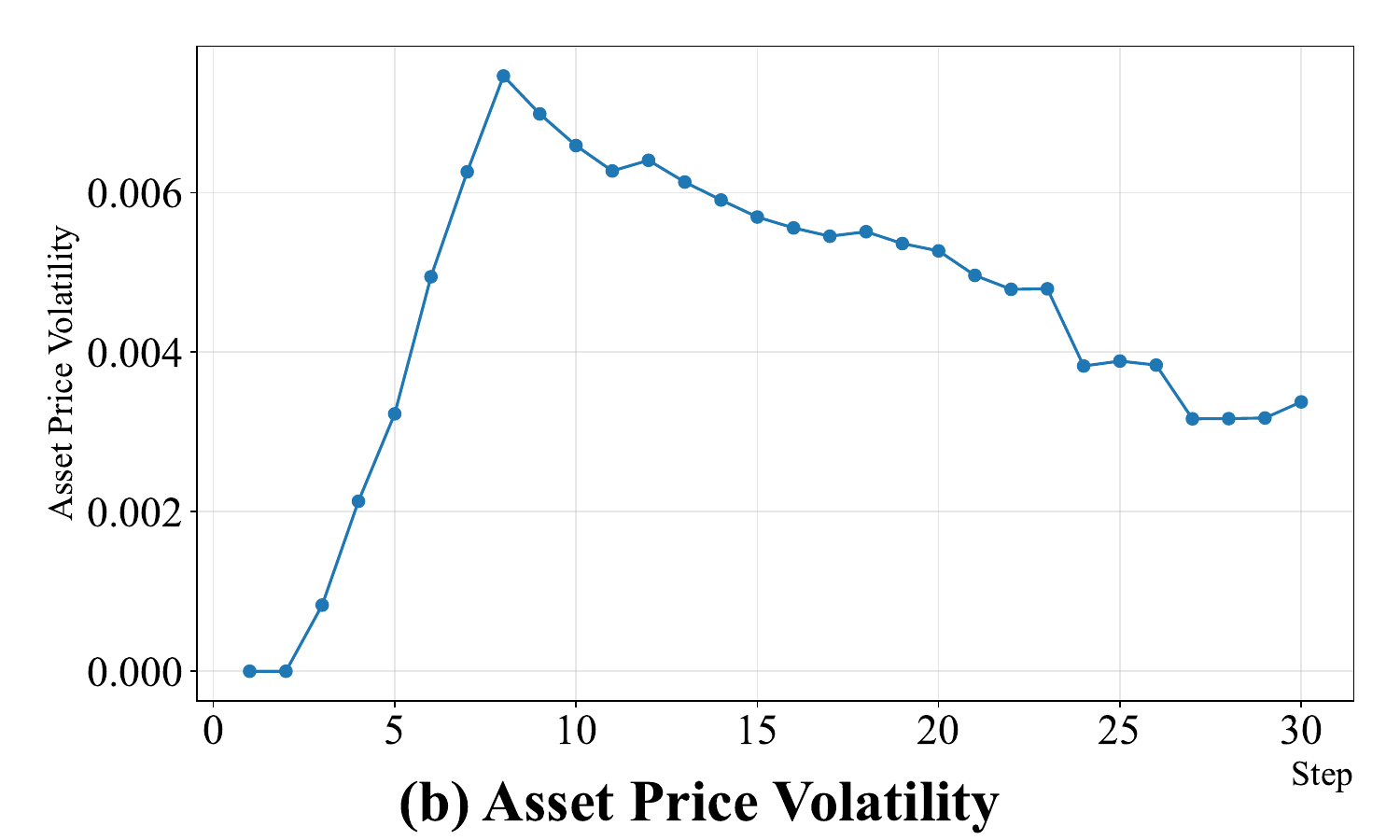}\\[1ex]
    \includegraphics[width=0.49\columnwidth]{ 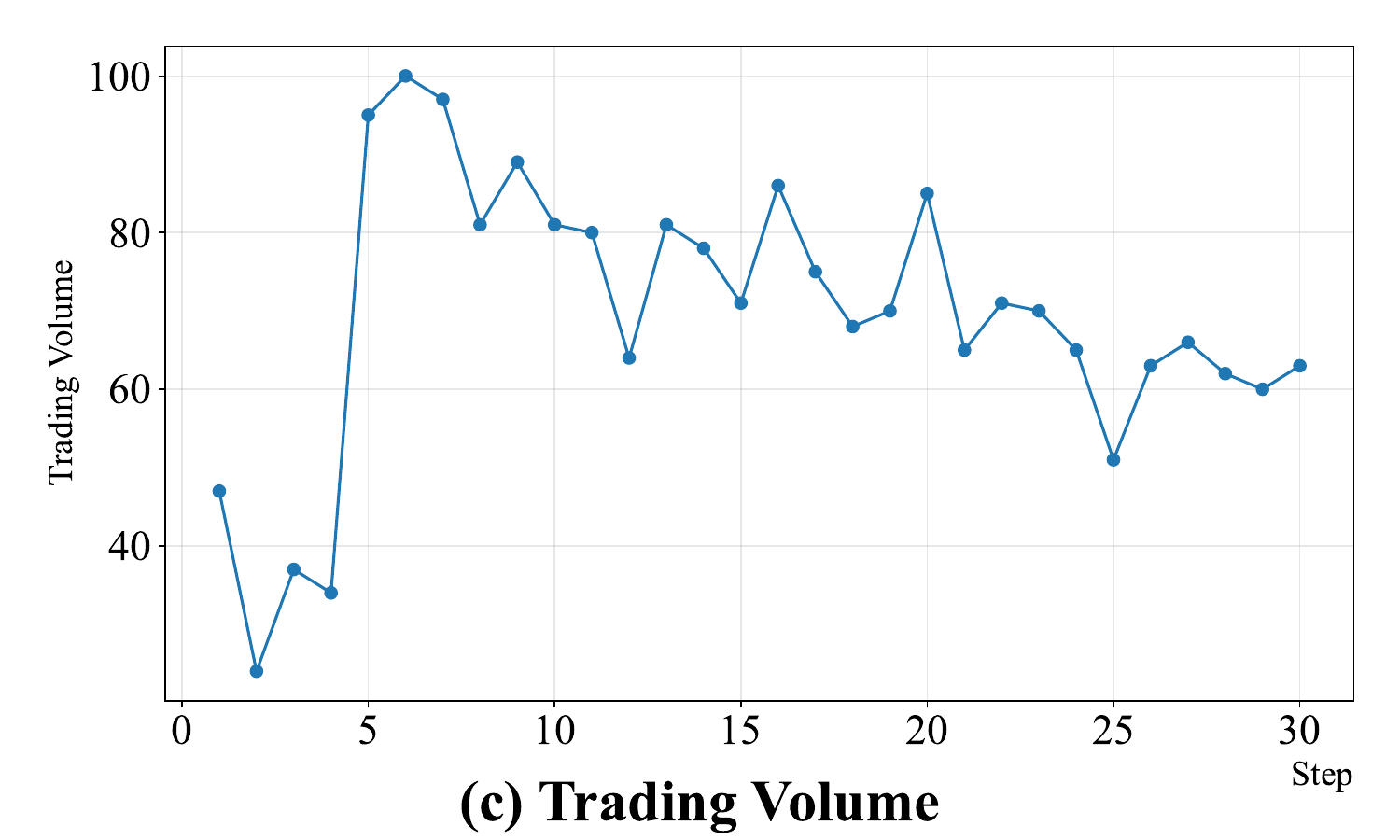}
    \includegraphics[width=0.49\columnwidth]{ 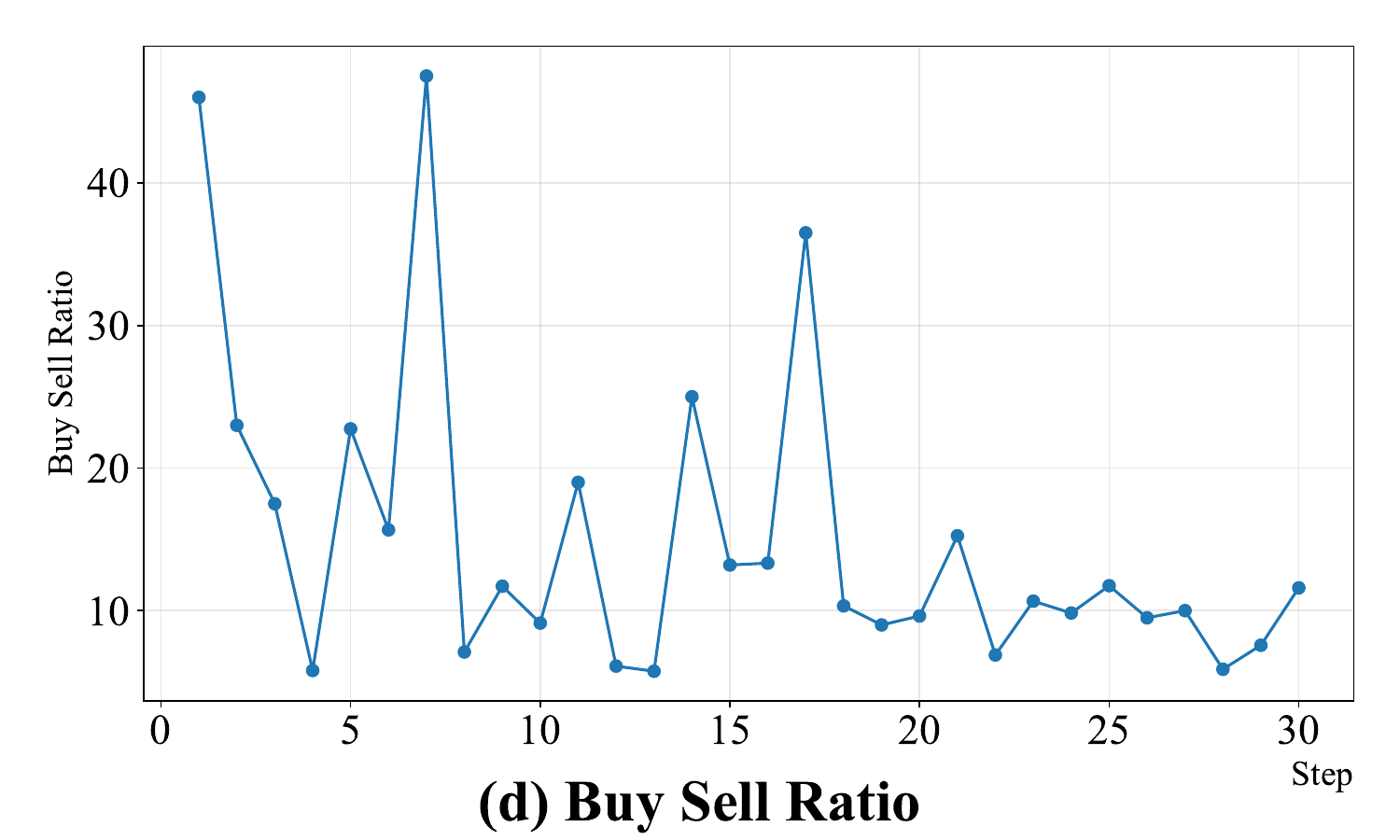}\\[1ex]
    \includegraphics[width=0.49\columnwidth]{ 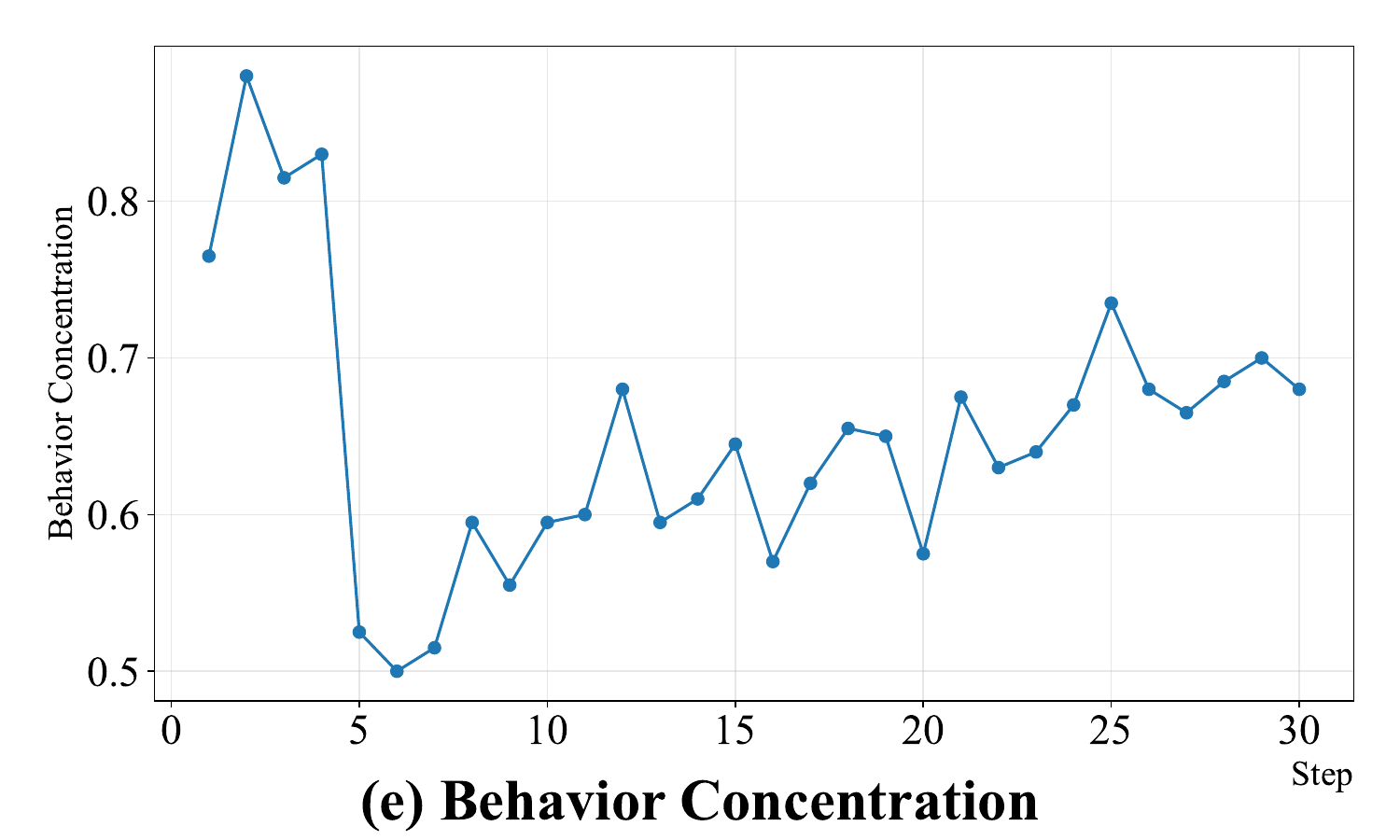}
    \includegraphics[width=0.49\columnwidth]{ 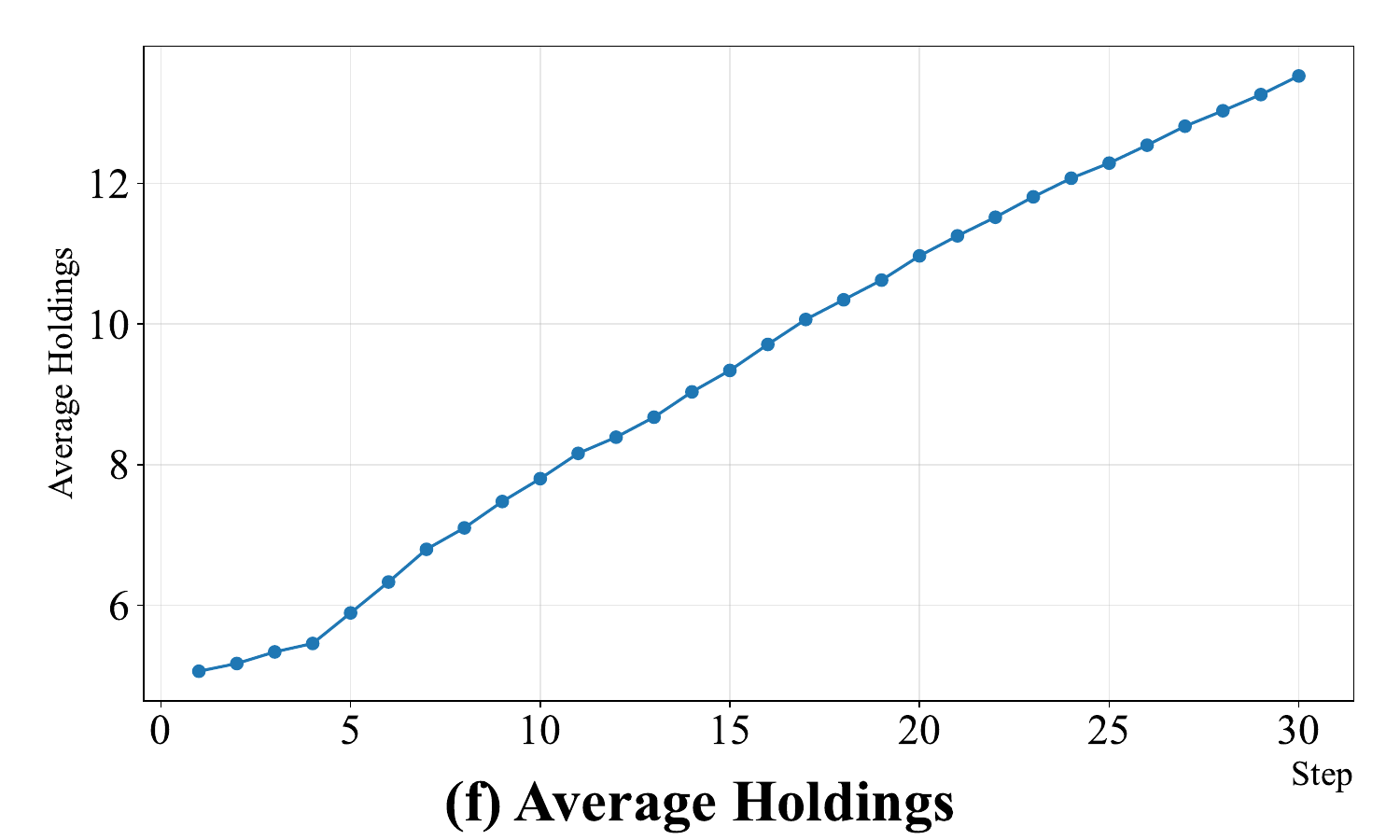}
  \caption{Results of Herding Effect Experiment.}
  \label{fig:herding_effect}
\end{figure}

As shown in Figure \ref{fig:herding_effect}, the simulation manifests clear herding behavior. Asset prices rose significantly over the simulation period, with price volatility reaching 0.0074. The buy/sell ratio was heavily skewed toward buying (5.75 to 47.50), and decision homogeneity averaged 0.65. Average holdings increased by over 167\%. The results successfully replicate theoretical features of financial herding, including price amplification and behavioral synchronization.

\subsection{Custom Satisfaction and Loyalty}
This experiment investigates the dynamic interplay between service quality, word-of-mouth (WOM), and purchasing behavior. Resident agents interact within a social network where positive or negative experiences propagate to connected peers, adjusting collective sentiment. Each simulation cycle proceeds in four phases: (1) computing macro indicators; (2) making parallel decisions on employment and consumption; (3) executing decisions to update income and social expressions; and (4) logging metrics.

\begin{figure}[!ht]
  \centering
  \includegraphics[width=0.49\columnwidth]{ 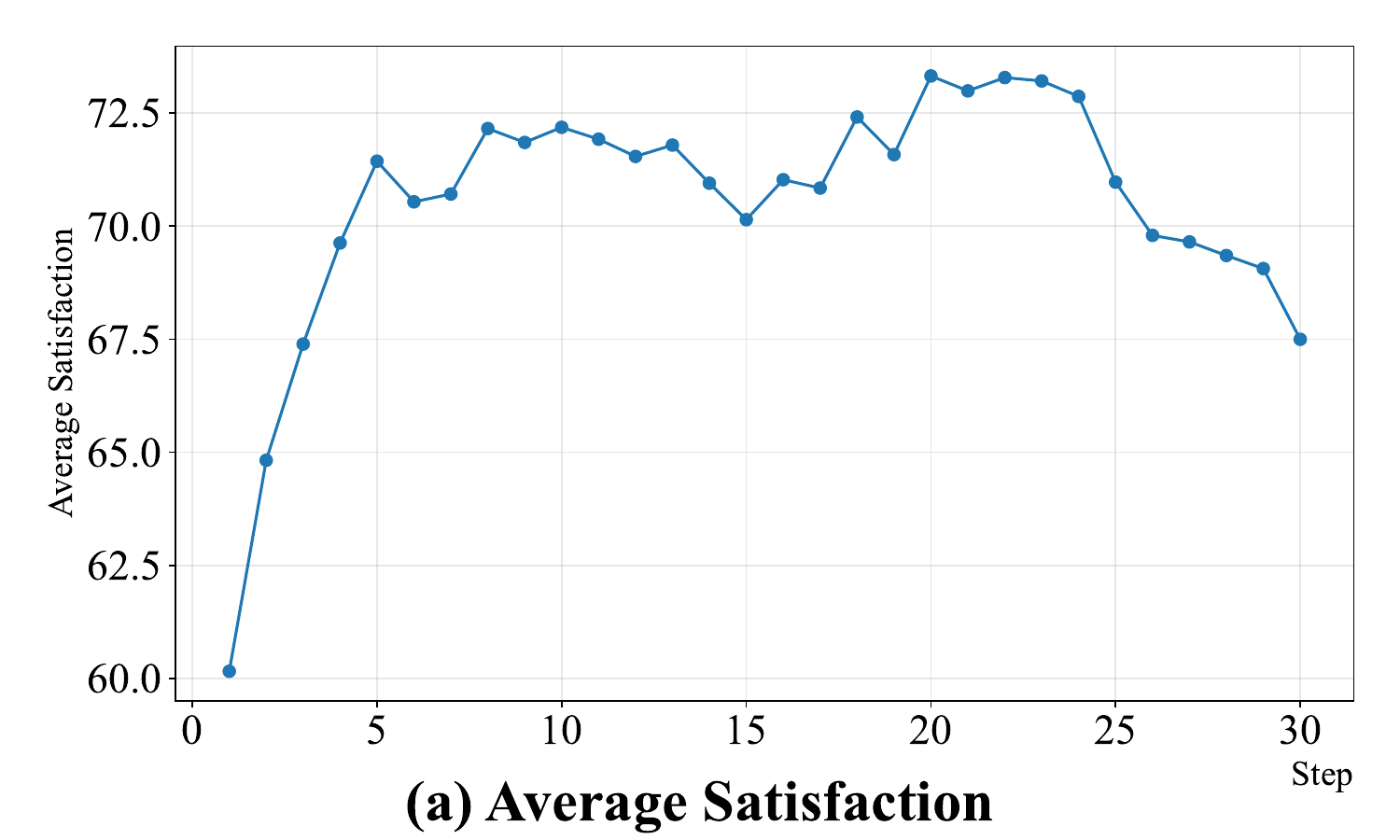}
  \includegraphics[width=0.49\columnwidth]{ 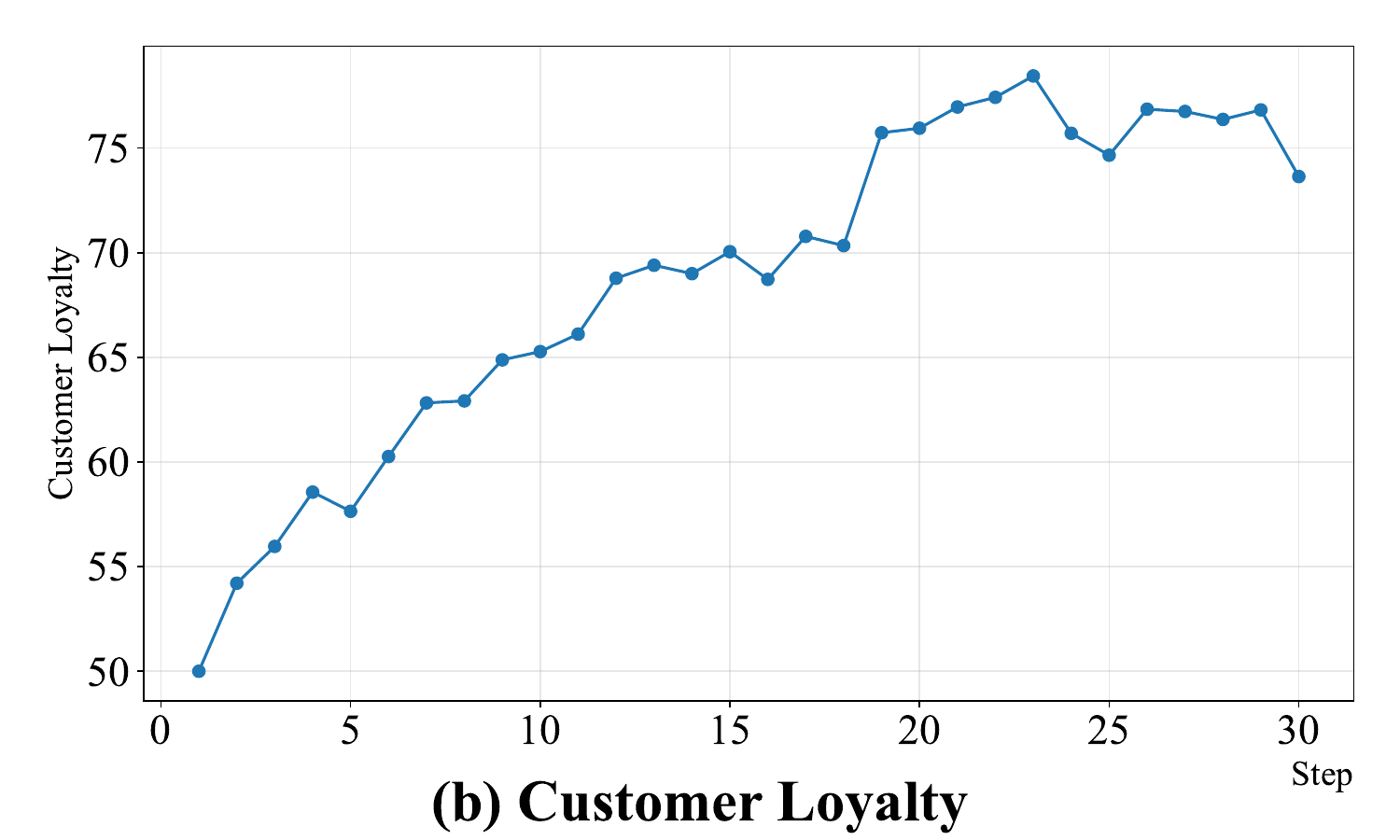} \\[1ex]
  \includegraphics[width=0.49\columnwidth]{ 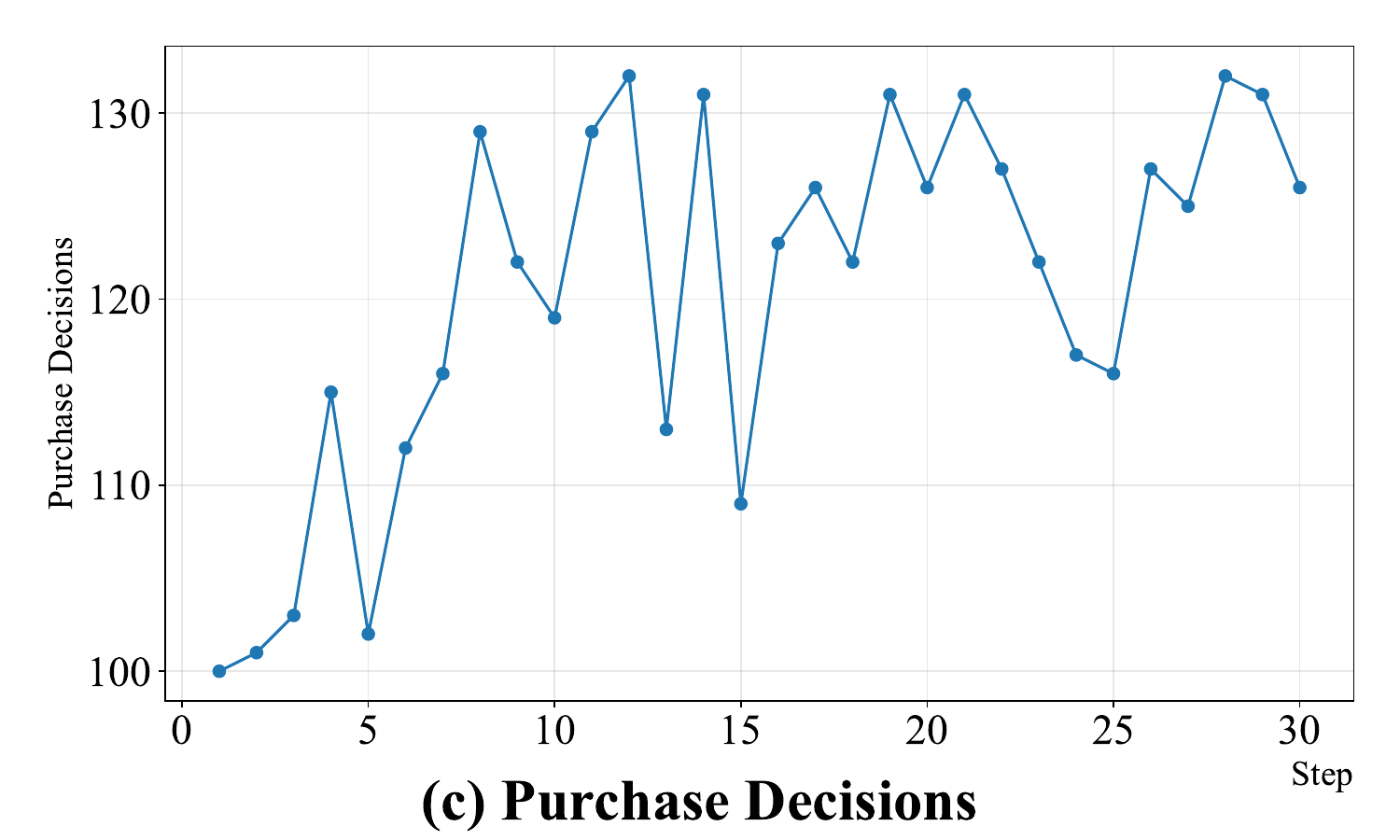}
  \includegraphics[width=0.49\columnwidth]{ 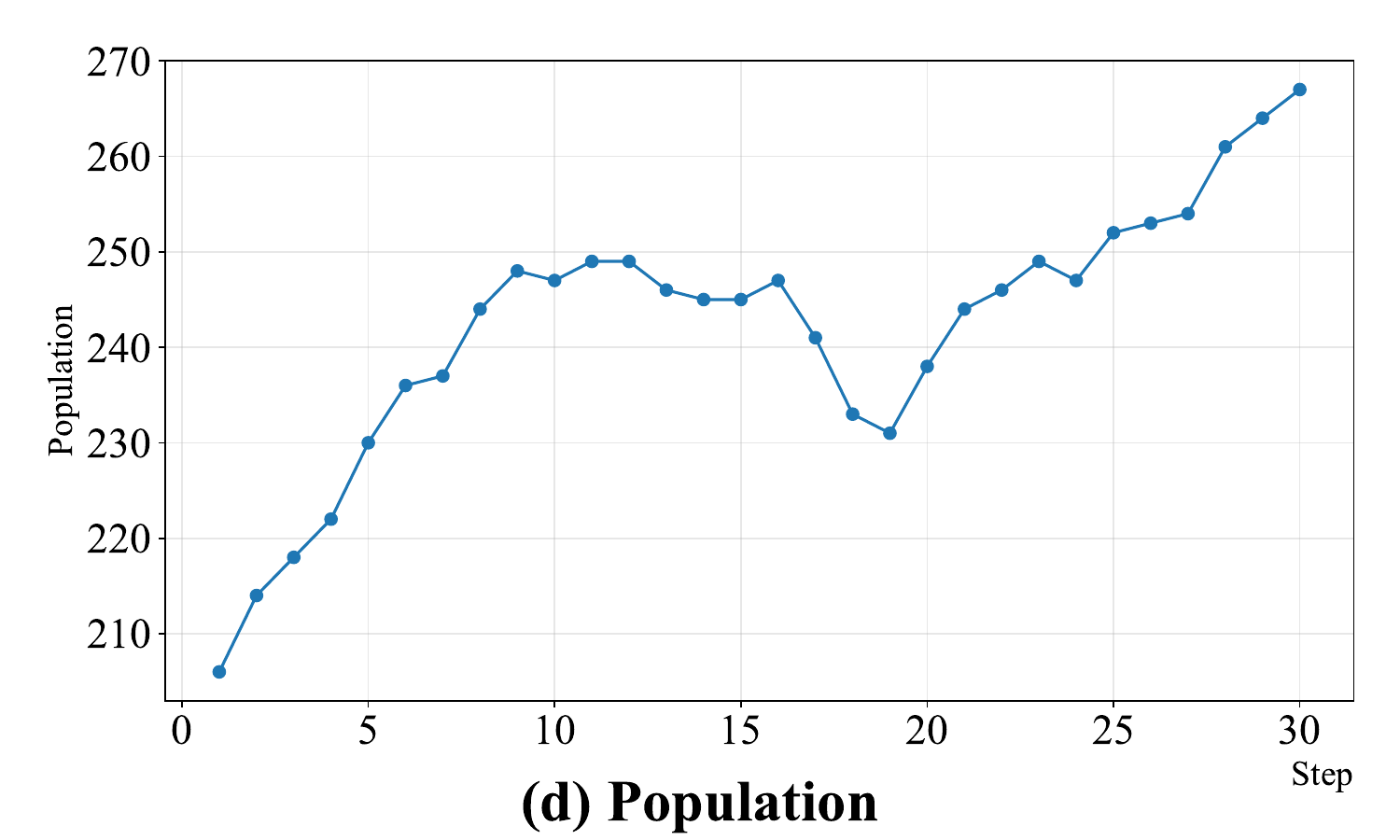}
  \caption{Results of Custom Satisfaction Experiment.}
  \label{fig:custom_satisfaction}
\end{figure}

Results in Figure \ref{fig:custom_satisfaction} show that average satisfaction rose from 60.16 to 73.32, and loyalty increased from 50.00\% to 75.94\%. A decline after step 24 suggests population growth outpaced service capacity. Notably, loyalty exhibits hysteresis, declining slower than satisfaction, which confirms the "stickiness" of established customer relationships. The simulation validates that effective word-of-mouth mechanisms and sustained quality drive a virtuous cycle of satisfaction and consumption.

\subsection{Asset Bubble Formation}
This model replicates a classic asset market \citep{smith1988bubbles} where a synthetic asset pays fixed dividends and expires after $T$ periods. Fundamental value is defined as $V_t = D \times (T - t)$. Trader agents with diverse risk preferences and strategies (momentum or value-based) interact via a continuous double auction mechanism.

\begin{figure}[!ht]
  \centering
  \includegraphics[width=0.49\columnwidth]{ 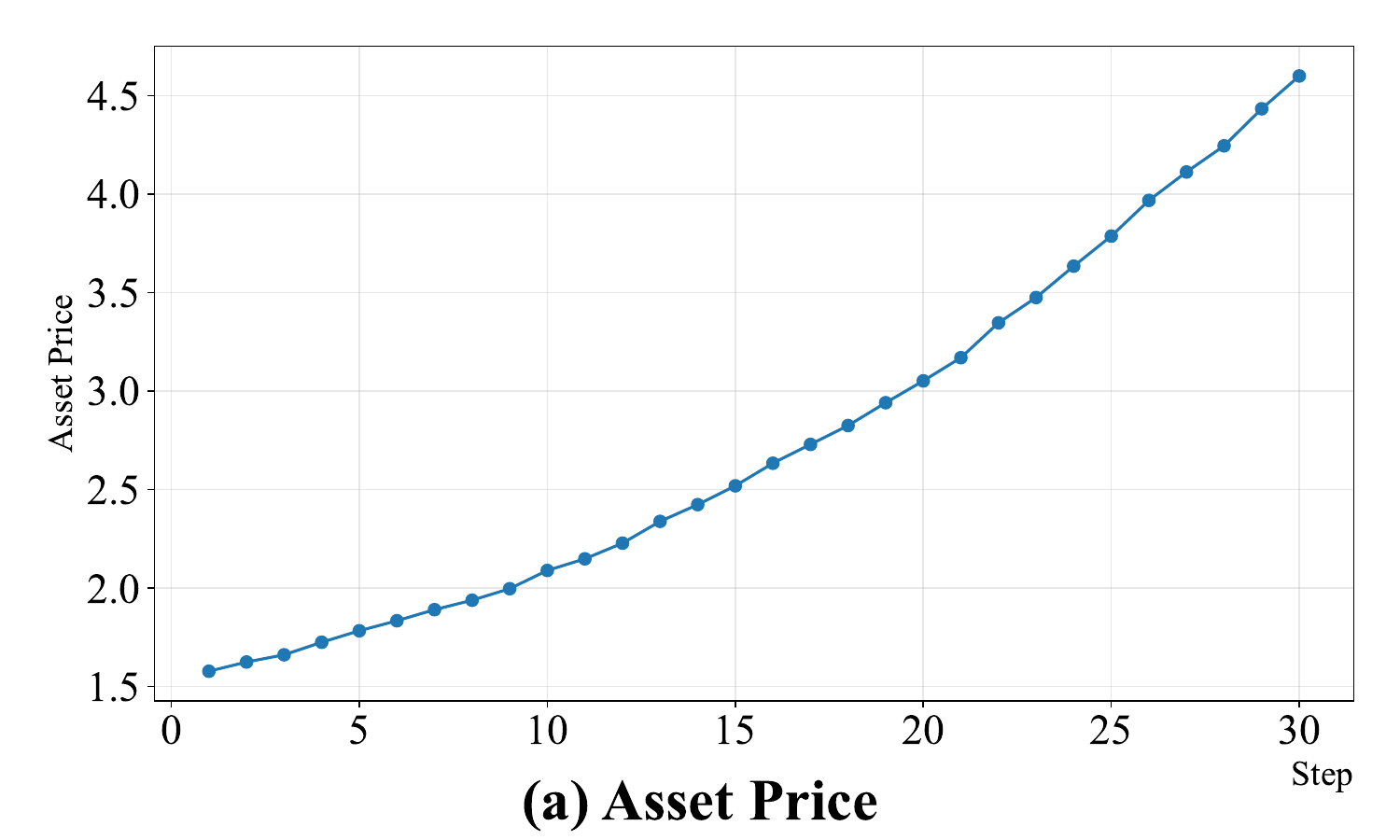}
  \includegraphics[width=0.49\columnwidth]{ 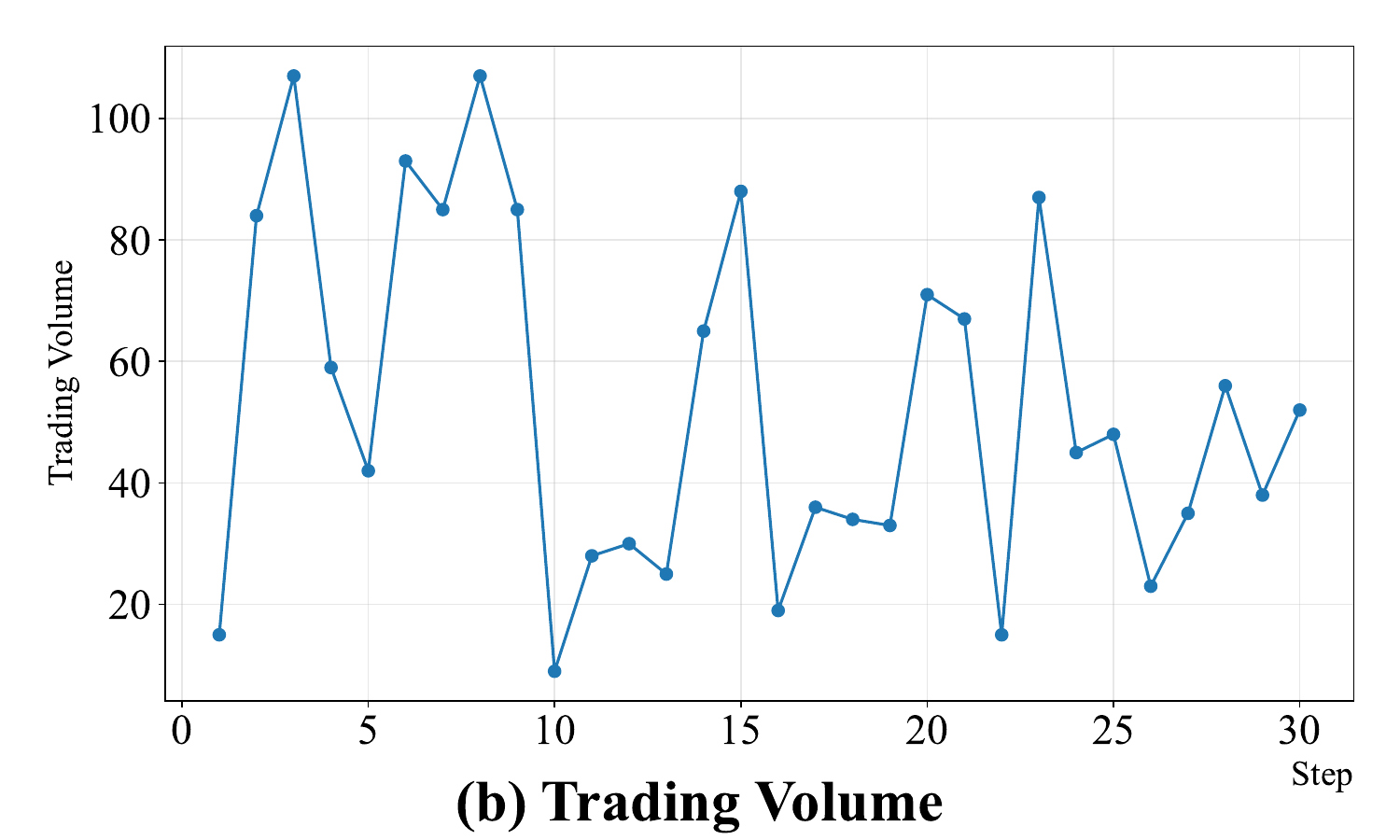} \\[1ex]
  \includegraphics[width=0.49\columnwidth]{ 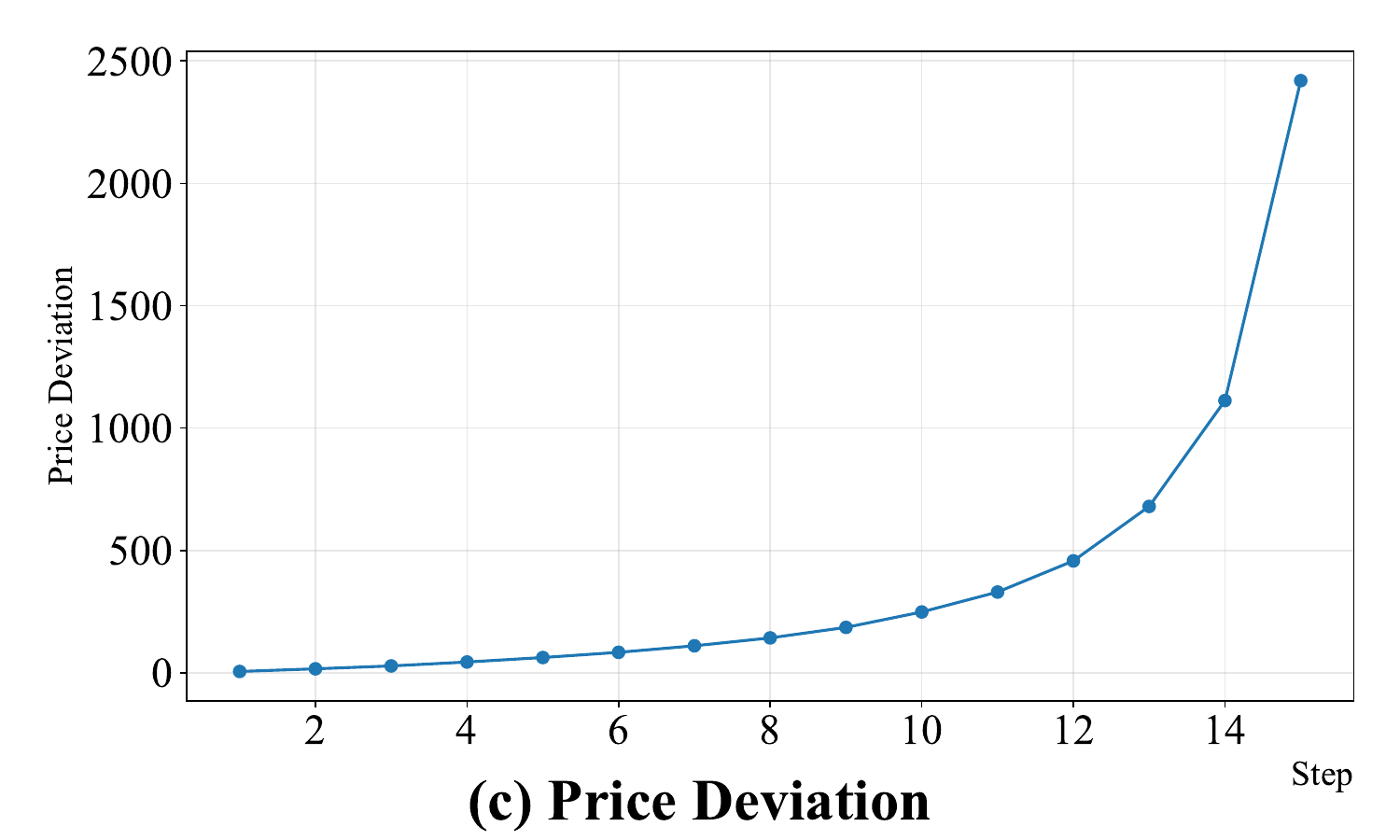}
  \includegraphics[width=0.49\columnwidth]{ 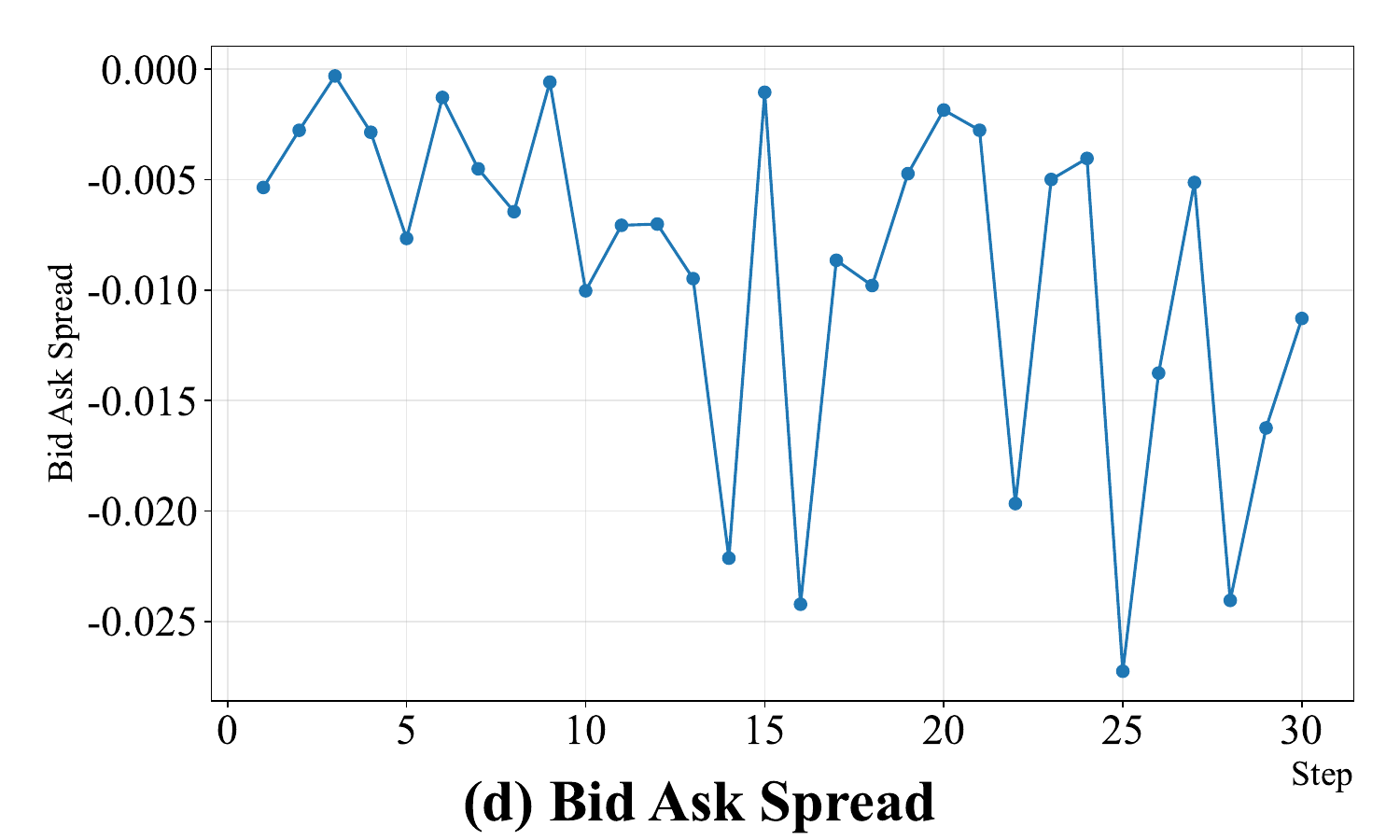}
  \caption{Results of Asset Bubble Experiment.}
  \label{fig:asset_bubble}
\end{figure}

As shown in Figure \ref{fig:asset_bubble}, the asset price rose from 1.58 to 4.60 (a 191\% increase) despite fundamental value declining to zero. Price deviation from fundamentals peaked at over 2400\% in period 15. Trading volume was volatile with notable spikes, and the bid-ask spread narrowed over time. These results demonstrate that LLM-driven agents can endogenously generate asset bubbles through speculative trading and decision feedback.

\subsection{Schelling Segregation Model}
This study reconstructs \citet{schelling1971dynamic}'s model on a grid to observe macro-level spatial segregation. Agents relocate if neighbor similarity falls below a threshold. Segregation is measured by the Index of Dissimilarity ($D$):
\begin{equation}
D = \frac{1}{2} \sum_{i=1}^{n} \left| \frac{a_i}{A} - \frac{b_i}{B} \right|.
\end{equation}

\begin{figure}[!ht]
  \centering
    \includegraphics[width=0.49\columnwidth]{ 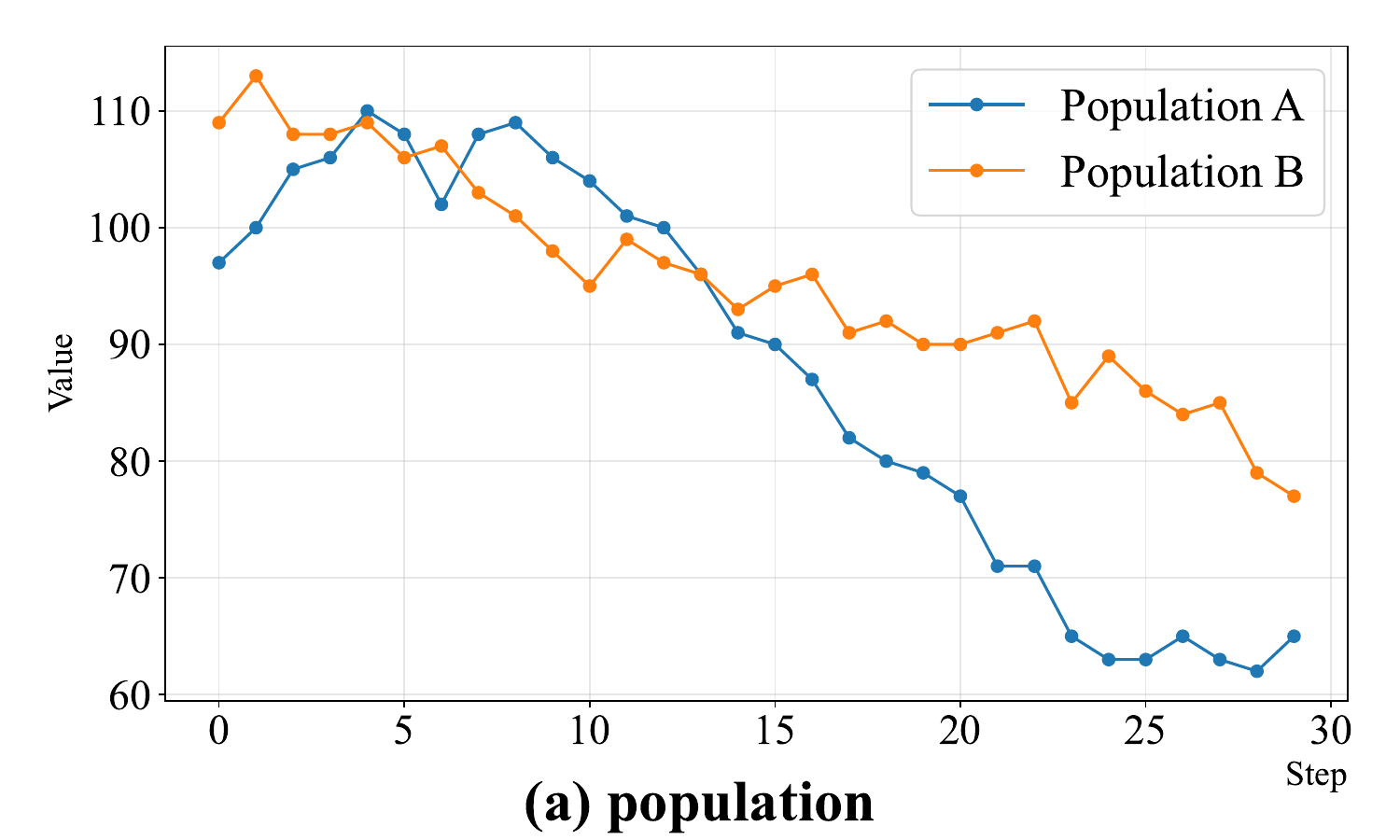}
    \includegraphics[width=0.49\columnwidth]{ 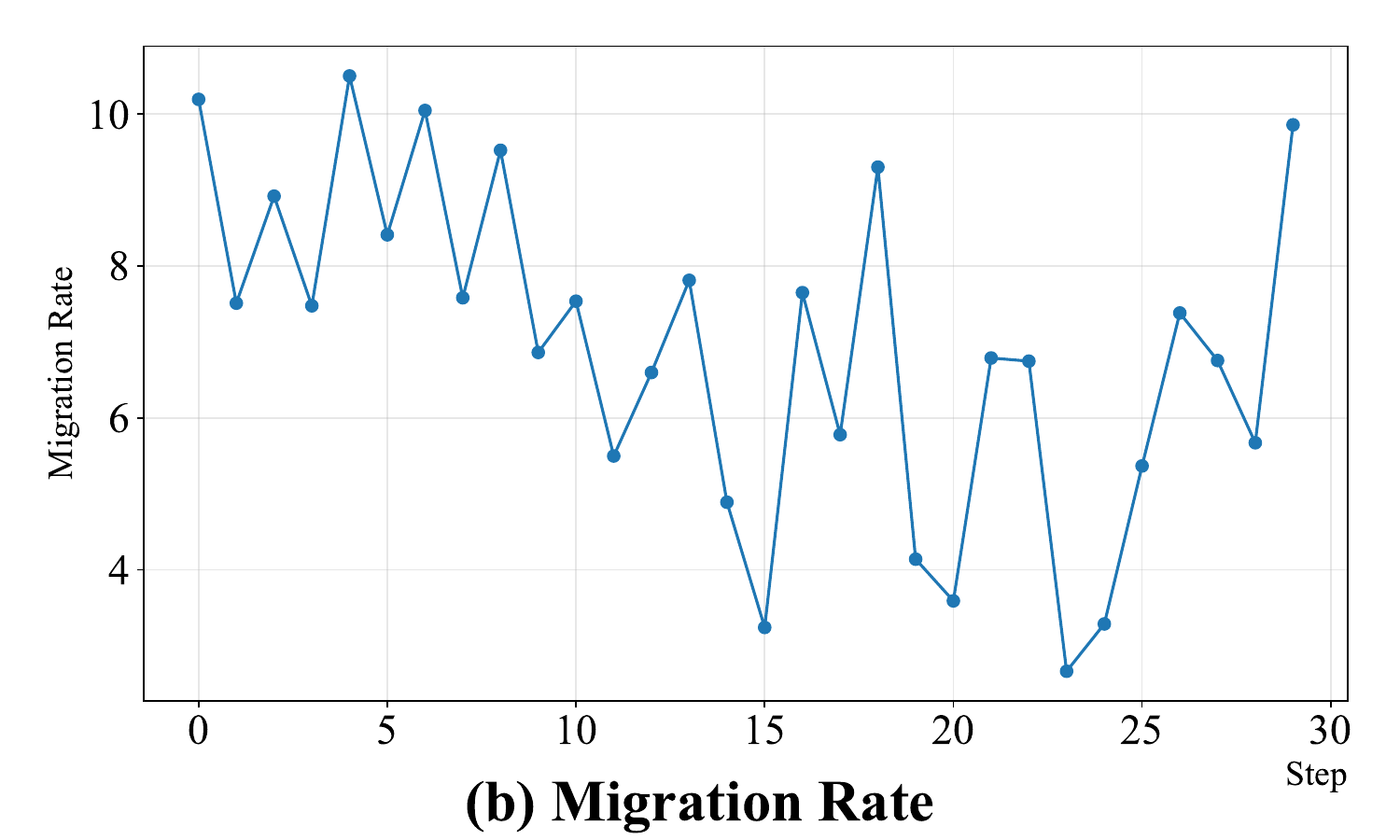} \\[1ex]
    \includegraphics[width=0.49\columnwidth]{ 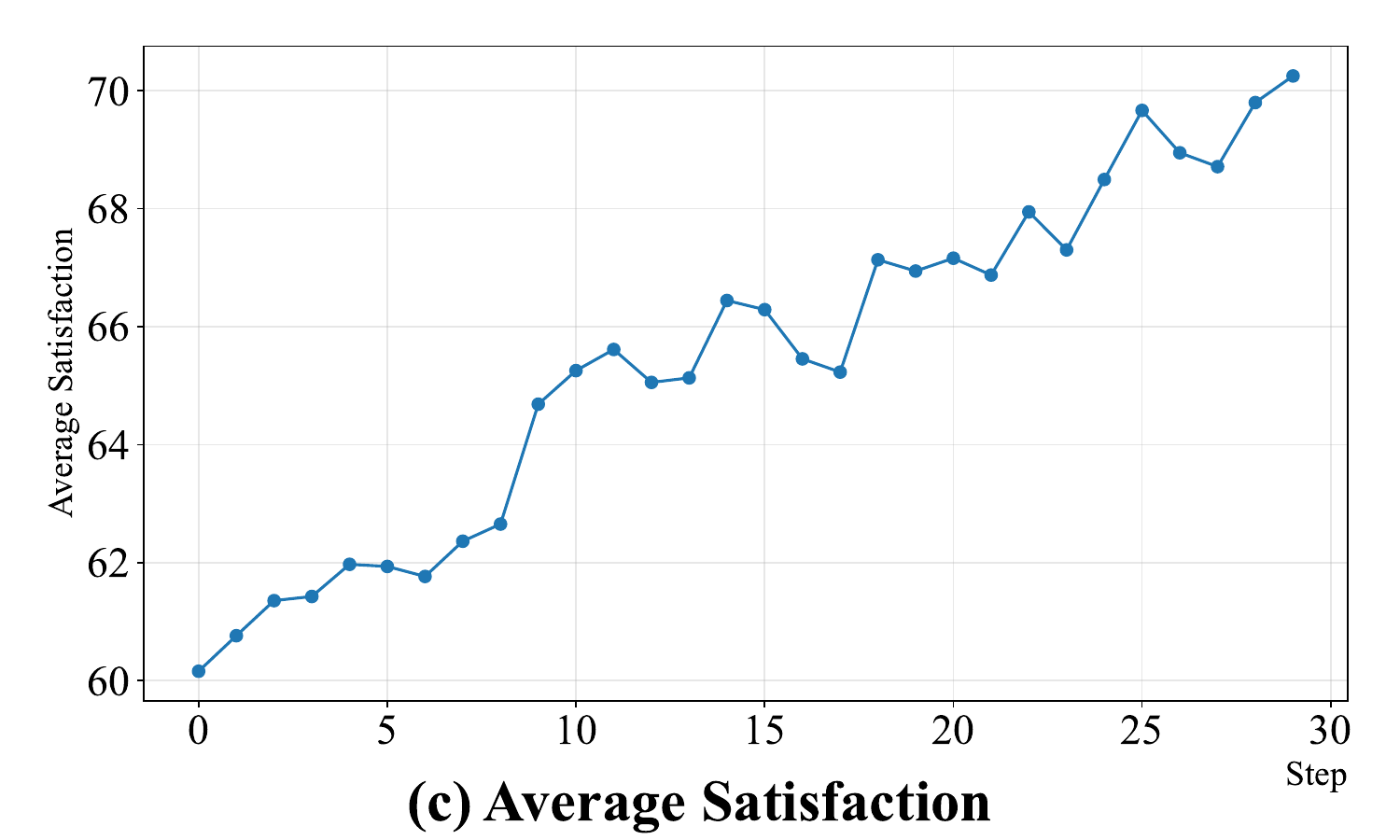}
    \includegraphics[width=0.49\columnwidth]{ 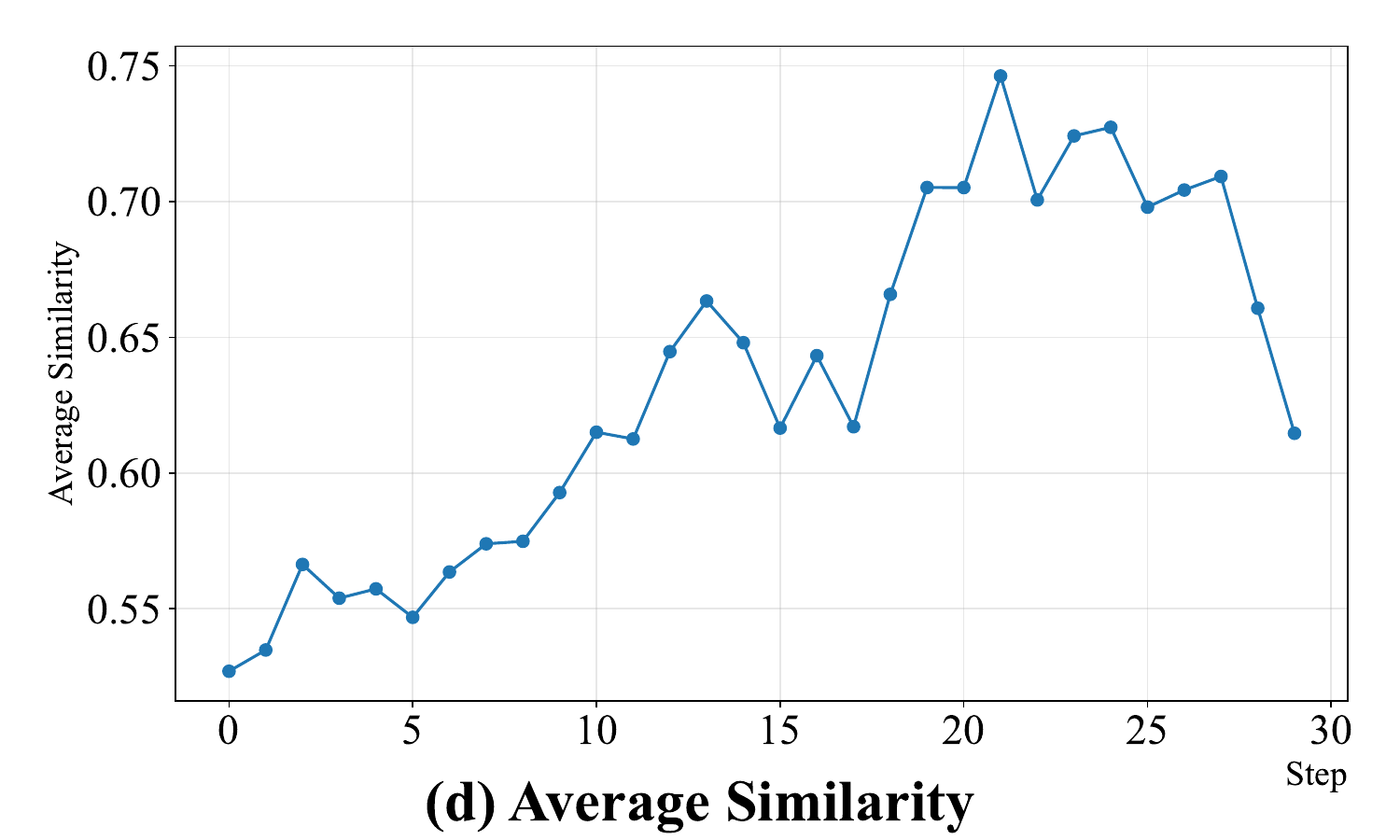} \\[1ex]
    \includegraphics[width=0.49\columnwidth]{ 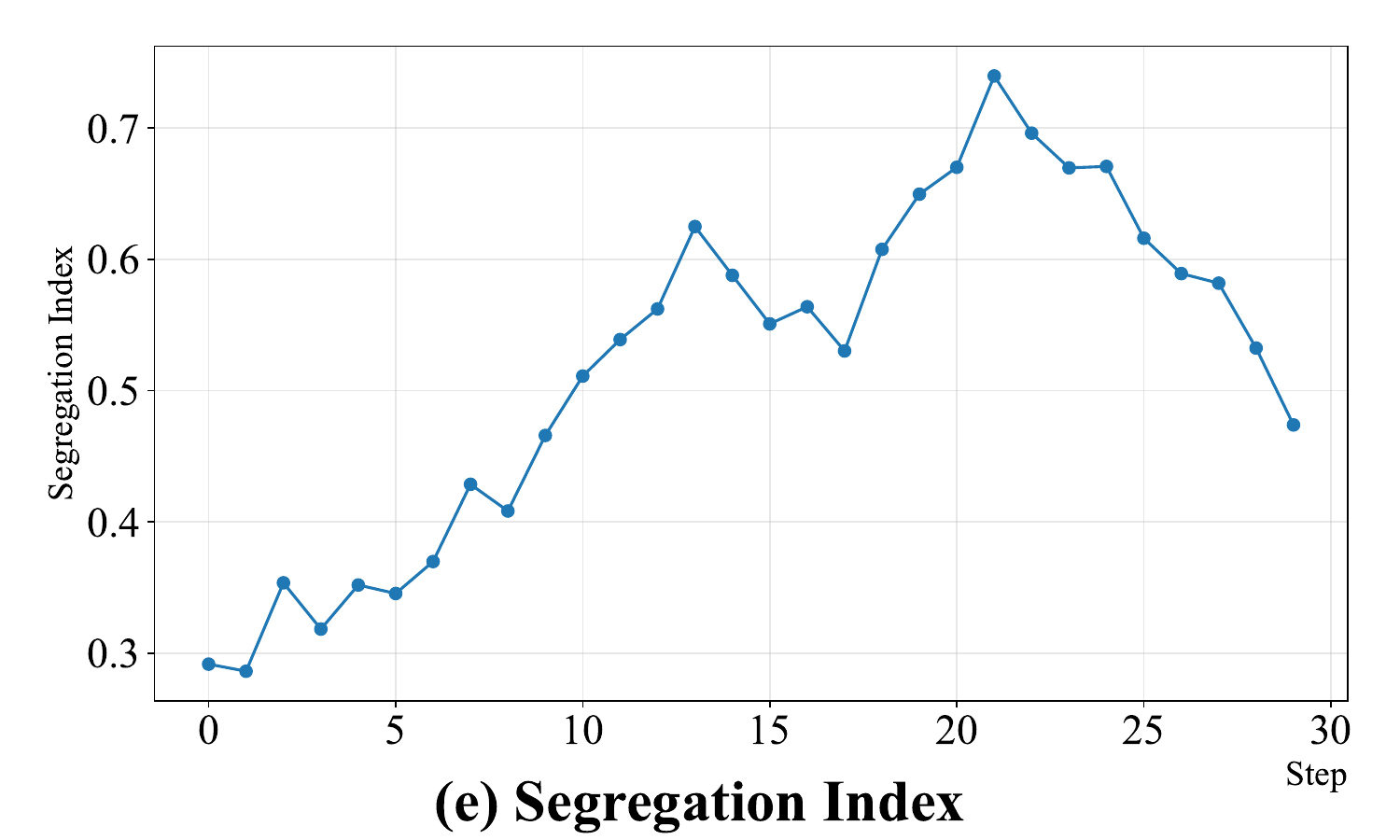}
    \includegraphics[width=0.49\columnwidth]{ 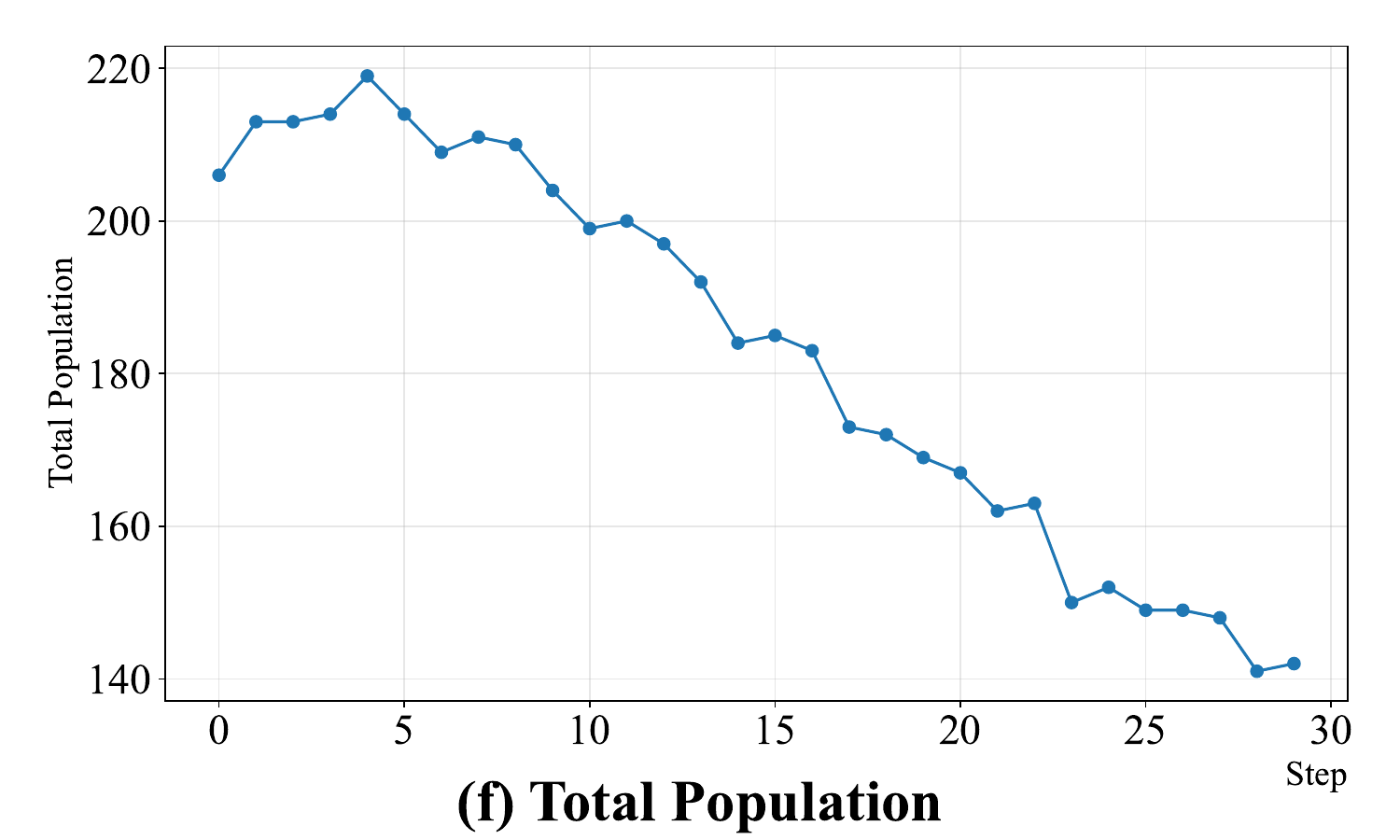}
  \caption{Results of Schelling Model Experiment.}
  \label{fig:Schelling_model}
\end{figure}

Results in Figure \ref{fig:Schelling_model} show the segregation index rising from 0.29 to 0.74, while neighbor similarity increased from 0.53 to 0.75. Average satisfaction improved from 60.16 to 66.87, illustrating the "Schelling paradox" where individuals become happier as society becomes more segregated. The experiment confirms that mild micro-level preferences can lead to pronounced macro-level segregation patterns through self-organizing processes.

\section{Robustness Analysis}
\label{app:api_compatibility}
To verify framework robustness and flexibility, we evaluated four LLMs as cognitive engines: GPT-3.5 Turbo, GPT-4o, DeepSeek, Qwen-235B-A22B, and Qwen3-8B. All trials were conducted with $N=2000$ residents over $T=15$ steps, maintaining identical economic structures (jobs, wages, market rules) and social dynamics (interaction matrices). We stochastically assigned unique random seeds to each trial to ensure that results reflect emergent behavior under non-deterministic conditions rather than simple replication.

\begin{figure}[!ht]
  \centering
    \includegraphics[width=0.49\columnwidth]{ 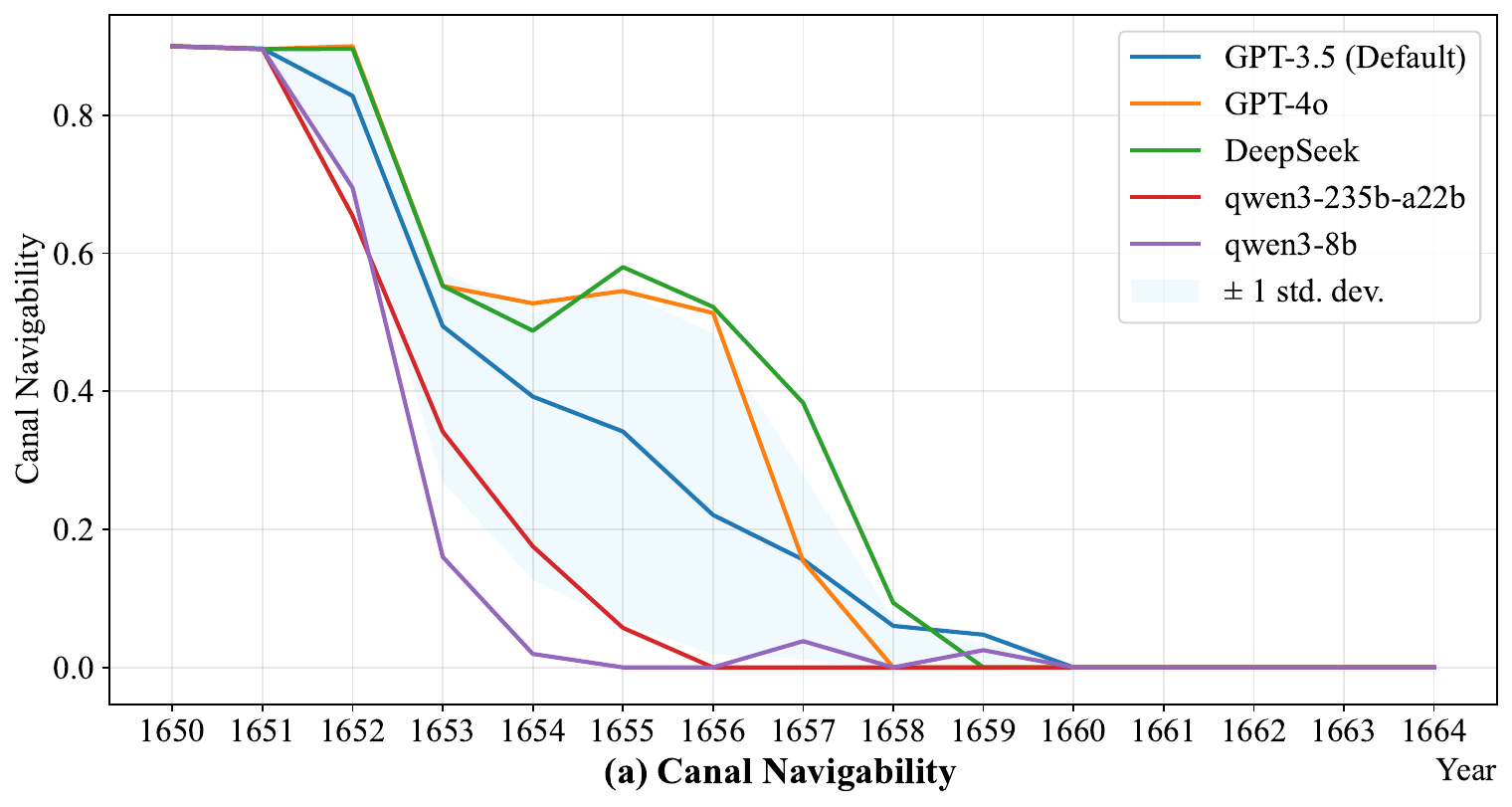}
    \includegraphics[width=0.49\columnwidth]{ 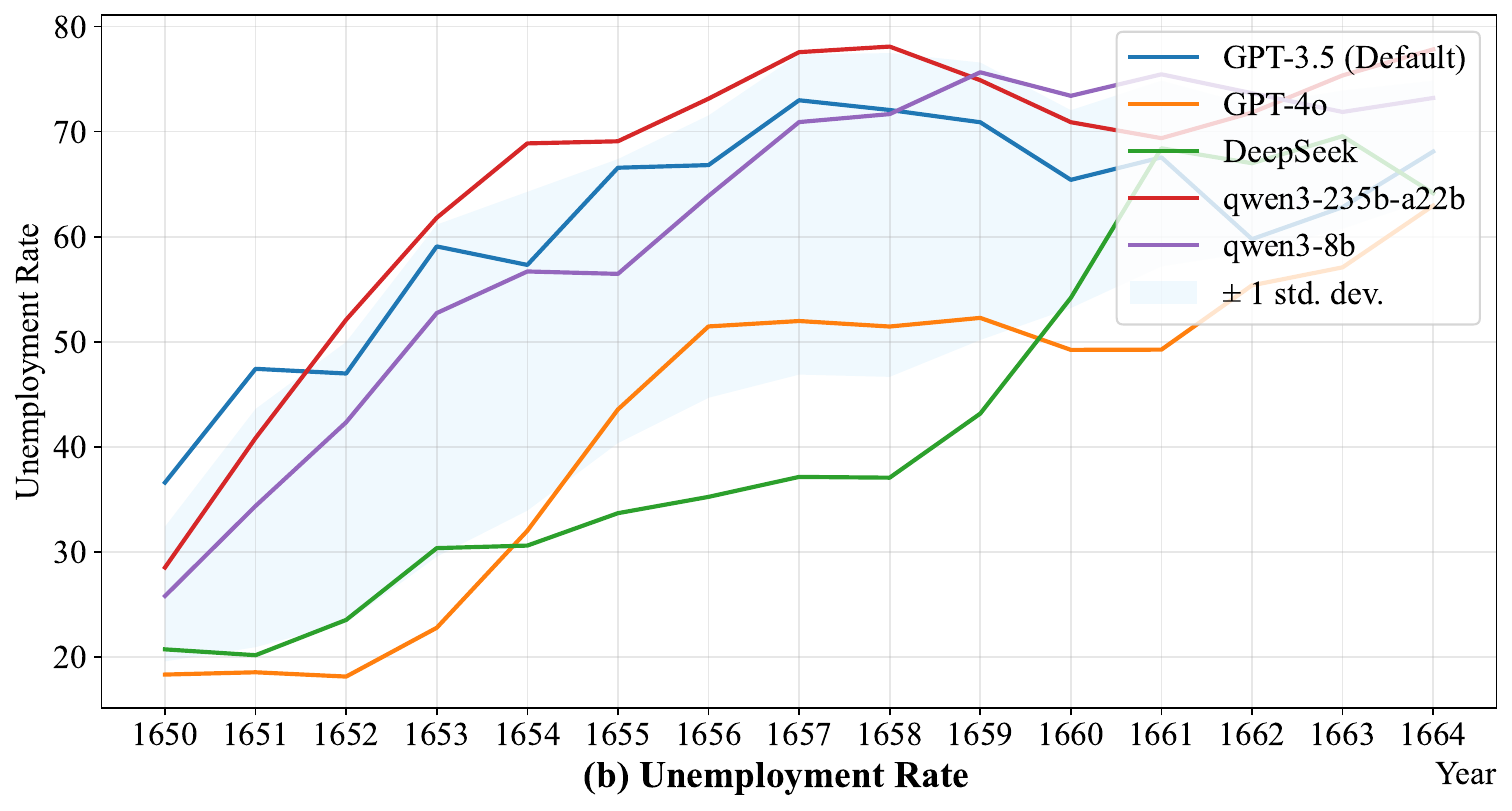} \\[1ex]
    \includegraphics[width=0.49\columnwidth]{ 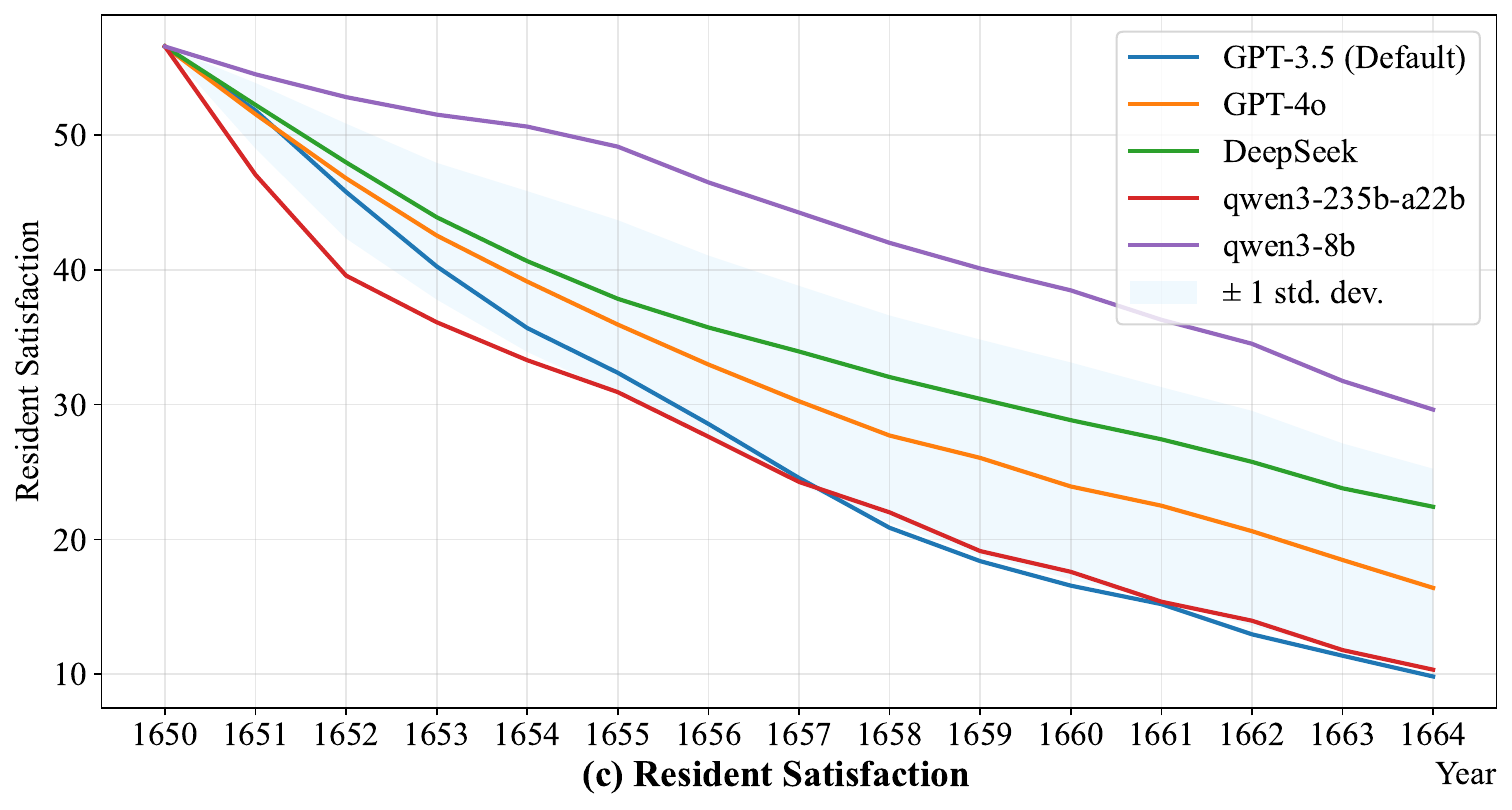}
    \includegraphics[width=0.49\columnwidth]{ 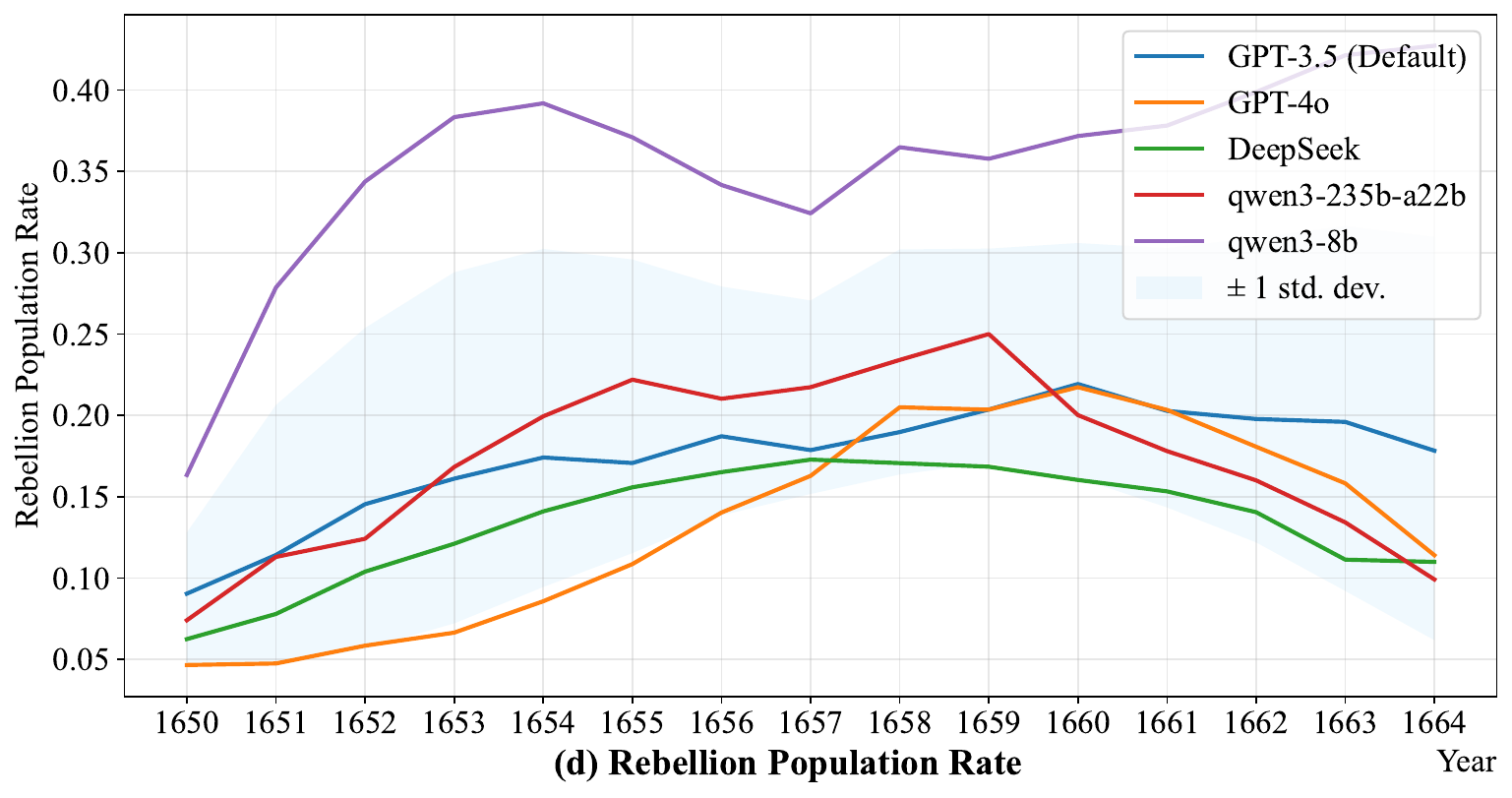}
  \caption{Consistency of simulation trends across different API providers.}
  \label{fig:api_analysis}
\end{figure}

As illustrated in Figure \ref{fig:api_analysis}, the simulation results across different LLMs exhibit a high degree of qualitative consistency. Specifically, Figure \ref{fig:api_analysis}(a) shows that all models successfully capture the inevitable collapse of river navigability due to siltation. 
This physical decay directly triggers a sharp rise in the unemployment rate (Figure \ref{fig:api_analysis}b) and a continuous decline in average social satisfaction (Figure \ref{fig:api_analysis}c). 
The social unrest, represented by the rebellion strength ratio in Figure \ref{fig:api_analysis}(d), consistently peaks as satisfaction hits critical lows before tailing off due to systemic exhaustion. 

In contrast, the Qwen3-8B model not only exhibits a significantly higher rebellion rate throughout the simulation but also displays a distinct and less consistent temporal trend.

Notably, as shown in Figure \ref{fig:api_analysis}(d), the Qwen3-8B model not only exhibits a significantly higher rebellion rate throughout the simulation but also displays a distinct and less consistent temporal trend. This divergence suggests potential challenges in Qwen3-8B's ability to fully capture the intricate dynamics and constraints of the simulated environment and agent behaviors. While other larger LLMs demonstrate a more stable and accurate understanding of the complex economic and social interdependencies,  the smaller Qwen3-8B model struggles with the intricate reasoning for policy evaluation and long-term behavioral consistency. Its higher rebellion rate could indicate a less precise interpretation of satisfaction thresholds or the mechanisms of rebellion propagation, leading to amplified conflict in the simulation.

The convergence of these trajectories across different model architectures demonstrates that the emergent socio-economic phenomena are driven by the internal logic of our framework rather than the specific biases of a single LLM, especially for larger, more capable models. This confirms that the framework is highly stable and robust to the choice of capable LLM provider for scenarios requiring complex reasoning. 

\section{Details Regarding Scalability and Robustness} 
% \section{System Performance Details} 
\label{app:performance}
\subsection{Scaling Effect Analysis}

\begin{figure}[!ht]
  \centering
    \includegraphics[width=0.49\columnwidth]{ 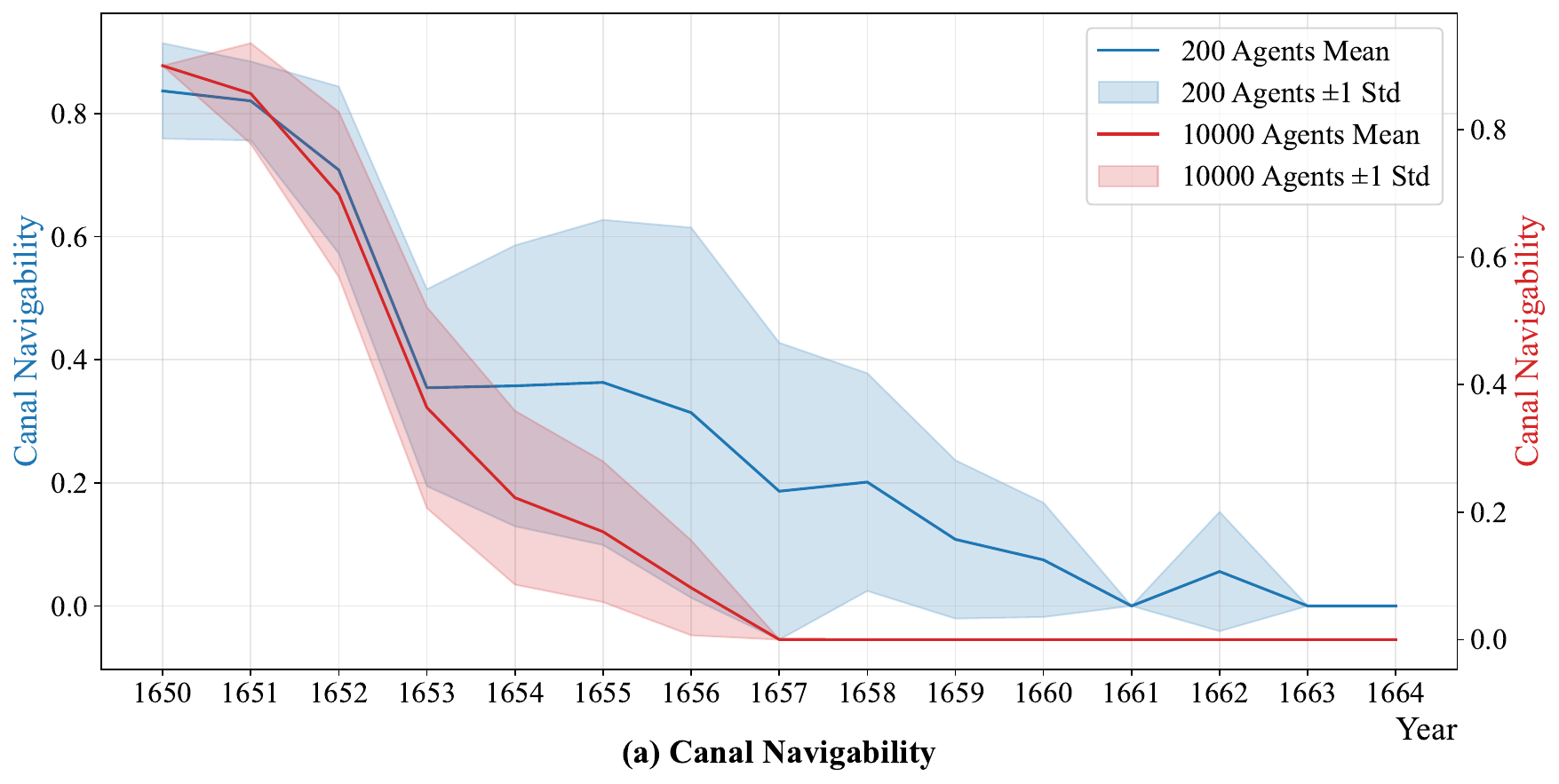} 
    \includegraphics[width=0.49\columnwidth]{ 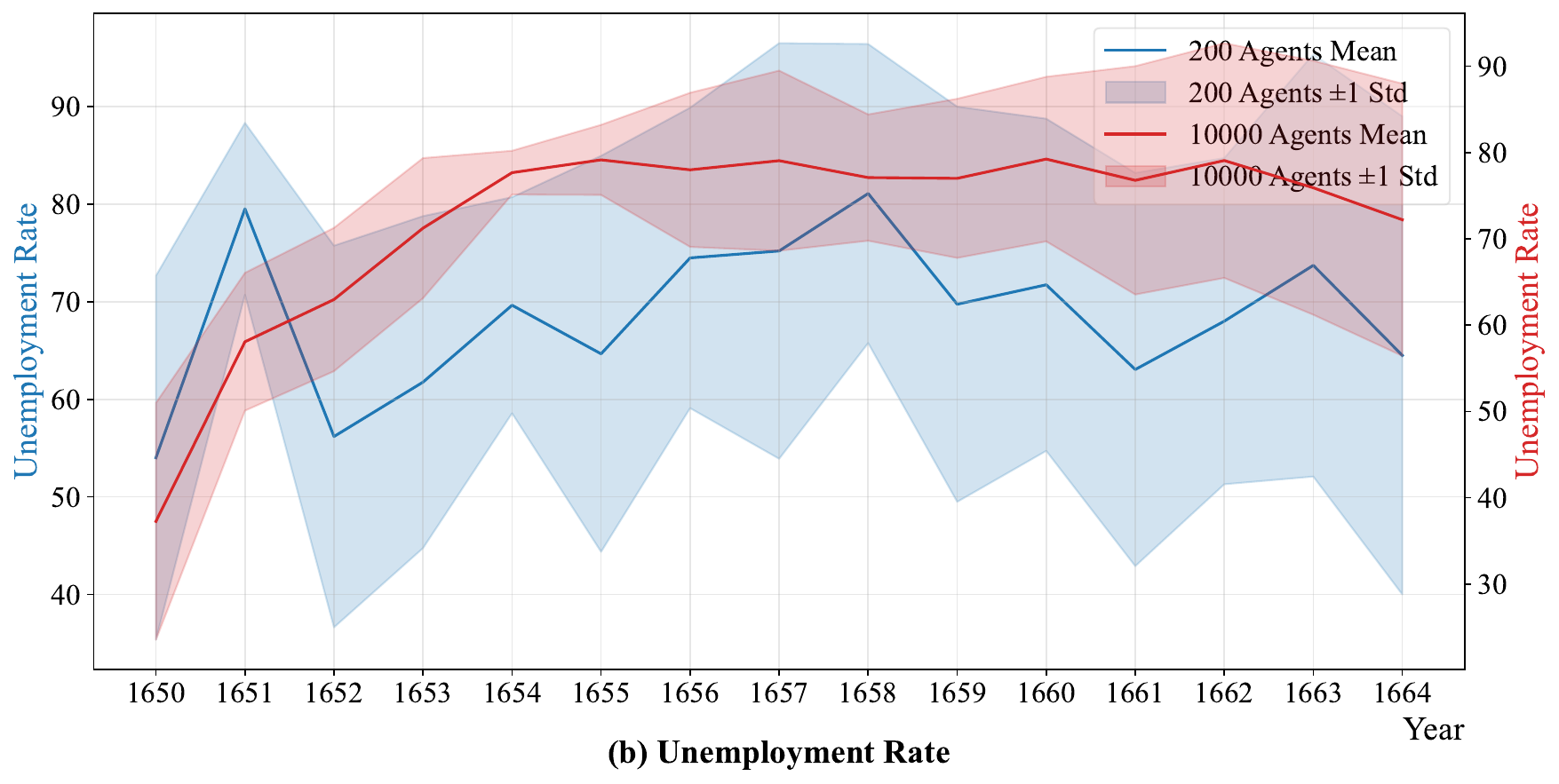} \\[1ex]
    \includegraphics[width=0.49\columnwidth]{ 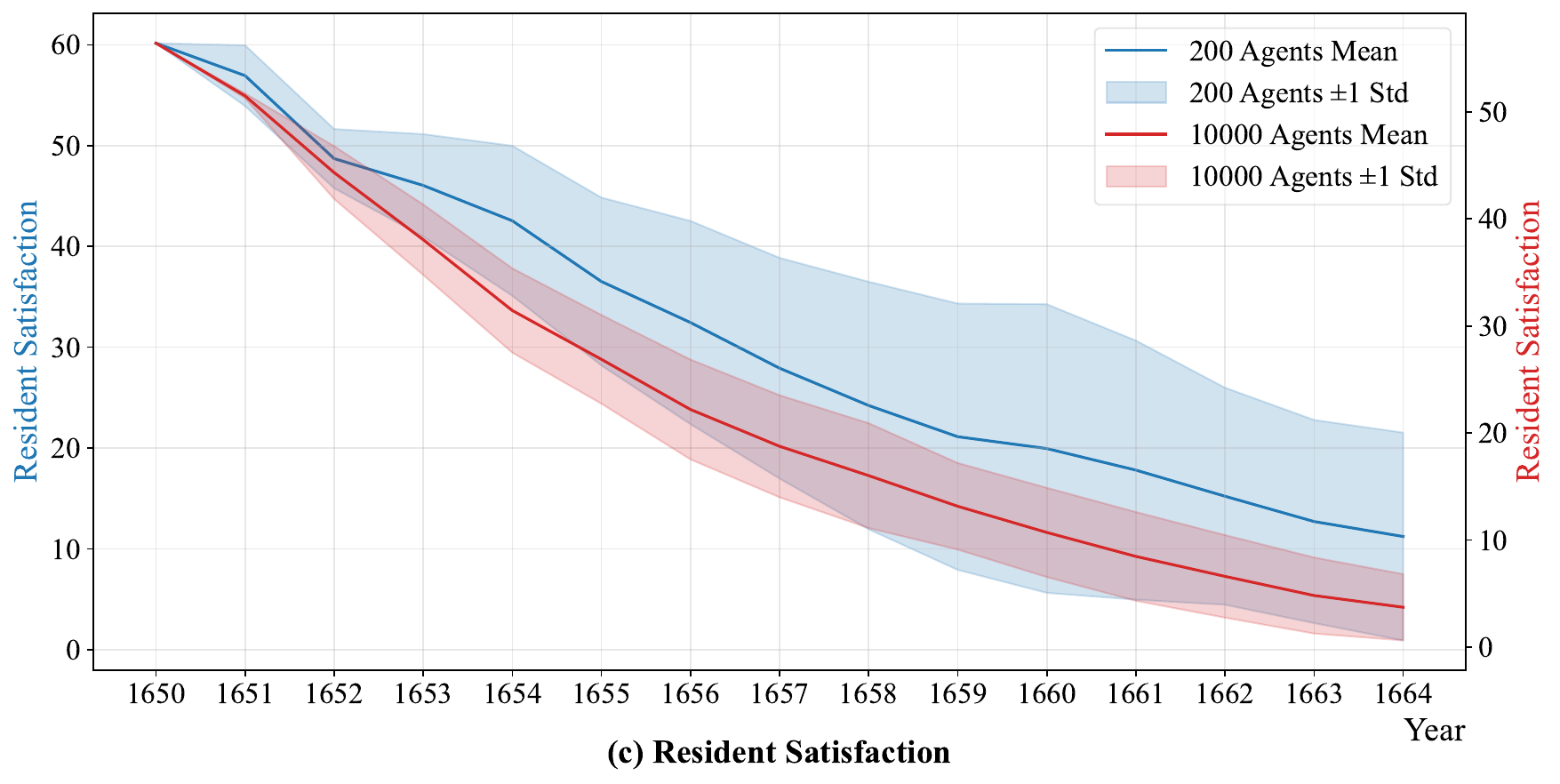}
    \includegraphics[width=0.49\columnwidth]{ 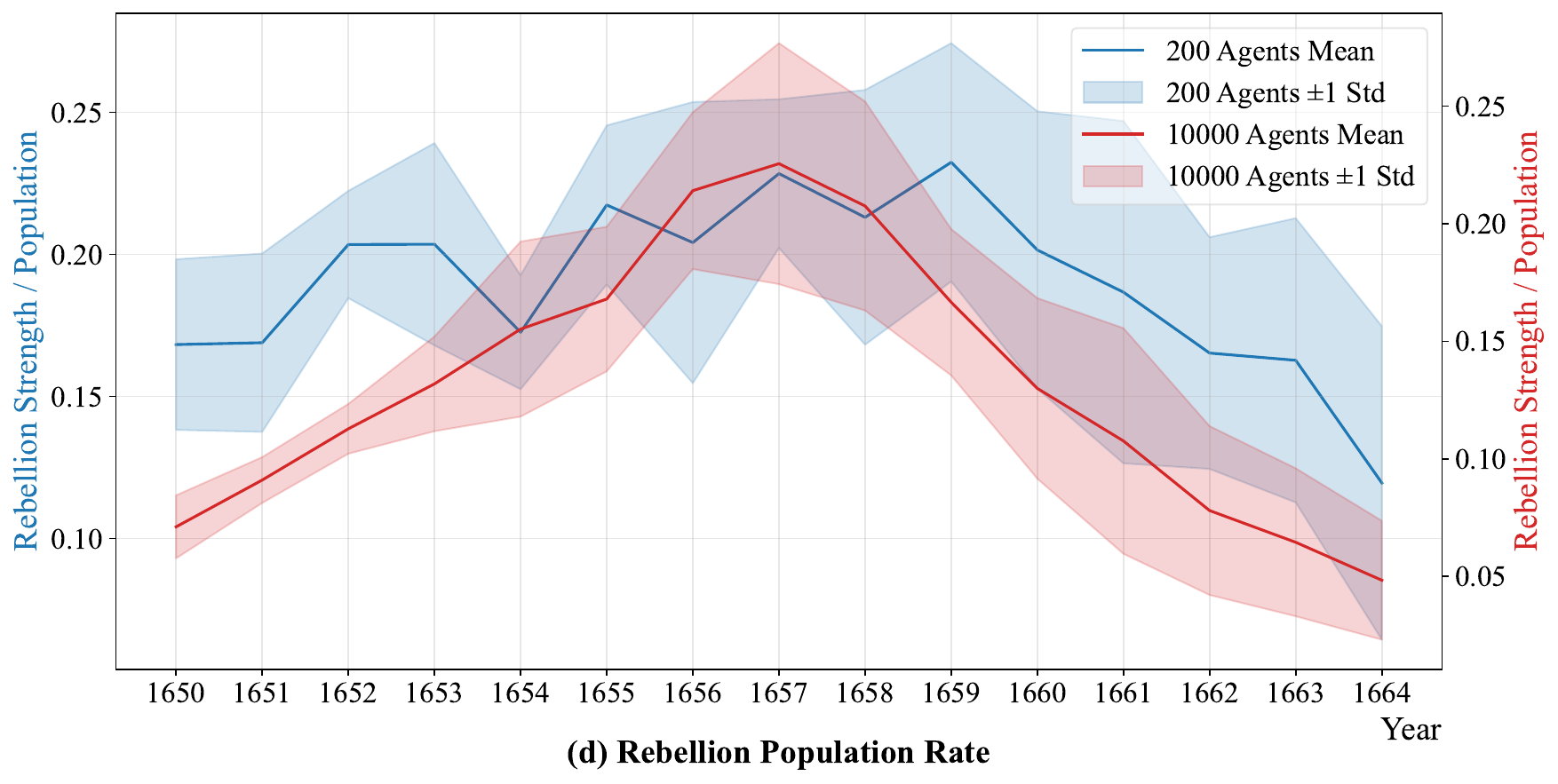}
  \caption{Comparison Between Different Agent Scale.}
  \label{fig:agent_scale}
\end{figure}

We investigate the scaling effects and behavioral robustness by comparing two configurations: 200 versus 10,000 agents. As shown in Figure \ref{fig:agent_scale}, simulations with 10,000 agents exhibit significantly mitigated fluctuations across all four sub-figures, as evidenced by consistently lower variance compared to the 200-agent case. Moreover, the larger agent scale yields both qualitative and quantitative differences in the observed outcomes. For example, in sub-figure (a), the 10,000-agent simulation terminates earlier and with markedly lower variance. Sub-figure (d) reveals a distinct transition occurring in 1657 under the 10,000-agent setting, which aligns precisely with the termination time observed in sub-figure (a). This correspondence can be attributed to the closure of the canal, which triggered substantial population outflow, primarily among communities residing along the canal and dependent on it for livelihood, thereby reducing the pool of potential participants in the rebellion.
Collectively, these results indicate that scaling up the agent population effectively reinforces the macroscopic robustness and clarity of the simulation outcomes, supporting more reliable and discernible conclusions.

\subsection{Efficiency Analysis}

We evaluate the computational efficiency of the \textit{Eco3S} framework by measuring runtime under varying simulation configurations, particularly with respect to the number of agents and simulation steps. As summarized in Table \ref{tab:efficiency_transposed}, execution time scales sublinearly with increases in both agent count and temporal horizon, demonstrating the framework’s capacity to robustly scale to large scale socio-economic modeling.
For instance, simulating the Canal Decay and Origins of Governance scenarios with 2,000 agents over 10 steps takes 30:56.2 and 39:05.1 minutes, respectively. When the number of steps is doubled to 20 (with agent count held constant at 2,000), runtime increases modestly to 33:05.8 and 59:06.0 minutes, reflecting the relatively low per-step computational overhead once the agent population is initialized.
By contrast, reducing the agent count to 500 (with 10 steps) yields runtime of 10:53.7 and 20:41.2 minutes, more than a quarter of the time required for 2,000 agents. Consistent with the theoretical time complexity($\mathcal{O}(\frac{N \times T}{P})$) analyzed, this sub-proportional scaling is achieved by leveraging concurrent processing to compute agent decisions in parallel, minimizing sequential bottlenecks in the simulation loop. 
Notably, the dominant constraint on scalability in practice is not the framework’s architecture, but external LLM platform rate limits which govern the throughput of language model queries.

% \begin{table*}[t]
% \centering
% \small
% \setlength{\tabcolsep}{4pt} % 略微缩小列间距以适应宽度
% \begin{tabular}{lcccccccccc}
% \toprule
% \textbf{Agents} & \multicolumn{3}{c}{100} & \multicolumn{3}{c}{500} & \multicolumn{3}{c}{2000} & 10000 \\
% \cmidrule(lr){2-4} \cmidrule(lr){5-7} \cmidrule(lr){8-10} \cmidrule(lr){11-11}
% \textbf{Steps}  & 5 & 10 & 20 & 5 & 10 & 20 & 5 & 10 & 20 & 5 \\ 
% \midrule
% Canal Decay     & 02:18.8 & 06:03.7 & 09:33.9 & 03:51.0 & 10:53.7 & 12:28.4 & 15:33.3 & 30:56.2 & 33:05.8 & 1:22:50.6 \\
% Origins of Gov. & 01:33.7 & 03:04.9 & 18:36.9 & 05:05.6 & 20:41.2 & 34:06.5 & 18:19.3 & 39:05.1 & 59:06.0 & -       \\
% \bottomrule
% \end{tabular}
% \caption{Comparison between different simulation scales}
% \label{tab:efficiency_transposed}
% \end{table*}

\begin{table}[!ht]
\centering
\small
\setlength{\tabcolsep}{5pt}
\renewcommand{\arraystretch}{1.08}
\begin{threeparttable}
\begin{tabular}{cccc}
\toprule
\textbf{Agents} & \textbf{Steps} & \textbf{Canal Decay} & \textbf{Origins of Gov.} \\
\midrule
\multirow{3}{*}{100}
 & 5  & 02:18.8 & 01:33.7 \\
 & 10 & 06:03.7 & 03:04.9 \\
 & 20 & 09:33.9 & 18:36.9 \\
\midrule
\multirow{3}{*}{500}
 & 5  & 03:51.0 & 05:05.6 \\
 & 10 & 10:53.7 & 20:41.2 \\
 & 20 & 12:28.4 & 34:06.5 \\
\midrule
\multirow{3}{*}{2,000}
 & 5  & 15:33.3 & 18:19.3 \\
 & 10 & 30:56.2 & 39:05.1 \\
 & 20 & 33:05.8 & 59:06.0 \\
\midrule
10,000 & 5 & 1:22:50.6 & -- \\
\bottomrule
\end{tabular}
\begin{tablenotes}[flushleft]
\footnotesize
\item Runtime is reported as MM:SS.s, or H:MM:SS.s for runs exceeding one hour. ``--'' indicates that the configuration was not evaluated.
\end{tablenotes}
\end{threeparttable}
\caption{Runtime comparison across different simulation scales.}
\label{tab:efficiency_transposed}
\end{table}

\section{System Interface and Operational Workflow}
\label{app:system_interface}
\begin{figure}[!ht]
  \centering
  \includegraphics[width=\columnwidth]{ 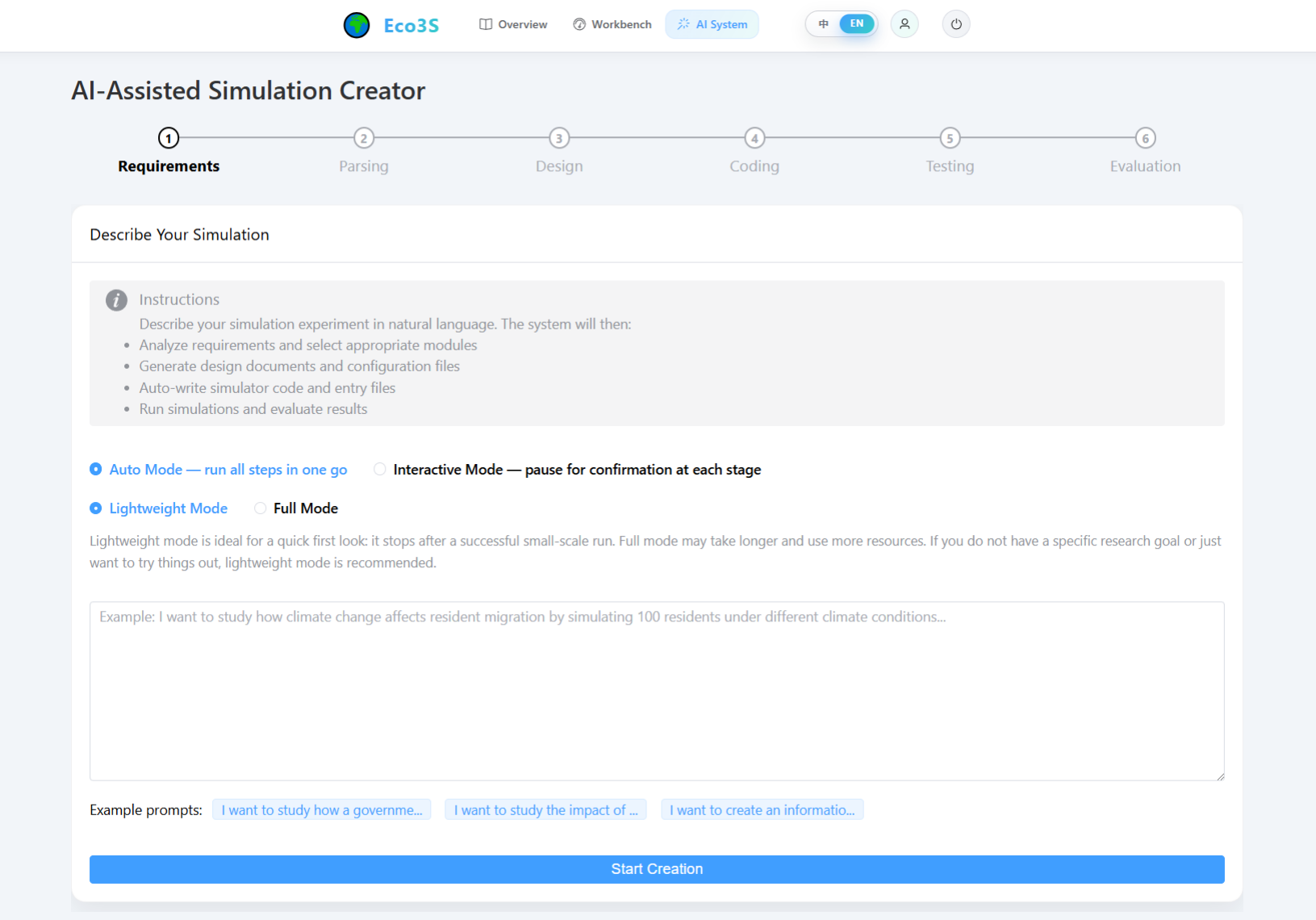}
  \caption{AI-assisted simulation creation interface.}
  \label{fig:ai_creation}
\end{figure}

Eco3S provides a web-based platform that supports the complete workflow from simulation creation to execution and monitoring. As shown in Figure~\ref{fig:ai_creation}, researchers can describe their simulation requirements in natural language through the AI-assisted creator interface.

\begin{figure}[!ht]
  \centering
  \includegraphics[width=\columnwidth]{ 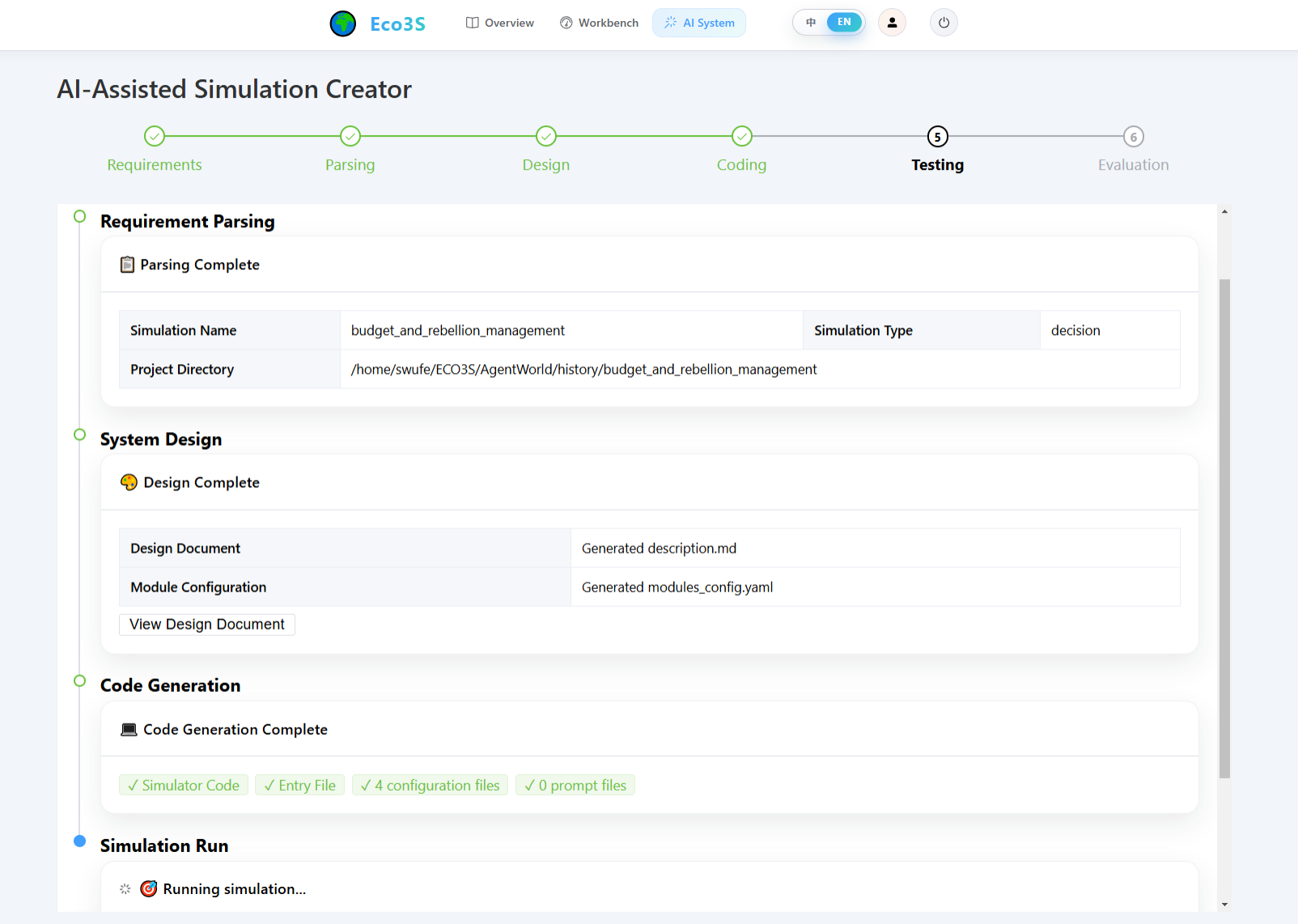}
  \caption{Runtime interface for the multi-phase simulation generation pipeline.}
  \label{fig:pipeline_runtime}
\end{figure}

The system translates these requirements into a structured simulation through a multi-phase generation pipeline. Figure~\ref{fig:pipeline_runtime} shows the corresponding runtime interface, where researchers can inspect the current phase, execution progress, and generation logs.

\begin{figure}[!ht]
  \centering
  \includegraphics[width=\columnwidth]{ 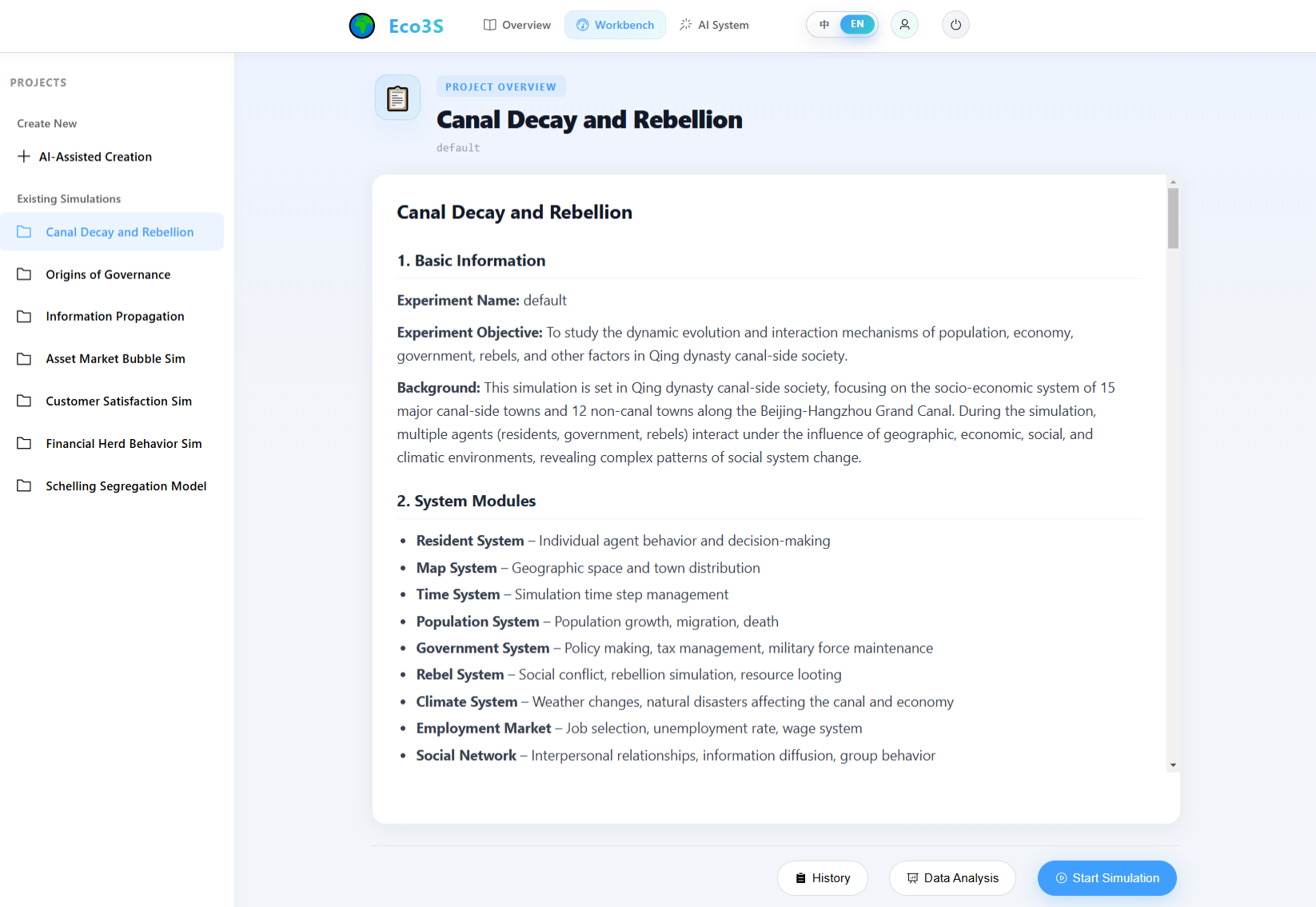}
  \caption{Dashboard for managing generated simulation projects.}
  \label{fig:system_dashboard}
\end{figure}

After generation, simulation projects and their configurations can be reviewed and managed through the dashboard shown in Figure~\ref{fig:system_dashboard}. This interface provides a centralized entry point for inspecting project descriptions and initiating subsequent operations.

\begin{figure}[!ht]
  \centering
  \includegraphics[width=\columnwidth]{ 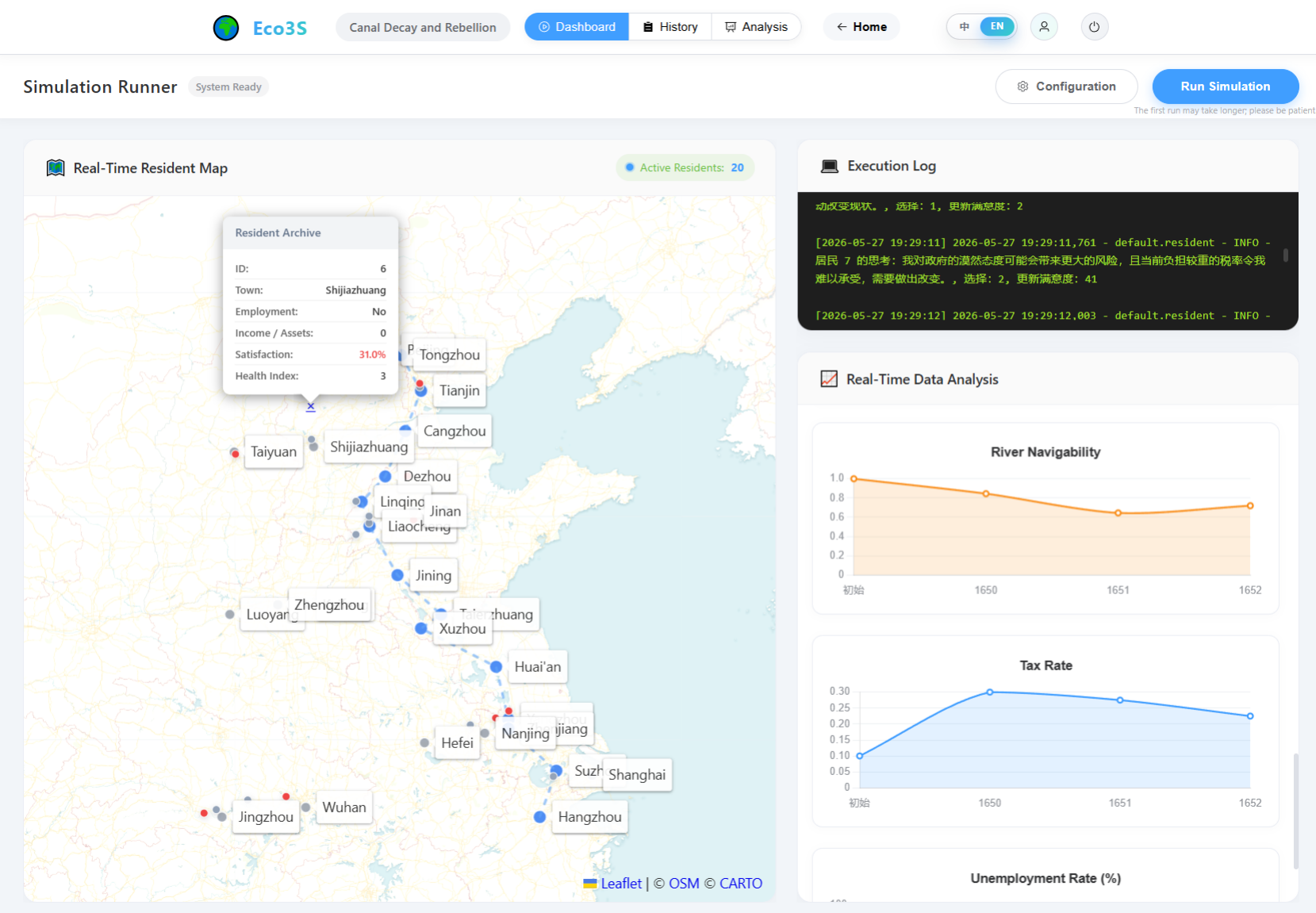}
  \caption{Simulation execution and real-time monitoring interface.}
  \label{fig:simulation_execution}
\end{figure}

During simulation execution, the monitoring interface in Figure~\ref{fig:simulation_execution} presents live charts of key metrics together with a dynamic map of agent locations and states. This synchronized view allows researchers to examine aggregate trends and spatial behaviors as the simulation evolves.

\section{Detailed Comparison with Existing Social Simulation Studies}

Supplementary Table\ref{tab:studies_comparison} situates Eco3S within representative agent-based social simulation studies spanning social media, societal phenomena, economic environments, and other application scenarios. We compare these studies along six capabilities that are central to the design of Eco3S: evolving environments, counterfactual mechanisms, reproduction of established economic studies, automated simulation construction, real-time human feedback, and automated result analysis. A checkmark indicates that the corresponding capability is explicitly supported or demonstrated in the cited work, whereas a cross indicates that it is not reported as part of the presented system. The comparison highlights that prior studies typically focus on a subset of these capabilities, while Eco3S integrates all six within a unified simulation platform.

\begin{table*}[!ht]
\caption{Comparison between different social simulation studies}
\label{tab:studies_comparison}
\centering
\footnotesize
\setlength{\tabcolsep}{3pt}
\resizebox{0.99\linewidth}{!}{%
\begin{tabular}{l l *{6}{c}}
\toprule
\textbf{Scenario} & \textbf{Paper} & \tabincell{c}{Env-evolution} & \tabincell{c}{Counterfactual\\ Mechanism} & \tabincell{c}{Economic Studies\\ Reproduction} & \tabincell{c}{Auto-simulation} & \tabincell{c}{Real-time \\ Human Feedback} & \tabincell{c}{Auto-analysis} \\
\midrule
\multirow{8}{*}{Social Media} & S3~\citep{gao2023s3} & \xmark & \xmark & \xmark & \xmark & \xmark & \xmark \\
& ~\citet{li2023quantifying} & \xmark & \xmark & \xmark & \xmark & \xmark & \xmark \\
& ~\citet{tornberg2023simulating} & \xmark & \xmark & \xmark & \xmark & \xmark & \cmark \\
& Y SOCIAL~\citep{rossetti2024social} & \xmark & \xmark & \xmark & \xmark & \xmark & \xmark \\
& ~\citet{mou2024unveiling} & \xmark & \xmark & \xmark & \xmark & \xmark & \xmark \\
& FPS~\citep{liu2024skepticism} & \xmark & \xmark & \xmark & \xmark & \xmark & \xmark \\
% & FUSE~\citep{liu2024tiny} & \xmark & \xmark & \xmark & \xmark & \xmark & \cmark \\
& ~\citet{wang2025decoding} & \xmark & \xmark & \xmark & \xmark & \xmark & \xmark \\
& ~\citet{Wang2025User} & \xmark & \xmark & \xmark & \xmark & \cmark & \xmark \\
\midrule
\multirow{12}{*}{Social Behavior} & ~\citet{xiao2023simulating} & \xmark & \cmark & \xmark & \xmark & \xmark & \xmark \\
& Generative Agents~\citep{park2023generative} & \xmark & \cmark & \xmark & \xmark & \xmark & \xmark \\
% & ~\citet{williams2023epidemic} & \xmark & \cmark & \xmark & \xmark & \xmark & \xmark \\
& ~\citet{chopra2025limits} & \xmark & \cmark & \xmark & \xmark & \xmark & \xmark \\
& CRSEC~\citep{Ren2024Emergence} & \xmark & \xmark & \xmark & \xmark & \xmark & \xmark \\
& Lyfe Agents~\citep{zhao2024lyfe} & \xmark & \xmark & \xmark & \xmark & \xmark & \xmark \\
& ~\citet{wu2024shall} & \xmark & \xmark & \xmark & \xmark & \xmark & \xmark \\
& ElectionSim~\citep{zhang2024electionsim} & \xmark & \xmark & \xmark & \xmark & \xmark & \xmark \\
& Concordia~\citep{touzel2024simulation} & \xmark & \xmark & \xmark & \xmark & \xmark & \xmark \\
& WarAgent~\citep{hua2023war} & \xmark & \xmark & \xmark & \xmark & \xmark & \xmark \\
& OASIS~\citep{yang2024oasis} & \xmark & \xmark & \xmark & \xmark & \xmark & \xmark \\
& SocioVerse~\citep{zhang2025socioverse} & \xmark & \xmark & \xmark & \xmark & \xmark & \xmark \\
& TeachTune~\citep{jin2025teachtune} & \xmark & \xmark & \xmark & \cmark & \cmark & \xmark \\
\midrule
\multirow{5}{*}{Economic Phenomenon} & EconAgent~\citep{li2024econagent} & \xmark & \xmark & \xmark & \xmark & \xmark & \xmark \\
& GenSim~\citep{tang2025gensim} & \xmark & \xmark & \xmark & \xmark & \xmark & \xmark \\
& CompeteAI~\citep{zhao2024competeai} & \xmark & \xmark & \xmark & \xmark & \xmark & \xmark \\
& AgentSociety~\citep{piao2025agentsociety} & \xmark & \xmark & \xmark & \xmark & \xmark & \xmark \\
& YuLan-OneSim~\citep{wang2025yulanonesim} & \xmark & \cmark & \xmark & \cmark & \xmark & \cmark \\
\midrule
& \textbf{\textit{Eco3S}(ours)} & \cmark & \cmark & \cmark & \cmark & \cmark & \cmark \\
\bottomrule
\end{tabular}%
}

\end{table*}

% \end{document}

\end{document}